\documentclass[iicol, sn-apa]{sn-jnl}%

\jyear{2022}%

\usepackage{times}
\usepackage[T1]{fontenc}

\usepackage{natbib}
\usepackage{multicol}

\usepackage{siunitx}
\usepackage{amsmath}
\usepackage{amssymb}
\usepackage{amsfonts}
\DeclareMathAlphabet{\mathcal}{OMS}{cmsy}{m}{n}

\usepackage{nicefrac}
\usepackage{soul}
\usepackage{graphics} %
\usepackage{epsfig} %
\usepackage{mathptmx} %
\usepackage{hyperref}
\usepackage[export]{adjustbox}

\usepackage{xcolor}%
\usepackage{pifont}
\usepackage{ragged2e}
\usepackage{booktabs}
\usepackage{adjustbox}
\usepackage{graphbox}
\usepackage{tabularx}
\usepackage{scrextend}

\usepackage{subcaption}

\usepackage{threeparttable}

\usepackage{xspace}
\newcommand{\methodname}{DiSECt\xspace}

\title{\methodname: A Differentiable Simulator for Parameter Inference and Control in Robotic Cutting}

\author*[1]{\fnm{Eric} \sur{Heiden}}\email{heiden@usc.edu}
\author[2]{\fnm{Miles} \sur{Macklin}}
\author[2]{\fnm{Yashraj} \sur{Narang}}
\author[2,3]{\fnm{Dieter} \sur{Fox}}
\author[2,4]{\fnm{Animesh} \sur{Garg}}
\author[2,5]{\fnm{Fabio} \sur{Ramos}}

\affil*[1]{\orgname{University of Southern California}, \orgaddress{\city{Los Angeles}, \country{USA}}}
\affil[2]{\orgname{NVIDIA}, \orgaddress{\country{USA}}}
\affil[3]{\orgname{University of Washington}, \orgaddress{\city{Seattle}, \country{USA}}}
\affil[4]{\orgname{University of Toronto \& Vector Institute}, \orgaddress{\city{Toronto}, \country{Canada}}}
\affil[5]{\orgname{University of Sydney}, \orgaddress{\city{Sydney}, \country{Australia}}}

\definecolor{codegreen}{rgb}{0,0.6,0}
\definecolor{codegray}{rgb}{0.5,0.5,0.5}
\definecolor{codepurple}{rgb}{0.58,0,0.82}
\definecolor{backcolour}{rgb}{0.95,0.95,0.92}

\lstdefinestyle{mystyle}{
    backgroundcolor=\color{backcolour},   
    commentstyle=\color{codegreen},
    keywordstyle=\color{magenta},
    numberstyle=\tiny\color{codegray},
    stringstyle=\color{codepurple},
    basicstyle=\ttfamily\footnotesize,
    breakatwhitespace=false,         
    breaklines=true,                 
    captionpos=b,                    
    keepspaces=true,                 
    numbers=left,                    
    numbersep=5pt,                  
    showspaces=false,                
    showstringspaces=false,
    showtabs=false,                  
    tabsize=2
}

\begin{document}

\newcommand{\traj}[0]{\zeta}

\abstract{
Robotic cutting of soft materials is critical for applications such as food processing, household automation, and surgical manipulation. As in other areas of robotics, simulators can facilitate controller verification, policy learning, and dataset generation. Moreover, \textit{differentiable} simulators can enable gradient-based optimization, which is invaluable for calibrating simulation parameters and optimizing controllers. In this work, we present \methodname: the first differentiable simulator for cutting soft materials. The simulator augments the finite element method (FEM) with a continuous contact model based on signed distance fields (SDF), as well as a continuous damage model that inserts springs on opposite sides of the cutting plane and allows them to weaken until zero stiffness, enabling crack formation. Through various experiments, we evaluate the performance of the simulator. We first show that the simulator can be calibrated to match resultant forces and deformation fields from a state-of-the-art commercial solver and real-world cutting datasets, with generality across cutting velocities and object instances. We then show that Bayesian inference can be performed efficiently by leveraging the differentiability of the simulator, estimating posteriors over hundreds of parameters in a fraction of the time of derivative-free methods. Next, we illustrate that control parameters in the simulation can be optimized to minimize cutting forces via lateral slicing motions. Finally, we conduct experiments on a real robot arm equipped with a slicing knife to infer simulation parameters from force measurements. By optimizing the slicing motion of the knife, we show on fruit cutting scenarios that the average knife force can be reduced by more than $40\%$ compared to a vertical cutting motion. We publish code and additional materials on our project website at \url{https://diff-cutting-sim.github.io}.

}


\maketitle

\section{Introduction}
\label{sec:intro}

Robotic cutting of soft materials is critical for various real-world applications, including food processing, household automation, surgical manipulation, and manufacturing of deformable objects. As in other areas of robotics, simulators can allow researchers to verify controllers, train control policies, and generate synthetic datasets for cutting, as well as avoid expensive or time-consuming real-world trials. In addition, cutting is inherently destructive and irreversible; thus, accurate and efficient simulators are indispensable for automating safety-critical tasks like robotic surgery.

However, simulating the cutting of soft materials is challenging. Cutting involves diverse physical phenomena, including contact, friction, elastic deformation, damage, plastic deformation, crack initiation, and/or fracture. A variety of simulation methods have been introduced, including mesh-based approaches (e.g., extensions of the finite element method (FEM)), particle-based approaches (e.g., smoothed particle hydrodynamics (SPH)), and hybrid techniques. For accuracy, these methods often require computationally-expensive re-meshing, or simulation of millions of particle interactions.

\begin{figure}[t]
    \centering
    \includegraphics[width=.95\columnwidth]{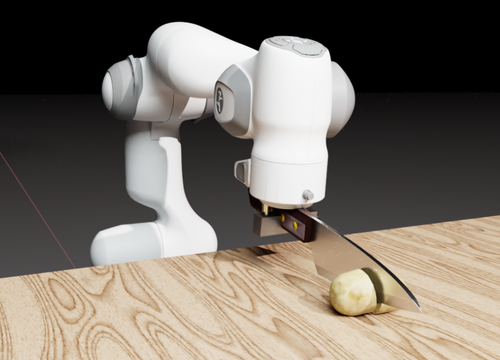}
    \caption{A rendering from our differentiable cutting simulator. \methodname provides accurate gradients of the cutting process, allowing us to efficiently fit model parameters to real-world measurements, and optimize cutting motions.}
    \label{fig:teaser}
\end{figure}

Moreover, a physically-accurate forward simulator is necessary but not sufficient for accurate and useful predictions. The material properties of real-world soft materials (e.g., ripening fruits, diseased tissue) are often unknown and highly heterogeneous, mandating calibration of material models. In addition, ideal trajectories for cutting may not be known beforehand, requiring efficient optimization of control actions. In other fields, the need to infer material and control parameters has motivated the development of differentiable simulators, which can harness efficient, gradient-based optimization methods.

We present \methodname, the first differentiable simulator for the cutting of soft materials (\autoref{fig:teaser}). The simulator has several key features. First, contact between the cutting instrument (i.e., knife) and the object is resolved using a continuous signed distance field (SDF) representation of the knife. Second, a mesh-based representation of the object is used; given a predefined cutting plane, virtual nodes are inserted along the surface in a preprocessing step that occurs only once. Third, cracks are introduced using a continuous damage model, where springs are inserted on either side of the cutting plane. Under application of force over time, the springs can progressively weaken until zero stiffness, producing a crack. The continuous contact and damage models enable differentiability, and the springs and particles provide hundreds of degrees of freedom that can be used to calibrate the simulator against ground-truth data. Gradients of any simulation parameter are computed using automatic differentiation via source-code transformation.

This work offers the following contributions:
\begin{enumerate}
    \item The first differentiable simulator for cutting soft materials.
    \item A comparison of gradient-based and derivative-free methods for calibrating the simulator against ground-truth data from commercial solvers and real-world datasets. The comparison demonstrates that the differentiability of the simulator enables highly efficient estimation of posteriors over hundreds of simulation parameters.
    \item A performance evaluation of the calibrated simulator for unseen cutting velocities and object instances. It demonstrates that the simulator can accurately predict resultant forces and nodal displacement fields, with simulation speeds that are orders of magnitude faster than a comparable commercial solver.
    \item An application of the simulator to robotic control, optimizing knife motion trajectories to minimize cutting forces under time constraints.
    \item Experimental results on a real robot that validate our approach in parameter inference and trajectory optimization, where optimized slicing trajectories yield cutting motions that require significantly less force than vertical cuts.
\end{enumerate}

This article is an extended version of our work on DiSECt which we presented at the Robotics: Science and Systems (RSS) 2021 conference \citep{heiden2021disect}. Besides the extensive supplemental materials that include additional results and details of our work, we present new results from our own real-robot experiments in \autoref{sec:robot-opt-cutting} where we collected a dataset of force profiles and 3D meshes of common fruits, which we plan to release in conjunction with our already published simulation code\footnote{DiSECt source code: \url{https://github.com/NVlabs/DiSECt}}. By applying our pipeline of parameter inference, parameter transport to novel mesh topologies, and slicing trajectory optimization to such real data, we achieve efficient slicing motions that exhibit a strong reduction in required knife force to cut common fruits. Additionally, we provide code snippets in \autoref{sec:code-snippets} to serve as tutorial for the applications of DiSECt, which include parameter inference and knife motion optimization.

\section{Related Work}
\label{sec:related}

\subsection{Modeling and Simulation of Cutting}

\noindent
\paragraph*{Analytical modeling}
Cutting is a branch of elastoplastic fracture mechanics~\citep{atkins2009science}, where theoretical analysis has primarily focused on metal cutting~\citep{merchant1945mechanics} and brittle materials~\citep{griffith1921fracture}. 
A comprehensive treatment of the mechanics involved in cutting of metals, biomaterials and non-metals is given in~\cite{atkins2009science}. 
Analytical models of the forces acting on a knife as it cuts through soft materials have been derived in~\cite{zhou2006cut1, mu2019robotic}.

\noindent
\paragraph*{Mesh-based simulation}
Among the numerical methods that implement such analytical models, the Finite Element Method (FEM)~\citep{hughes2012fem, belytschko2013nonlinear} is a commonly used technique to simulate deformable bodies. FEM solves partial differential equations over a given domain (defined by a mesh) by discretizing it into simpler elements, such as tetrahedra (which we use in this work). 
Without modifying the mesh topology, the Extended FEM (X-FEM)~\citep{moes1999xfem} augments mesh elements by enrichment functions to model fracture mechanics processes~\citep{khoei2014extended,jerabkova2009xfem,koschier2017xfem}.

In classical FEM, topological changes that result from cutting and other fracture processes mandate an adaption of the mesh resolution so that the propagation of the crack can be accurately simulated. Approaches in mechanical engineering \citep{areias2017steiner} and computer graphics~\citep{wu2015survey,bielser2003state,burkhart2010adaptive,koschier2014adaptive,paulus2015virtual} have been introduced that re-mesh the simulation domain to accommodate cuts and other forms of cracks.

Virtual node algorithms (VNA)~\citep{molino2004vna,sifakis2007arbitrary,wang2014vna} duplicate mesh elements that intersect with the cutting surface, resulting in elements with portions of real material and empty regions. At the cutting interface, virtual nodes are introduced that allow for a two-way coupling of contact and elastic forces with the underlying mesh of the separated parts. We leverage VNA to augment our FEM simulation with extra degrees of freedom to allow the fine-grained simulation of contact between the material and the knife. Furthermore, by augmenting the mesh only once at the beginning of the cut, we avoid discontinuous re-meshing operations and are able to compute gradients from our simulator.

\paragraph*{Mesh-free and hybrid approaches}
So-called mesh-free Lagrangian methods, such as smoothed-particle hydrodynamics (SPH)~\citep{monaghan1977sph,monaghan1992sph}, element-free Galerkin (EFG)~\citep{belytschko1994efg}, and smoothed-particle Galerkin (SPG)~\citep{wU2017spg}, simulate continuous media by many particles that cover the simulation domain and the dynamics is determined by a kernel that defines the interaction between spatially close particles.
Hybrid approaches have been proposed, such as the Material Point Method (MPM)~\citep{hu2018mlsmpmcpic, wang2019mpm, wolper2019cdmpm, wolper2020anisompm}, that combine Eulerian (grid-based) and Lagrangian (particle-based) techniques. 
Position-based Dynamics (PBD) \citep{muller2007position} has been explored to simulate cutting in interactive surgery simulators~\citep{berndt2017pbd, pan2015real}.

\noindent
\paragraph*{Robotic cutting}
In~\cite{long2013robotic}, a robotic meat-cutting system is introduced that scans objects online to simulate deformable objects and uses impedance control to steer the knife.
In~\cite{thananjeyan2017multilateral,swirl-ijrr18}, a simplified mass-spring model is used for learning to cut planar surfaces. Such trained policies are then transferred to a physical robotic surgery system.
By studying the mechanics of cutting, \cite{mu2019robotic} derive control strategies involving pressing, pushing, and slicing motions. Similarly, \cite{zhou2006cut1, zhou2006cut2} analyze cutting ``by pressing and slicing,'' while taking into account the blade sharpness of the knife. Constrained optimization is used in~\cite{wijayarathne2020cut} to optimize cutting trajectories while accounting for contact forces, whereas in our control experiment we furthermore account for the coupling of elastic and contact force between the knife and the deformable object being cut.
In \cite{jamdagni2019robotic}, FEM simulation for robotic cutting is devised where crack propagation is simulated at the cross sections on a high-resolution 2D mesh, which is coupled with a coarser 3D mesh simulation. Meshes are obtained from laser scans of biomaterials, and force profiles recorded from a force sensor attached to a robot end-effector as it vertically cuts through various foodstuffs. In this work, we leverage this dataset of real-world cutting trajectories.
Fully data-driven models have been learned to facilitate model-predictive control in robotic cutting~\citep{lenz2015deepmpc,mitsioni2019mpc}.

\subsection{Differentiable Simulation}

Differentiable simulation has gained increasing attention, as it allows for the use of efficient gradient-based optimization algorithms to tune simulation parameters or control policies \citep{giftthaler2017autodiff,carpentier2018analytical,peres2018lcp, heiden2019ids, koolen2019rbd-julia, hu2020difftaichi, qiao2020scalable, heiden2020lidar, geilinger2020add, murthy2021gradsim, heiden2021neuralsim}.
Although finite differencing may be used to approximate the derivative of a simulation output, this approach suffers from accuracy problems and does not scale to large numbers of parameters~\citep{Margossian_2019}. In this work we employ reverse-mode automatic differentiation to obtain gradient information via source-code transformation~\citep{griewank_ad, innes2019zygote}. Recent work has shown the potential for source-code transformation to generate efficient parallel kernels for reverse-mode differentiation using graphics-processing units (GPUs)~\citep{hu2019chainqueen, hu2020difftaichi}.

\subsection{Parameter Inference for Simulators}

Simulation-based inference is a methodology that has emerged across various fields of science~\citep{cranmer2020sbi}. In parameter identification, optimization-based approaches are often utilized to obtain point estimates of the simulation parameters (such as for constitutive equations~\citep{mahnken2017identification, hahn2019real2sim}) that minimize the model error as measured from the dynamics of the real system. More often, when analytical gradients are unavailable or too expensive to obtain, optimization with finite differencing or gradient-free methods have been applied~\citep{mehta2020user, narang2021tactile}, particularly for dynamical systems through system identification~\citep{ljung1999sysid}.

Probabilistic inference techniques, on the other hand, seek to infer a distribution of simulation parameters that allows downstream applications to evaluate the uncertainty of the estimates. Such methods have been applied to learn conditional densities of simulation parameters given trajectories from the simulator and the real system~\citep{toni2008abc,papamakarios2016epsilon,ramos2019bayessim,hsu2019likelihoodfree,matl2020granular,matl2020stressd,heiden2022pds}.

Parameter inference may also be integrated in a closed-loop system~\citep{chebotar2019closing,ramos2019bayessim,mehta2019adr,mozifian2019learning}, where the currently estimated posterior over simulation parameters guides domain randomization and policy learning. Such approaches iteratively reduce the sim2real gap.

\section{\methodname: Differentiable Simulator for Cutting}
\label{sec:model}

The design of our simulator for cutting biomaterials is motivated by three aspects:
\begin{enumerate}
    \item The model has to be physically plausible such that the effects of changing physical quantities from continuum and fracture mechanics can be observed realistically.
    \item While the cutting process is an inherently discontinuous operation, wherever possible, gradients of the simulation parameters must be efficiently obtainable.
    \item The simulator must allow for sufficient degrees of freedom so that fine differences in material properties can be identified at localized places where the knife cuts through a heterogeneous material.
\end{enumerate}

\begin{figure}
    \centering
    \includegraphics[width=\columnwidth]{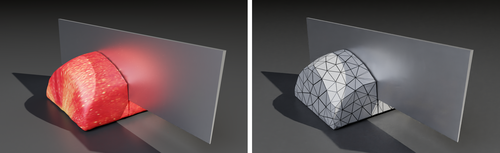}
    \caption{Visualization of an apple slice. We use a tetrahedral FEM-based model of the apple generated from scanned real-world data~\citep{jamdagni2019robotic}.}
    \label{fig:fem_mesh}
\end{figure}

\begin{figure}
    \centering
    \includegraphics[width=0.32\columnwidth,trim=0 0 0 0,clip]{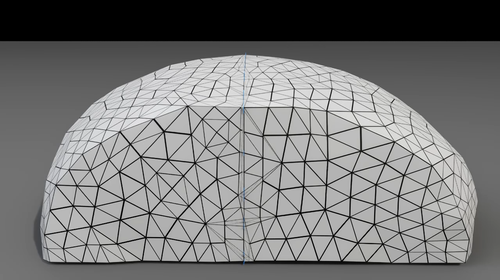}
    \hfill
    \includegraphics[width=0.32\columnwidth,trim=0 0 0 0,clip]{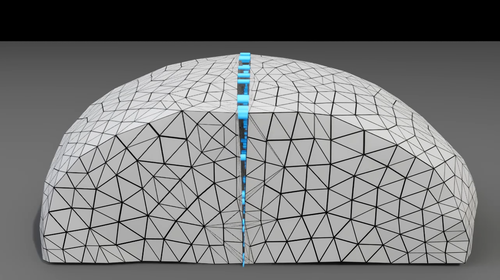}
    \hfill
    \includegraphics[width=0.32\columnwidth,trim=0 0 0 0,clip]{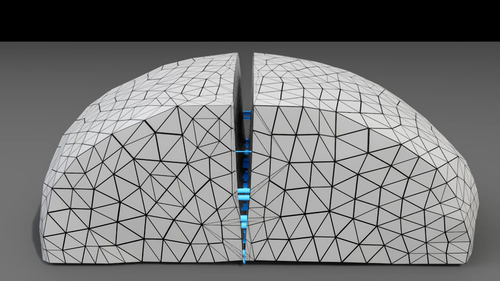}
    \caption{To smoothly model damage, we insert springs (shown in blue) between cut elements. We incrementally reduce the spring stiffness based on damage over time, proportional to the contact force applied from the knife. From left at $t=\SI{0}{\second}$, $t=\SI{0.8}{\second}$, $t=\SI{1.2}{\second}$. Here, we have removed the knife from visualization to show the inserted springs clearly.}
    \label{fig:cut_spring_evolution}
\end{figure}

\subsection{Continuum Mechanics}
\label{sec:fem}

We implement the Finite Element Method (FEM) to simulate the dynamics of the soft materials used throughout this work. Based on a tetrahedral discretization of the cutting target object, elastic forces are computed that take into account material properties, such as Young's modulus, Poisson's ratio, and density.

We employ a Neo-Hookean constitutive model following the strain-energy density function from~\cite{smith2018neohookean}, which has been designed to model biological tissue and preserve volume when the mesh undergoes large deformations:
\begin{align}
    \Psi = \frac{\mu}{2}(I_C - 3) + \frac{\lambda}{2}(J-\alpha)^2 - \frac{\mu}{2}\text{log}(I_C + 1)\label{eq:neohookean}.
\end{align} 
Here $\lambda, \mu$ are the Lam\'e parameters\footnote{Lam\'e parameters are a reformulation of Young's modulus and Poisson's ratio; hence we use these terms interchangeably.} and $\alpha$ is a constant. $J = \text{det}(\mathbf{F})$ is the relative volume change, $I_C = \text{tr}(\mathbf{F}^T\mathbf{F})$ is the first invariant of the Cauchy-Green deformation tensor, and $\mathbf{F}$ is the deformation gradient. We give the material properties for our experimental objects in \autoref{tab:materials}. Integrating over each tetrahedral element and summing the energy from \eqref{eq:neohookean} over all elements gives the total elastic potential energy. Forces $\mathbf{f}_{\text{elastic}}$ are derived from the energy gradient analytically and integrated using a semi-implicit Euler scheme. In addition to elastic forces, we include a strain-rate dissipation potential to model internal damping \citep{banderas2018}.

\begin{figure}
    \centering
    \includegraphics[width=\columnwidth]{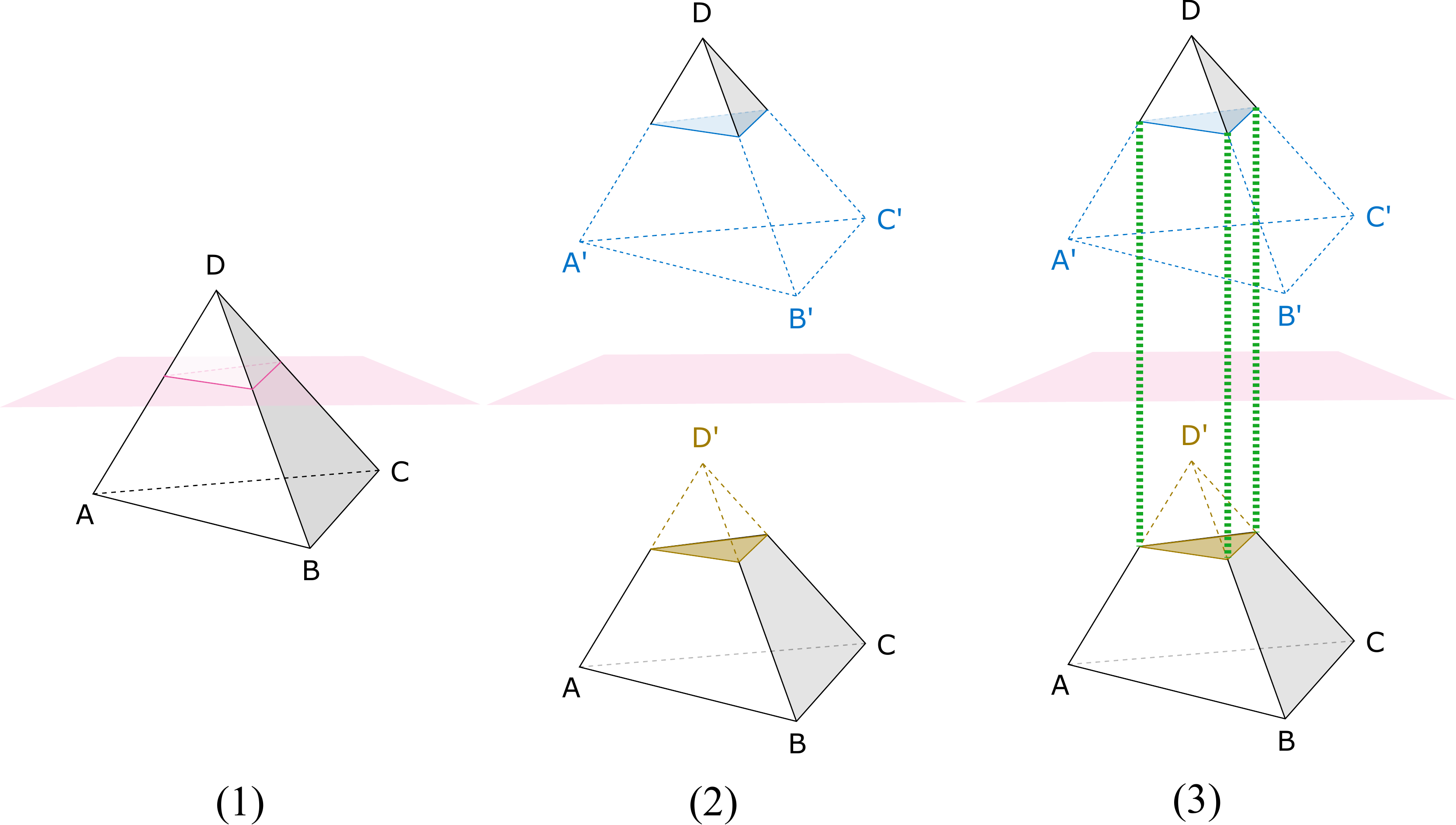}
    \caption{Pre-processing of a tetrahedral mesh (visualized by a single tet) prior to being cut along a given cutting surface (red plane).
    Given the original mesh, (1) intersecting elements are identified, and (2) duplicated such that the intersection geometry is retained in each part of the cut elements. Virtual nodes are inserted where the edges are intersecting the cutting surface. (3) The original and duplicated elements are connected by springs (green) between the virtual nodes.}
    \label{fig:preprocessing}
\end{figure}

\subsection{Mesh Pre-Processing}
\label{sec:preprocess}

Before the simulation begins, the cutting surface for the entire cut is given as a triangle mesh. We implement the Virtual Node Algorithm from~\cite{sifakis2007arbitrary}, which, as shown in \autoref{fig:preprocessing}, duplicates the mesh elements (tets) that intersect with the cut (subfigure 2), so that the tet above the cutting surface has a portion of material above the cut, and one empty portion below the cut (and vice versa for the reciprocal element). Next, we insert virtual nodes at the intersection points on the edges. These nodes are only represented by their barycentric coordinate $u\in[0,1]$.

Similar to the two-way coupled ``soft particles'' described in~\cite{sifakis2007hybrid}, each virtual node's 3D position $\tilde{\mathbf{x}}$ and velocity $\tilde{\mathbf{v}}$ is defined entirely by the coordinates of its two parent vertices, indexed $i$ and $j$:
$$
\tilde{\mathbf{x}} = (1-u) \mathbf{x}_i + u \mathbf{x}_j  \quad \qquad \tilde{\mathbf{v}} = (1-u) \mathbf{v}_i + u \mathbf{v}_j.
$$
The edge sections along the non-empty portions of the duplicated mesh elements participate in contact dynamics and propagate the resulting contact forces back to their parent nodes, as described in~\autoref{sec:contact}. Edge sections from the empty portions of the mesh create no collisions and are solely updated from the FEM dynamics of the parent vertices.

In the final step of this mesh preprocessing phase (subfigure 3), springs are inserted that connect the virtual nodes on both sides of the cutting surface which originally belonged to the same edge of the mesh. These springs allow us to simulate damage occurring during the cutting process in an entirely continuous (and thereby differentiable) manner, by weakening their stiffness values as knife contact forces are applied over time.

While our approach in its current form assumes that the entire cutting surface is given before the cutting simulation begins, interactive cutting applications could be accommodated by interweaving this mesh augmentation step with the actual cutting simulation. Nonetheless, the practicality of using such an interactive approach for parameter inference remains to be validated.

\subsection{Contact Dynamics}
\label{sec:contact}

\begin{figure}
    \centering
    \includegraphics[width=\columnwidth]{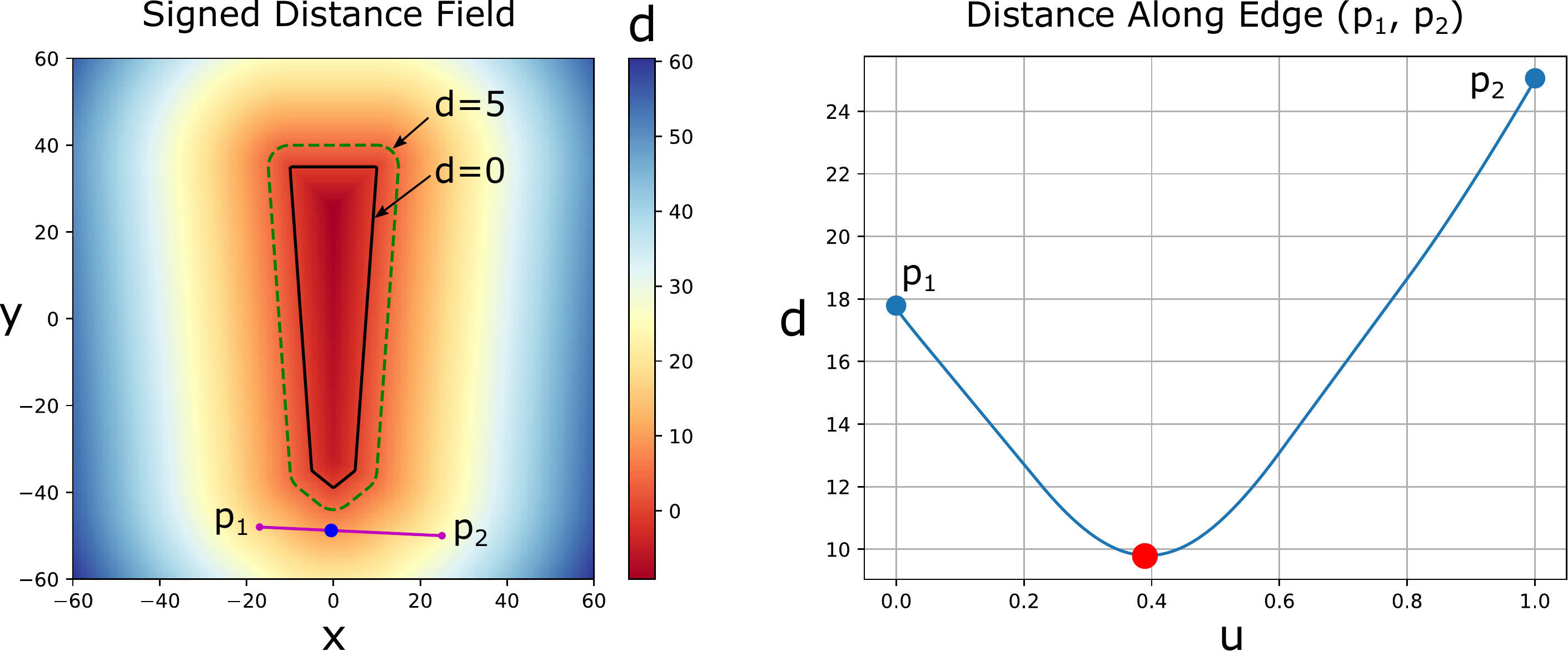}
    \caption{2D slice of the signed distance field (SDF) for the knife shape (not true to scale), where the color indicates distance $d$. The knife's boundary at $d=0$ is indicated by a solid black contour line.
    \textit{Left:} at distances greater than zero the SDF becomes more rounded.
    \textit{Right:} distance $d$ along the edge $(p_1,p_2)$ with barycentric edge coordinate $u$ varying between $0$ ($p_1$) and $1$ ($p_2$). The closest point found by Algorithm~\ref{alg:frank-wolfe} is shown in red.
    Distances are exemplary and do not represent the actual dimensions used in the simulator (see Appendix~\ref{tab:parameters}).}
    \label{fig:knife_sdf}
\end{figure}

Following \cite{macklin2020sdf}, we implement a contact model that represents the knife shape by a signed distance function (SDF) (\autoref{fig:knife_sdf}) that interacts with the tetrahedral mesh of the object being cut. To find the closest point between an edge from the mesh and the SDF, we run 20 iterations of the Frank-Wolfe algorithm (Algorithm~\ref{alg:frank-wolfe} in appendix), that uses the gradient information from the SDF to find a locally optimal solution for the barycentric coordinate $u\in[0,1]$ with the smallest distance. Using this coordinate, we can compute the penetration depth and contact normal by querying the SDF and its gradient.
Penalizing collisions, the contact normal force is computed as the squared penetration depth in the direction of the normal (zero if no collision). In combination with the relative velocity between the knife and the mesh vertices, friction forces are computed following the continuous friction model from~\cite[Equation 4.5]{brown2017contact}.

Analogous to the knife-mesh contact dynamics, we simulate contact forces between the object mesh and the ground that it rests on via the same penalty-based contact model. Here, the forces are computed between mesh vertices, represented by spheres as collision geometry, and the ground represented by a half space. To prevent the object from sliding off the table during the cut, as in~\cite{jamdagni2019robotic}, we apply boundary conditions that fix mesh vertices in place when they fulfill both the following conditions: (1) touching the ground and (2) being located \SI{1}{\cm} away from the cutting plane.

\subsection{Damage Mechanics}
\label{sec:damage}

Physically, damage refers to a macroscopic reduction in stiffness or strength of a material caused by the formation and growth of microscopic defects (e.g., voids and microcracks). For fruits and vegetables, which often have limited plastic deformation regimes before failure, damage can be approximated by a reduction in the elastic modulus of the material, or in a discrete mesh-based formulation, in the components of the stiffness matrix.

As follows, our model for damage mechanics leverages the springs that have been introduced in the final step in \autoref{sec:preprocess}. 
As the knife applies force to the cutting interface, the stiffnesses of the springs that are in contact with the knife, and hence receive knife contact forces, are linearly decreased (see a visualization of this progressive weakening as the knife slides down the cutting interface in \autoref{fig:cut_spring_evolution}):
\begin{align}
\label{eq:loosen-spring}
    k_e^\prime &= k_e - \gamma \, \|\mathbf{f}_{\text{knife}}\|,
\end{align}
where $\mathbf{f}_{\text{knife}}$ is the force the knife applies on the spring (i.e., the contact force that is computed between the knife and the edge), and $\gamma\in[0,1]$ is a coefficient that controls the ``weakness'' of the spring (i.e., how easily the material weakens and separates as the knife applies force to it).

\begin{figure*}[ht!]
    \centering
    \renewcommand{\figheight}[0]{5cm}
    \begin{subfigure}[b]{0.3\textwidth}
    \centering BayesSim\\
    \includegraphics[width=\linewidth]{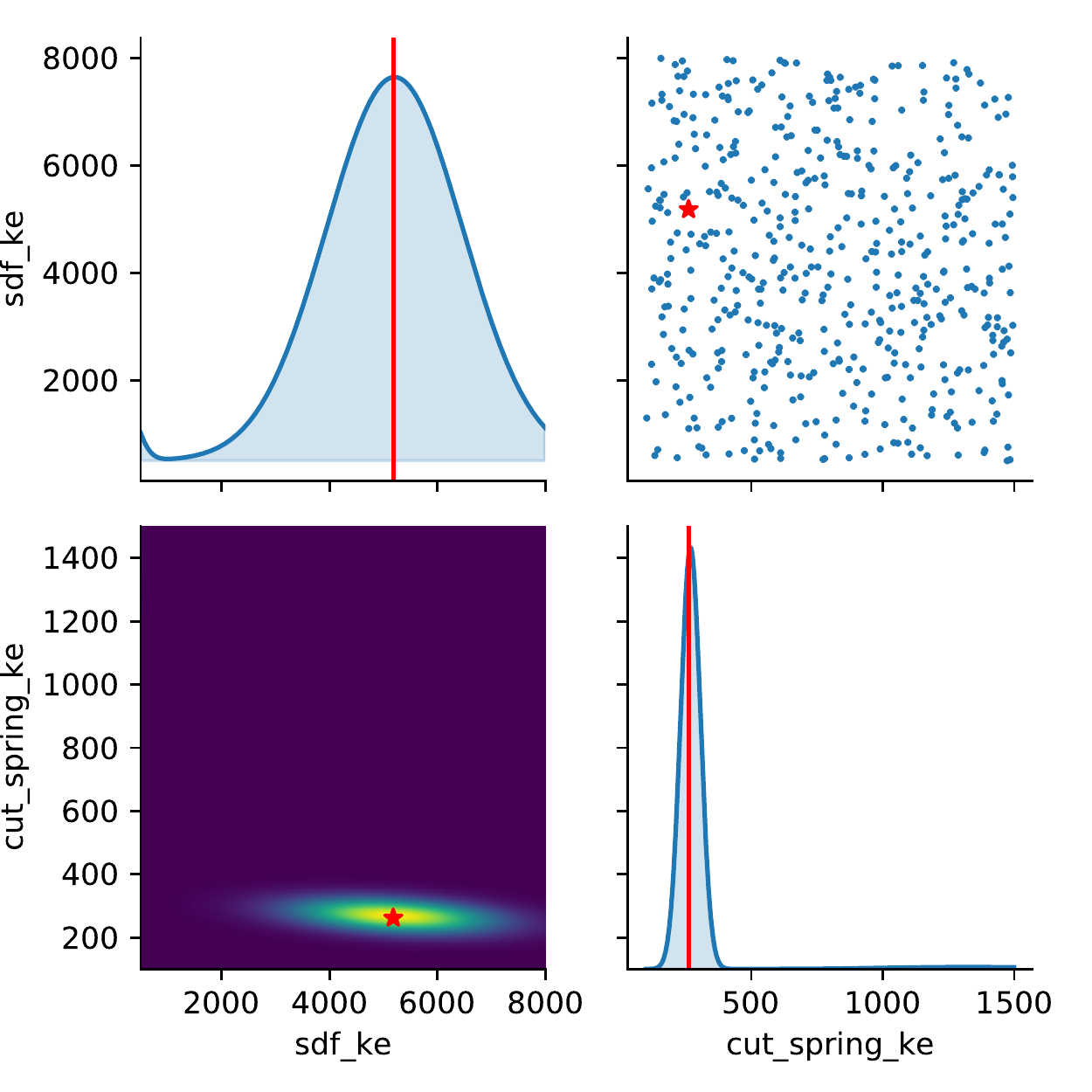}
    \end{subfigure}
    \hfill
    \begin{subfigure}[b]{0.3\textwidth}
    \centering SGLD\\
    \includegraphics[width=\linewidth]{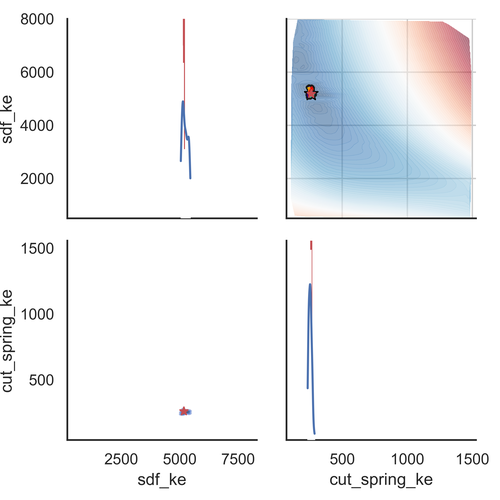}
    \end{subfigure}
    \hfill
    \begin{subfigure}[b]{0.3\textwidth}
    \centering HMC\\
    \includegraphics[width=\linewidth]{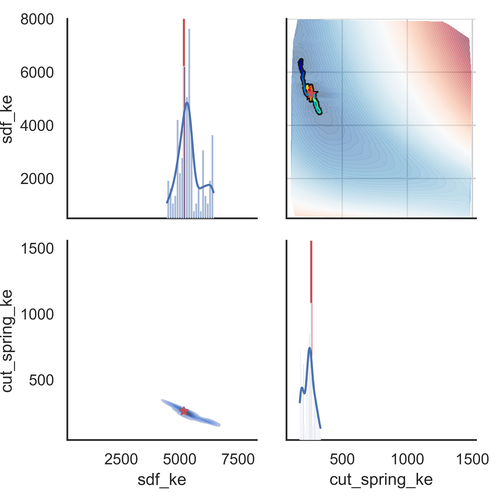}
    \end{subfigure}
    \caption{Parameter posterior inferred by BayesSim (left), Stochastic Gradient Langevin Dynamics (SGLD) after 90 burn-in iterations (center), and Hamiltonian Monte Carlo (HMC) after 50 burn-in iterations (right) for the two-dimensional parameter inference experiment from \autoref{sec:exp-infer-self}, in which cutting spring stiffness \texttt{sdf\_ke} and knife contact force stiffness \texttt{sdf\_ke} need to be estimated. The diagonals of each figure show the marginal densities for the two estimated parameters. The bottom-left sections show a heatmap of the joint distribution. The top-right scatter plot of the BayesSim figure shows the sampled parameters from the training dataset (blue) and the ground-truth (red). The top-right plot sections for SGLD and HMC visualize the loss landscape as a heatmap (where blue means smaller error than red), with the Markov Chain depicted by a colored line. Red lines and stars indicate ground-truth parameter values. SGLD and HMC leverage the gradients of our differentiable simulator and show a much sharper posterior distribution than BayesSim, which uses a dataset of 500 simulated force profiles.}
    \label{fig:actual-2d}
\end{figure*}

\begin{algorithm*}
\caption{Simulation Loop in \methodname}
\label{alg:overview}
\begin{algorithmic}[1]
\For{$i = 1 \dots \operatorname{time steps}$}
    \State Compute gravity and external contact forces $\mathbf{f}_{\text{ext}}$ between mesh and cutting board (half space).
    \State Compute elastic forces $\mathbf{f}_{\text{elastic}}$ following the constitutive model in \autoref{sec:fem}.
    \State Compute knife contact forces $\mathbf{f}_{\text{knife}}$ as described in \autoref{sec:contact}.
    \State Update cutting spring stiffness $k_e^\prime = k_e - \gamma \, \|\mathbf{f}_{\text{knife}}\|$.
    \State Compute cutting spring forces $\mathbf{f}_{\text{spring}}$.
    \State Semi-implicit Euler integration of mesh vertices: 
    \State \hspace{1em} $\mathbf{v}^{t+1} \leftarrow \mathbf{v}^t + \Delta t\textbf{M}^{-1}(\mathbf{f}_{\text{knife}} + \mathbf{f}_{\text{spring}} + \mathbf{f}_{\text{elastic}} + \mathbf{f}_{\text{ext}})$ 
    \State \hspace{1em} $\mathbf{x}^{t+1} \leftarrow \mathbf{x}^t + \Delta t\mathbf{v}^{t+1}$ 
    \State Euler integration of knife velocity (prescribed velocity trajectory).
\EndFor
\end{algorithmic}
\end{algorithm*}

We use a simulation time step of $\Delta t = 10^{-5}\si{\second}$ across all our experiments to accommodate the simulation of and stiff materials (such as apples or potatoes) and centimeter-scale meshes obtained by laser scans of real biomaterials. Our full simulation loop is described in Algorithm~\ref{alg:overview}.

\begin{figure*}
    \centering  %
    \renewcommand{\figheight}[0]{3.8cm}
    \includegraphics[height=\figheight,valign=t]{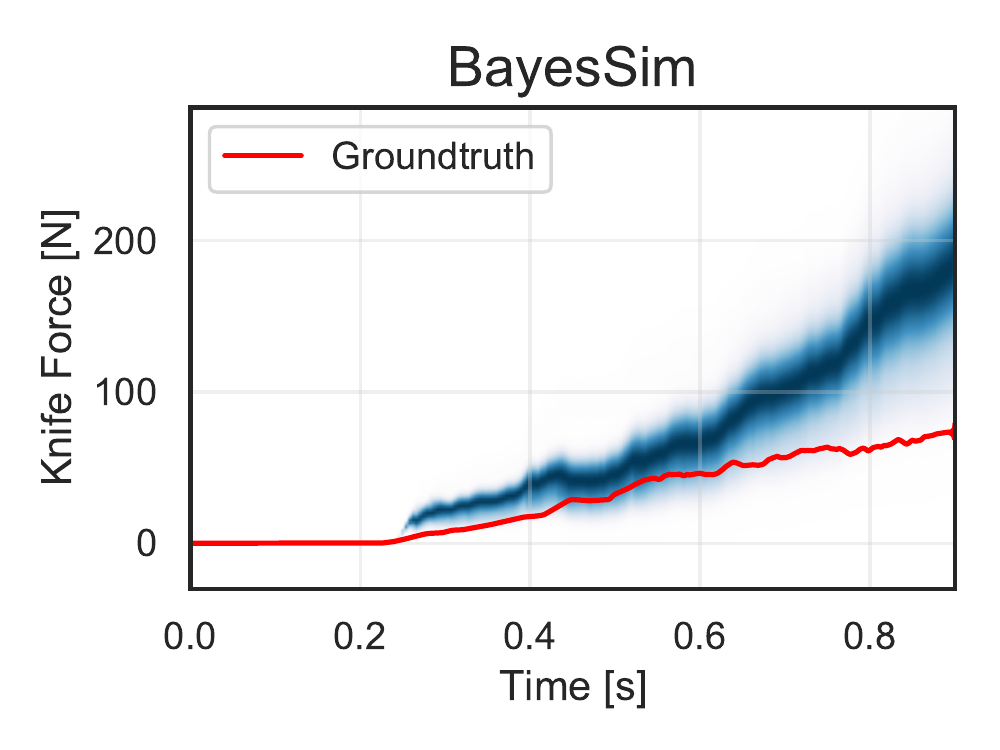}
    \includegraphics[height=\figheight,valign=t]{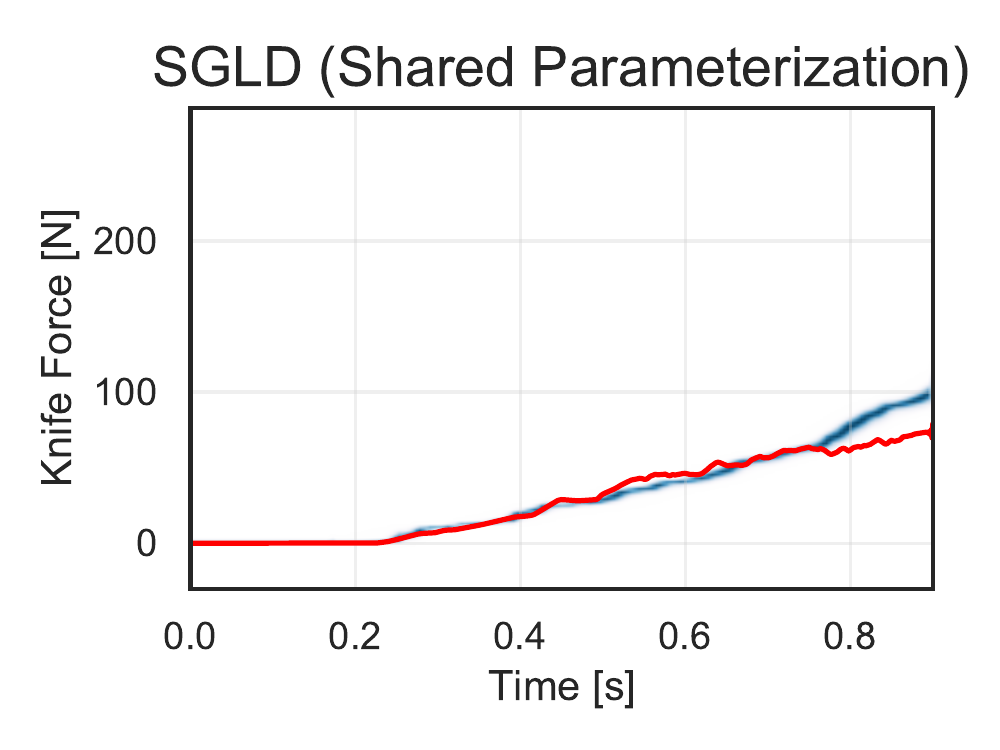}
    \includegraphics[height=\figheight,valign=t]{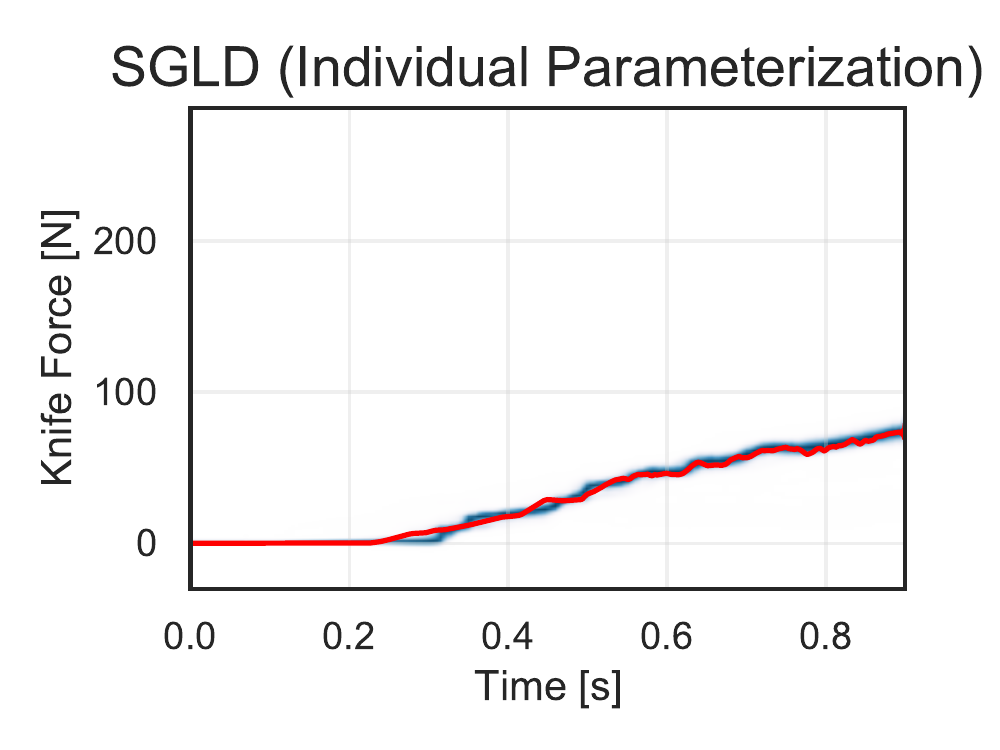}
    \caption{Kernel density estimation from 20 knife force trajectories rolled out by sampling from the posterior found by BayesSim (left) and SGLD applied in our differentiable simulator with shared (center) and individual parameterization (right). Shown in red is the ground-truth trajectory from a commercial simulation of cutting an apple with a hemispherical shape. Areas of higher density are shaded in dark blue.}
    \label{fig:ansys-apple-trajectory-density}
\end{figure*}

\section{Simulation Parameter Inference}
\label{sec:inference}

\methodname supports two modes of parameterizing the simulator: the cutting spring parameters shown in~Appendix~\autoref{tab:parameters} can be shared across all springs, or tuned individually per spring. The shared parameterization is low-dimensional since only one scalar per parameter type needs to be inferred. In contrast, when the cutting springs are individually parameterized, hundreds of variables need to be tuned.

While it is often possible to hand-tune parameters of low-dimensional, phenomenological models, the task of identifying the simulation parameters that can enable close prediction of real-world measurements is daunting, particularly for complex models such as ours with individually tuned spring parameters. To tackle the simulation calibration problem, we leverage automatic differentiation and GPU acceleration to efficiently compute gradients for all the parameters of our simulator. This allows us to use optimization techniques, such as stochastic gradient descent, to directly compute point estimates for the parameters. Moreover, we can leverage modern Bayesian inference methods and estimate posterior distributions for the high-dimensional parameter set given physical observations, such as the force profile of the knife while cutting real foodstuffs. The result is a simulator with the capacity to identify its own uncertainty about the physical world, leading to more robust simulations. We follow a conventional Bayesian approach and define $p(\theta)$ as the prior distribution over simulation parameters (see Appendix~\autoref{tab:parameters}) which, in our experiments, is a uniform distribution, 
and $p(\mathbf{\phi}^r \mid \theta)$
as the likelihood function given by
$p(\mathbf{\phi}^r\mid\theta)=\exp\{- \| \mathbf{\phi}_{\theta}^s - \mathbf{\phi}^r\|_L\}$,
where $\mathbf{\phi}^r$ corresponds to real trajectory observations such as knife forces, positions and velocities. $\mathbf{\phi}_{\theta}^s$ is the equivalent simulated trajectory, and $\| \cdot \|_L$ is the $L$-norm. In our experiments we use the $L1$, which we determined the most effective (see Appendix \ref{sec:loss_functions}).
With both prior and likelihood functions, we can compute the posterior as
$p(\theta\mid\mathbf{\phi}^r) = \frac{1}{Z} p(\mathbf{\phi}^r\mid\theta) p(\theta)$,
where $Z$ is a normalizing constant, also known as the marginal likelihood. 

In the next sections, we describe three techniques for simulator parameter inference. We start with a gradient-based approach that produces point estimates, followed by stochastic gradient Langevin dynamics (SGLD), a popular gradient-based Markov chain Monte Carlo technique that approximates the posterior as a set of particles. Finally, we describe the likelihood-free inference technique known as BayesSim, which does not make use of gradients and is used as a baseline, demonstrating the value of differentiable simulation.

\subsection{Gradient-Based Optimization}
\label{sec:baseline-adam}
As a baseline for probabilistic parameter inference, we present a solution based on stochastic gradient descent using the popular Adaptive Moment Estimation (Adam) optimizer~\cite{kingma2014adam} (see Appendix~\ref{sec:adam}), a first-order method that scales the parameter gradients with respect to their running averages and variances. Unlike the Monte Carlo posterior approximation obtained by SGLD, Adam will find a locally optimal point estimate to the parameters that minimizes the expected loss.
We define the loss as $l(\theta) = \log p(\mathbf{\phi}^r\mid\theta)$
and compute gradients with respect to the simulation parameters $\theta$ that minimize the loss between a real trajectory $\mathbf{\phi}^r$ and simulated trajectories $\mathbf{\phi}^s$.

\subsection{Stochastic Gradient Langevin Dynamics}
\label{sec:sgld}

A popular alternative to Adam for probabilistic inference is the stochastic gradient Langevin dynamics (SGLD) method~\cite{welling2011sgld} (Algorithm~\ref{alg:sgld}). SGLD combines the benefits of having access to parameter gradients with well-established sampling-based methods for probabilistic inference to significantly scale the parameter set to high dimensions at a tractable computational cost. The method can be seen as an iterative stochastic gradient optimization approach with the addition of Gaussian noise which is scaled by a preconditioner factor at every iteration. Given a sequence of trajectories generated by our simulator and a sequence of observed trajectories, $\Phi=\{\phi_i^s, \phi_i^r\}_{i=1}^N$, we can write the posterior distribution as 
\begin{align}
p(\theta\mid\Phi)\propto p(\theta)\prod_{i=1}^N p(\phi_i^r\mid\theta).
\end{align}

SGLD takes the energy function of the posterior denoted by $U(\theta) = -\sum_{i=1}^N \frac{1}{N} \log p(\Phi\mid\theta) - \log p(\theta)$ and samples from the posterior using the following rule:
\begin{align}
    \theta_{t+1} &= \theta_t - \frac{\alpha}{2} A(\theta_t) \nabla U(\theta_t) + \eta_t,
\end{align}
where $\eta_t\sim\mathcal{N}(0,A(\theta_t)\alpha) $, $\alpha$ is the learning rate, and $A$ is a preconditioner. After an initial burn-in phase necessary for the Markov chain to converge, $m$ samples can be stored to recover an approximate posterior given by $ p(\theta \mid \Phi) \approx \frac{1}{m} \sum_{i=1}^m \delta_{\theta_i}(\theta) $, where $\delta_{\theta_i}(x)$ is the Dirac delta function which is non-zero whenever $x = \theta_i$. 

Critical to SGLD's performance is the choice of an appropriate preconditioner. In this work we follow the extension proposed in~\cite{Li2016pSGLD} that computes the preconditioner as an approximation of the Fisher information matrix of the posterior distribution given by $A$. This approximation can then be sequentially updated using the gradient of the energy function. This is a similar process as the popular RMSProp~\cite{Tieleman2012rmsprop}. Specifically, the preconditioner $A\in\mathbb{R}^{m\times m}$ and momentum $V\in\mathbb{R}^m$ are updated as 
\begin{align}
V(\theta_t) &= \beta V(\theta_{t-1}) + (1-\beta) \nabla U(\theta_t)\odot \nabla U(\theta_t), \\
A(\theta_t) &= \operatorname{diag}\left(1 \oslash\left(\epsilon+\sqrt{V(\theta_t)}\right)\right),
\end{align}
where $\epsilon>0$ is a small diagonal bias (we choose $\epsilon=10^{-8}$) added to the preconditioner to prevent it from degenerating, and $\beta$ (which we set to $0.95$) is the exponential decay rate of the preconditioner. $\odot$ and $\oslash$ are element-wise multiplication and division operators.

\subsection{Likelihood-free Inference via BayesSim}
\label{sec:bayessim}

BayesSim is a likelihood-free technique to estimate the parameters of
a derivative-free simulator that implements a forward dynamics model which is used to generate trajectories for the given simulation parameters. The technique consists of learning a conditional density $ q(\theta\mid\mathbf{\phi}) $ (which BayesSim represents as a Gaussian mixture model), where
$\theta$ are the simulation parameters, and $\mathbf{\phi}$ are trajectories
or summary statistics of trajectories. For details about our particular BayesSim implementation, please see Appendix~\ref{sec:bayessim-extra}.

\section{Experiments}
\label{sec:experiments}

To evaluate \methodname, we first use synthetic data from our own simulator to evaluate probabilistic inference algorithms. Next, we identify the simulation parameters (Appendix \autoref{tab:parameters}) to closely match an industry-standard, high-fidelity simulation where we have access to the nodal forces and precise motion of the vertices resulting from the knife contact. Finally, we leverage real-world experimental data of measured knife force profiles to evaluate our sim2real transfer. 

\subsection{Parameter Inference from Synthetic Data}
\label{sec:exp-infer-self}

In our first experiment, we investigate how accurate the estimated posterior is, given synthetic force profiles from known simulation parameters. We create a dataset of 500 knife force trajectories by varying two of the parameters using the shared parameterization (\autoref{sec:inference}) from our proposed simulator as training domain. We choose such low dimensionality to ensure that we can obtain enough training data to sample from the uniform prior distribution which is crucial to the performance of likelihood-free methods, such as BayesSim.

The two parameters to be estimated are \texttt{sdf\_ke} (ranging from 500 to 8000), the stiffness of the contact model at the mesh surface, and \texttt{cut\_spring\_ke} (ranging from 100 to 1500), the stiffness of the cutting springs (cf. \autoref{sec:damage}) at the beginning of the simulation.

As shown in \autoref{fig:actual-2d}, BayesSim captures the posterior (see heatmap in the lower-left corner of the left subfigure) around the ground-truth parameters of \texttt{sdf\_ke} = 5100 and \texttt{cut\_spring\_ke} = 200. SGLD (center), however, yields a significantly sharper density estimate around the true values within 90 burn-in iterations, while Hamiltonian Monte-Carlo (HMC), another gradient-based Bayesian estimation algorithm, finds a slightly wider posterior within approximately 50 iterations.

\subsection{Parameter Inference from High-fidelity Simulator}
\label{sec:exp-ansys}

In our first set of experiments where the training data does not stem from our own model, we simulate cutting trajectories using a commercial, explicit dynamics simulator as a ground-truth source. Details on the simulation setup are in Appendix~\ref{sec:ansys-setup}. For each simulation, knife force and nodal motion trajectories were extracted. Each simulation was executed across 4 CPUs and took an average of \SI{1941}{\minute} ($>$\SI{32}{\hour}) to complete. For comparison, our simulator produces a cutting trajectory of \SI{1}{\second} duration (with a \SI{1e-5}{\second} simulation time step) within \SI{30}{\second} on an NVIDIA RTX 2080 GPU, and the gradient of the cost function (likelihood) within \SI{90}{\second}.

We estimate the following parameters $\theta$ in this experiment:
(1) \texttt{sdf\_ke}, 
(2) \texttt{sdf\_kd}, 
(3) \texttt{sdf\_kf},
(4) \texttt{sdf\_mu},
(5) \texttt{cut\_spring\_ke},
(6) \texttt{cut\_spring\_softness}, and
(7) \texttt{initial\_y} (for an explanation see \autoref{tab:parameters} in the appendix).

First, we train BayesSim in an iterative procedure where the currently estimated posterior over the simulation parameters is sampled to roll out a new set of trajectories that are added to the training dataset. Starting from 500 trajectories, we repeatedly sample 20 new parameters and refit BayesSim's mixture density network (MDN) to the updated training dataset. After 100 such iterations (i.e., 2000 additional roll-outs), we obtain the posterior shown in appendix \autoref{fig:ansys-apple-bayessim-posterior}.

For comparison, we train SGLD in \methodname{} for 300 iterations with the same parameterization as used for BayesSim, i.e., the parameters are represented by scalars that are shared across all cutting springs. As shown in \autoref{fig:ansys-apple-trajectory-density}, trajectories sampled from the estimated posterior of SGLD result in a significantly closer fit (center) to the ground-truth knife force profile compared to BayesSim (left). The average mean absolute error (MAE) of the roll-outs with \SI{4.714}{\newton} significantly outperforms BayesSim's average MAE of \SI{26.860}{\newton}. The gradient-based estimation method achieves such an outcome with less training data, as it only took 300 trajectory simulations, compared to the 2500 trajectories in total that served as the training dataset for BayesSim. If we allow the parameters to be optimized individually for each cutting spring (resulting in 1737 parameters in total), the resulting simulation becomes even closer (right), with an average MAE of \SI{3.075}{\newton}.

\begin{figure}
    \centering
    \includegraphics[width=\columnwidth]{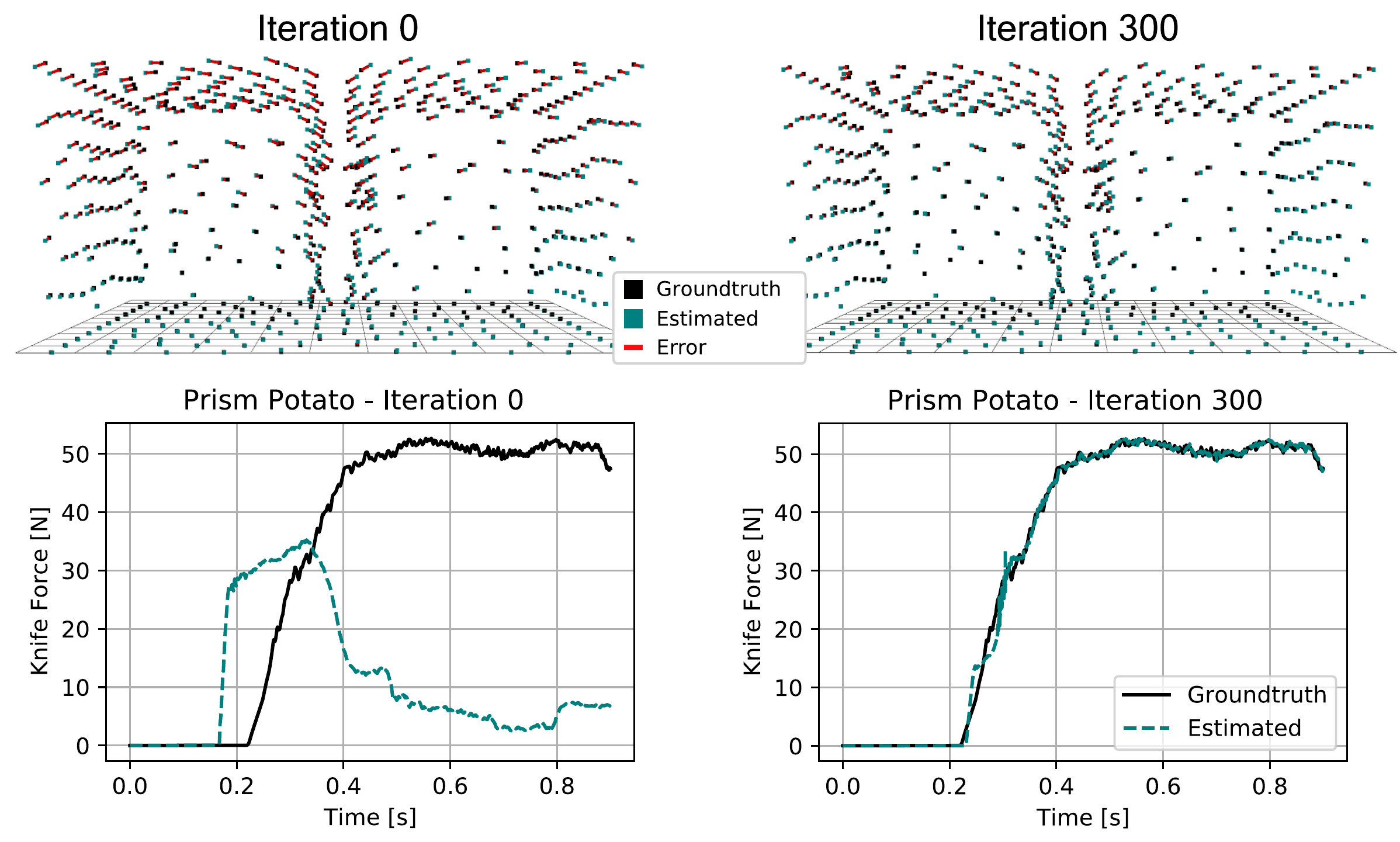}
    \caption{Results from simulation parameter optimization given the positions of the vertices (top row) and the knife force (bottom row) with a commercial simulation as ground-truth. \textit{Left:} before optimization, the vertices (top) at the last time step (\SI{0.9}{\second}) of the trajectory are visibly distinct between our simulation (blue) and the ground-truth (black), as shown by the red lines indicating the vertex difference. \textit{Right:} after 300 steps with the Adam optimizer, the vertices (top), as well as the knife force profile (bottom), are closer after the parameter inference.}
    \label{fig:ansys-node-motion}
\end{figure}

Based on the commercial ground-truth simulation, we collect additional object nodal displacement field trajectories in addition to the knife force profiles, which allows us to leverage another reference signal to calibrate \methodname. At 18 reference time steps within the trajectory roll-out, we compute the $L2$ error between the positions of the vertices in the ground-truth and our simulator, and add it to the overall cost (which previously only consisted of the $L1$ norm over knife force difference) with a tuned weighting factor. By optimizing the aforementioned parameters individually with the new cost function through Adam, after 300 iterations, we arrive at a simulation that not only closely matches the knife force profile from the ground-truth simulation, but also has a significantly reduced gap in the nodal motions between the two simulators (see \autoref{fig:ansys-node-motion}). Before the optimization, the mean Euclidean distance between the simulated nodes and the ground-truth nodal positions is \SI{1.554}{\mm} at the last time step ($t = \SI{0.9}{\second}$). The low initial error over the nodal motion is explained by the fact that the material properties have been set to already match a potato (\autoref{tab:materials}), leaving the cutting-related spring parameters as the only remaining variables to be estimated. After 300 iterations, this error reduced to \SI{1.289}{\mm}, while the knife force profile MAE decreased from \SI{28.366}{\newton} to \SI{0.549}{\newton}.

\subsection{Parameter Inference from Real World Measurements}
\label{sec:exp-infer-real}

\begin{figure}[t!]
    \centering
    \includegraphics[width=\columnwidth]{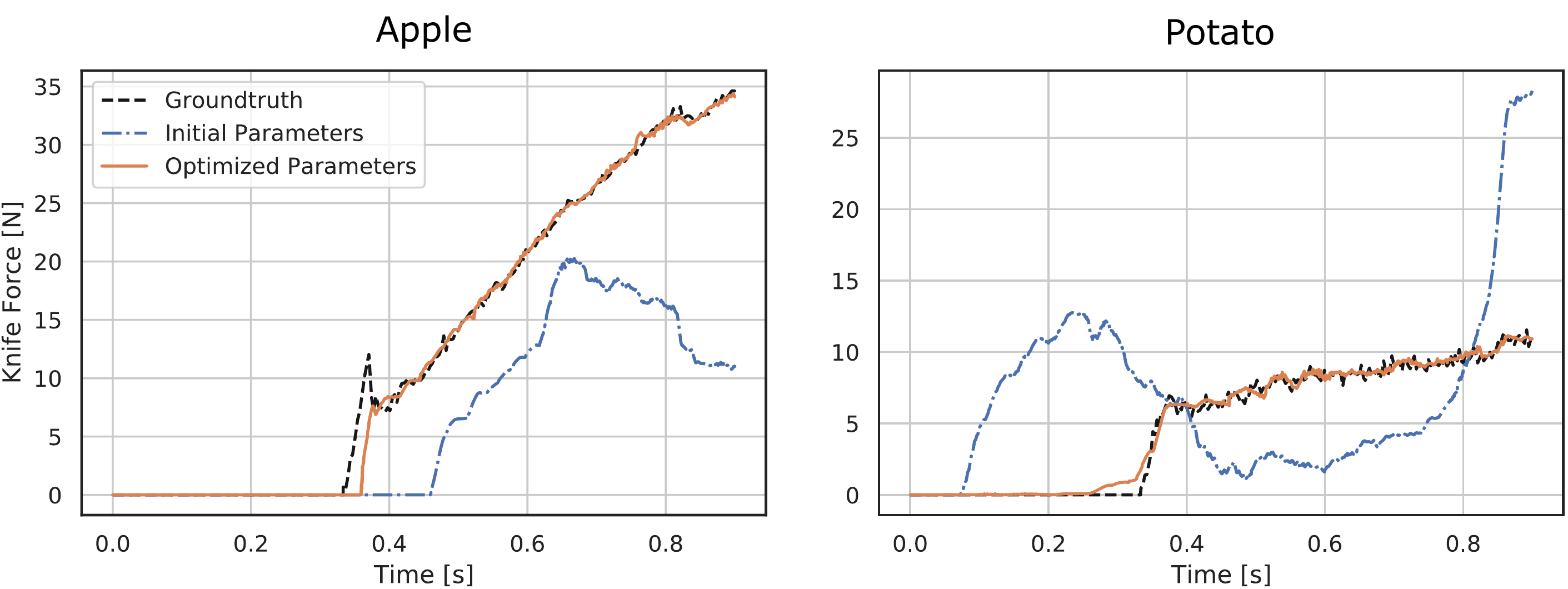}
    \caption{Results from optimizing simulation parameters in \methodname given real-world knife force profiles from vertical cutting of an apple (left) and a potato (right).}
    \label{fig:accuracy-real}
\end{figure}

In this experiment, we calibrate \methodname to real-world cutting trajectories.
The real-world dataset provided by~\cite{jamdagni2019robotic} contains 3D meshes created from laser scans of the actual objects being cut, as well as knife force profiles measured from a force sensor mounted between a robot end-effector and a knife. The knife dimensions are given in~\autoref{tab:parameters} and explained in \autoref{fig:knife-params}. The triangular surface meshes of the foodstuffs are discretized to tetrahedral meshes via the TetWild~\citep{hu2018tetwild} meshing library. After optimizing the simulation parameters with Adam for 300 iterations, the simulator closely matches the force profile of a knife cutting an apple (\autoref{fig:accuracy-real} left) with an MAE of \SI{0.253}{\newton}, and for a potato cutting action (right) achieves an MAE of \SI{0.379}{\newton}. We note that the optimization is stable without requiring restarts from different parameter configurations, and we found even relatively poor parameter initialization (such as the one shown in Figure \ref{fig:accuracy-real}) results in highly accurate prediction after the calibration.

\subsection{Generalization}
\label{exp:generalization}

While our simulator is able to calibrate itself very closely to a variety of sources of ground-truth signals -- whether these are knife force profiles obtained by a real robot cutting foodstuffs, or motion recordings from the mesh vertices in a commercial simulator that is entirely different from ours (see Appendix~\ref{sec:ansys-setup}) -- the question of overfitting arises.
In the following, we investigate how well the identified simulation parameters transfer to test regimes that differ from the training conditions under which these parameters were optimized.

\subsubsection{Generalization to longer simulations}
\label{sec:exp-ansys-duration}

As shown in \autoref{tab:parameters}, \methodname allows various dynamics parameters to be defined individually for each spring, resulting in hundreds of degrees of freedom. The larger parameterization provides more opportunity to find closely matching solutions, but may be prone to overfitting to the reference trajectory in certain settings. One of the pathological cases occurs when the goal is to predict the knife force trajectory for a duration longer than the time window seen during training from the reference force profile. In the experiment shown in \autoref{fig:exp-ansys-duration}, we optimize the simulation parameters using Adam on the first \SI{0.4}{\second} section from the reference trajectory. Within that segment, the individual parameterization clearly outperforms the shared parameterization with a MAE of \SI{0.306}{\newton} versus \SI{0.384}{\newton}. However, when we test the estimated parameters on a simulation with a duration of \SI{0.9}{\second}, the shared parameterization achieves a closer fit with a \SI{4.802}{\newton} MAE, compared to the individual tuning with a \SI{5.921}{\newton} MAE. Intuitively, as the knife slices downward with constant velocity (at \SI{50}{\mm\per\second}), fewer cutting springs are affected by the contact dynamics during a shorter roll-out since the knife does not progress far enough to reach the cutting springs closer to the ground. Hence, their parameters' gradients were zero during the estimation. Tuning the same kind of parameters uniformly allows all cutting springs to have an improved fit over the initialization, even when their interaction with the knife only becomes apparent at a later time.

\subsubsection{Generalization to different knife velocities}
\label{sec:exp-ansys-velocity}

\begin{figure}
    \centering
    \includegraphics[width=0.9\linewidth]{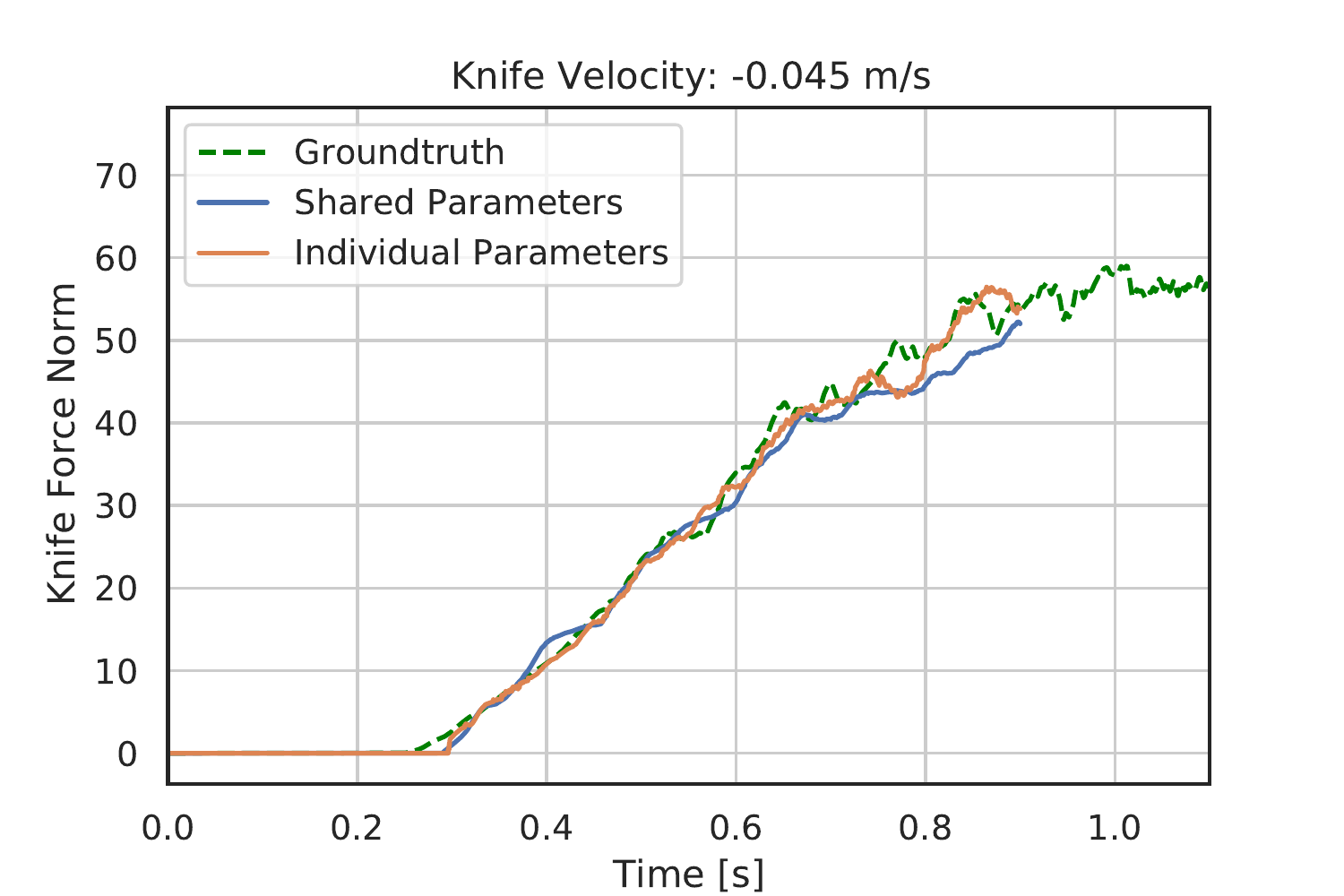}
    \caption{Knife force profiles for cutting a cylindrical mesh with cucumber material properties by simulating the parameters with $45\si{\mm\per\s}$ knife velocity downwards.
    The simulation parameters in the shared and individual parameterization have been inferred from a ground-truth trajectory with $50\si{\mm\per\s}$ knife velocity from a commercial solver (see \autoref{sec:exp-ansys-velocity}).}
    \label{fig:exp-ansys-velocity}
\end{figure}

We investigate how accurately \methodname predicts knife force profiles given the parameters that were inferred from a cutting trajectory with a knife downward velocity of \SI{50}{\mm\per\second}. At test time, we change the downward velocity to 35, 45, 55, and \SI{65}{\mm\per\second}. As shown in \autoref{fig:exp-ansys-velocity-apple} (and \autoref{fig:exp-ansys-velocity-cucumber} in the appendix), the individual parameterization significantly outperforms the shared parameterization in normalized mean absolute error (NMAE), i.e., the MAE between the ground-truth and estimated trajectory divided by the mean force of the ground-truth trajectory, achieving a four-fold more accurate result compared to the shared parameterization in many cases (see example knife force profile in \autoref{fig:exp-ansys-velocity}).

\begin{figure}
    \centering
    \includegraphics[width=0.8\columnwidth]{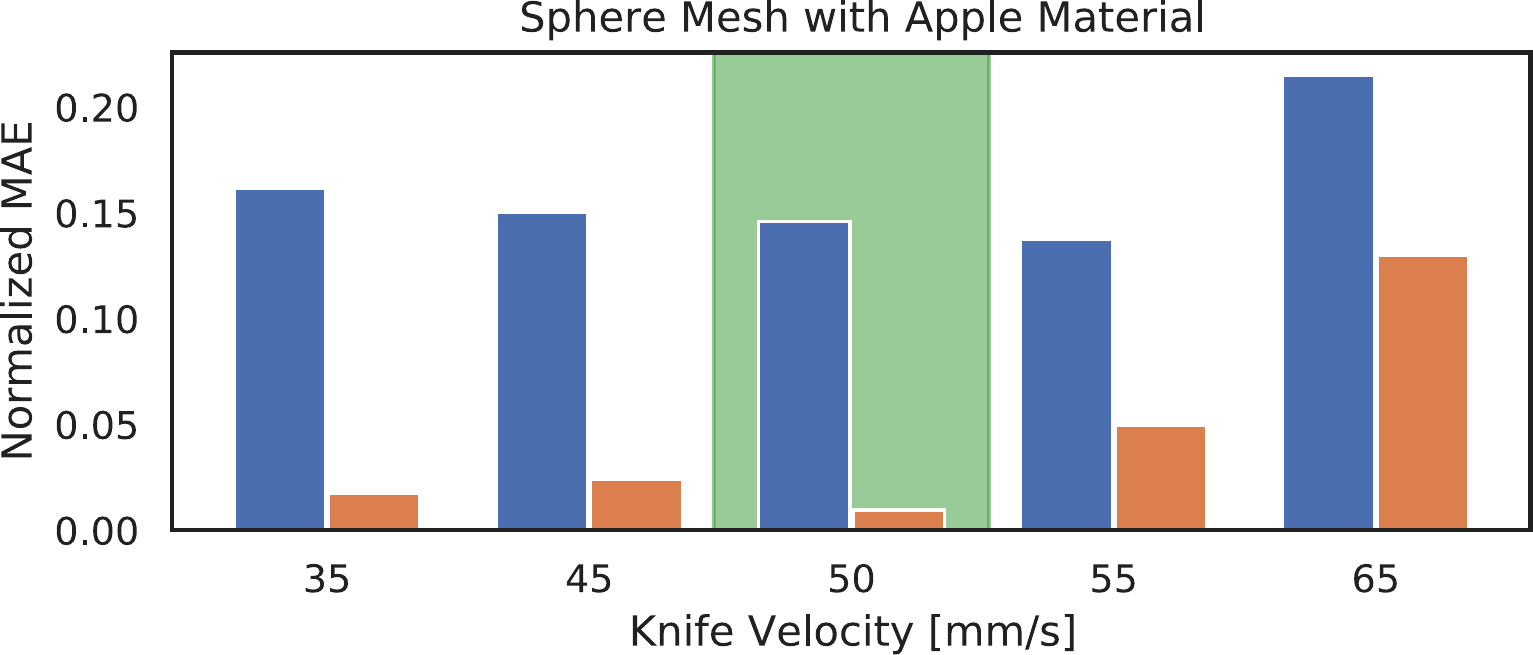}
    \caption{Velocity generalization results for a sphere shape with apple material properties given ground-truth simulations with different vertical knife velocities from a commercial simulator. Two versions of \methodname were calibrated: by sharing the parameters across all cutting springs (blue) and by optimizing each value individually (orange), given a ground-truth trajectory with the knife sliding down at \SI{50}{\mm\per\second} speed (highlighted in green). The normalized mean absolute error (MAE) is evaluated against the ground-truth by rolling out the estimated parameters for the given knife velocity.}
    \label{fig:exp-ansys-velocity-apple}
\end{figure}

\subsubsection{Generalization to different geometries}
\label{sec:exp-generalize-geometry}

\begin{figure}[t!]
    \centering
    \renewcommand{\figheight}[0]{3cm}
    \includegraphics[width=\columnwidth]{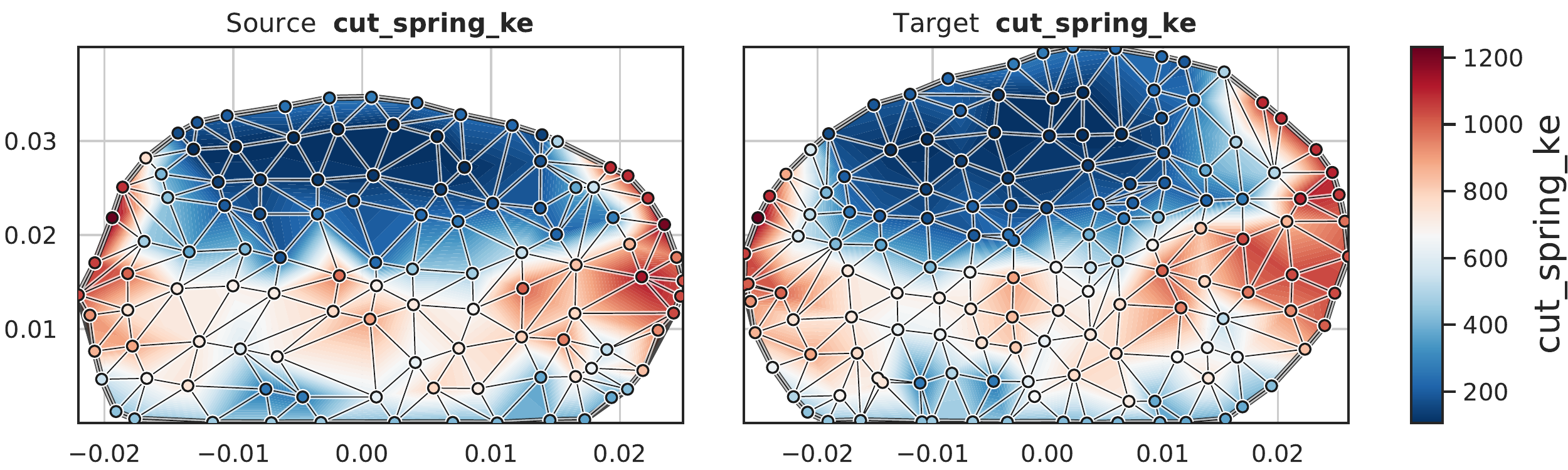}\\
    \includegraphics[width=\columnwidth]{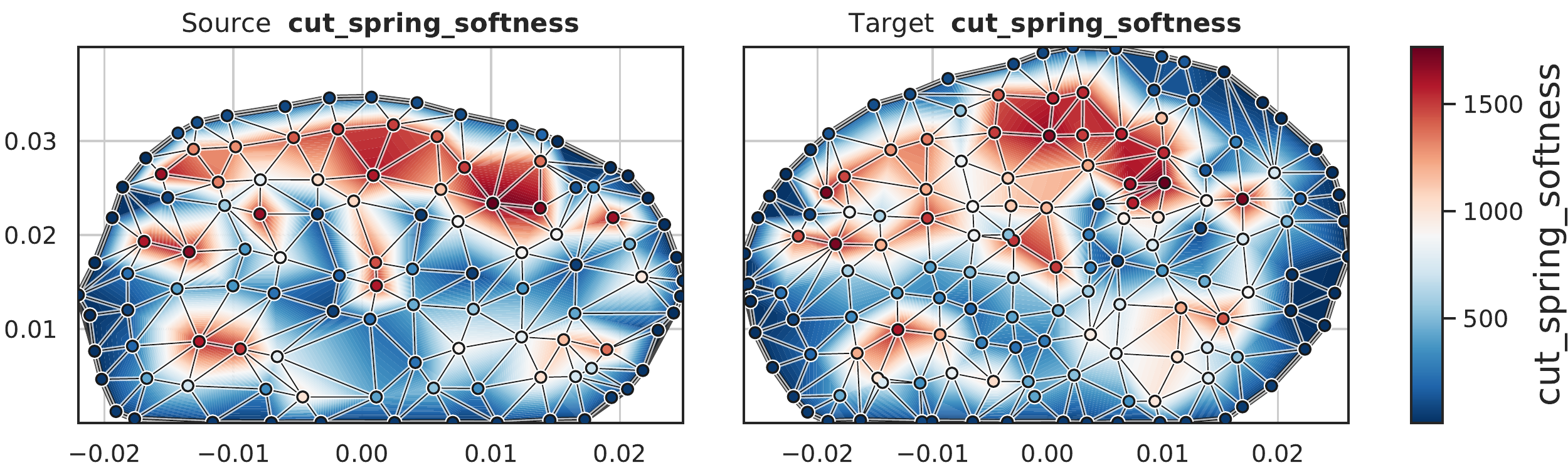}\\
    \caption{Transfer of cutting parameters between two different potato meshes from a real-world cutting dataset. For the model in the left column, the cutting spring parameters have been optimized individually for each cutting spring given a single trajectory of the knife force (only two of the parameter types are shown in both rows). These parameters have been transferred to the mesh on the right column via Optimal Transport with the Earth Mover's Distance (EMD) objective (more details for this mesh transfer example are shown in \autoref{fig:ybj-potato-optimal-transport}).}
    \label{fig:ybj-potato-generalization}
\end{figure}

Cutting the same type of biomaterial  can lead to drastically different knife force profiles, even when all the variables that influence the motion of the knife remain the same~\citep{jamdagni2019robotic}. 
This can be caused by different geometries even within the same object class, as no two fruits or vegetables of the same type are identical. Since the mesh topologies can differ significantly between the different geometric shapes that a foodstuff may have, a direct mapping between the virtual nodes (respective cutting springs) is not possible, which would allow the transfer of the individual parameters. Instead, we propose a weighted mapping of a combination of cutting spring parameters from the source mesh that are in proximity to the cutting springs of the target mesh.
We developed an optimal transport~\citep{peyre2019ot} method that receives as inputs the cutting spring coordinates in 2D (obtained by the mesh preprocessing step from \autoref{sec:preprocess}) at the cutting interface (shown in \autoref{fig:ybj-potato-generalization}) from a source domain, and a different set of 2D coordinates for the target domain. By minimizing the Earth Mover's Distance (EMD)~\citep{Rubner2000EMD} between the cutting spring vertices of the two meshes\footnote{Our implementation uses the Python Optimal Transport library~\citep{flamary2017pot}}, we find a weight matrix that allows us to compute spring parameters for the target domain as a weighted combination of the parameters from the source mesh. For more details, see \autoref{sec:optimal-transport}. Similar to the velocity generalization experiments, we observe a significantly improved NMAE performance (\autoref{tab:mesh-generalization}) in most cases when the cutting spring parameters are tuned individually (``NMAE OT'') in contrast to duplicating average of each spring parameter (``NMAE Avg'') from the source domain across all locations in the target domain. 

\begin{table}[]
    \centering
    \resizebox{\columnwidth}{!}{%
    \begin{tabular}{lllrr}
        \toprule
        \bf Material &
        \bf Source Mesh &
        \bf Target Mesh &
        \bf NMAE OT &
        \bf NMAE Avg \\
        \midrule
        Potato & Real Potato 1 & Real Potato 2 & 0.948 & 5.635 \\
        Potato & Real Potato 2 & Real Potato 1 & 1.360 & 6.981 \\
        Apple & Real Apple 2 & Real Apple 3 & 6.844 & 1.330 \\
        Apple & Real Apple 3 & Real Apple 2 & 3.857 & 19.749 \\\addlinespace[.5em]
        Potato & Cylinder & Prism & 3.470 & 12.001 \\
        Potato & Prism & Cylinder & 0.933 & 1.983 \\
        Potato & Prism & Sphere & 7.867 & 15.841 \\
        Potato & Sphere & Prism & 35.347 & 16.812 \\\addlinespace[.5em]
        Apple & Cylinder & Sphere & 4.261 & 14.457 \\
        Apple & Sphere & Cylinder & 1.839 & 1.602 \\
        Apple & Prism & Sphere & 0.920 & 4.531 \\
        Apple & Sphere & Prism & 13.883 & 45.983 \\\addlinespace[.5em]
        Cucumber & Cylinder & Sphere & 66.672 & 71.102 \\
        Cucumber & Sphere & Cylinder & 4.239 & 0.484 \\
        Cucumber & Cylinder & Prism & 58.138 & 64.407 \\
        Cucumber & Prism & Cylinder & 1.046 & 1.855 \\
        \bottomrule\\
    \end{tabular}
    }
    \caption{Mesh generalization results when the parameters from the source domain are transferred to the target domain via Optimal Transport (OT), and by averaging the parameters across all cutting springs. The numbers show the normalized mean absolute error (NMAE), i.e., the MAE divided by the mean of the respective ground-truth knife force profile, to make the results comparable across different material properties and geometries.}
    \label{tab:mesh-generalization}
\end{table}

Overall, our generalization experiments have shown that, although we optimize hundreds of parameters involved in the cutting dynamics at highly localized places, such representation still generalizes between various conditions. The successful transport of these parameters between two topologically different meshes based on their spatial correspondences indicates that our simulation parameters implicitly encode material properties that generalize across mesh topologies.

\subsection{Controlling Knife Velocity}
\label{sec:exp-control}

In real-world applications, automated cutting of foodstuffs may need to meet multiple competing objectives. In particular, force may be minimized to prevent peripheral damage to the object, or ensure human safety, while the velocity may be maximized to reduce the required time. For cutting of food and biomaterials, humans have intuitively developed the strategy of minimizing cutting force by pressing the knife vertically and simultaneously slicing horizontally (i.e., a sawing motion) \citep{atkins2004cutting, zhou2006cut1, zhou2006cut2}. Such a cutting action can be intuitively understood as follows: by definition, the work applied by the knife to the object is equal to the cutting force integrated over displacement, and by conservation of energy, also equal to the fracture energy required to introduce a cut in the vertical plane. By simultaneously slicing horizontally, the distance traveled by the knife will be greater, reducing the cutting force for the required fracture energy.

\subsection{Knife Motion Trajectory Optimization}
\label{sec:mdmm}
We represent the knife velocity trajectory by $k$ equidistant keyframes in time. Three parameters are to be optimized per keyframe $i$ (refer to \autoref{fig:knife-params} for the coordinate frame orientation w.r.t. the knife):
\begin{itemize}
    \item $a_i$: the amplitude of the lateral (along $z$ axis) sinusoidal  velocity
    \item $b_i$: the frequency of the lateral sinusoidal velocity
    \item $c_i$: the vertical (along $y$ axis) velocity 
\end{itemize}

\begin{figure}
    \centering
    \includegraphics[width=0.9\columnwidth]{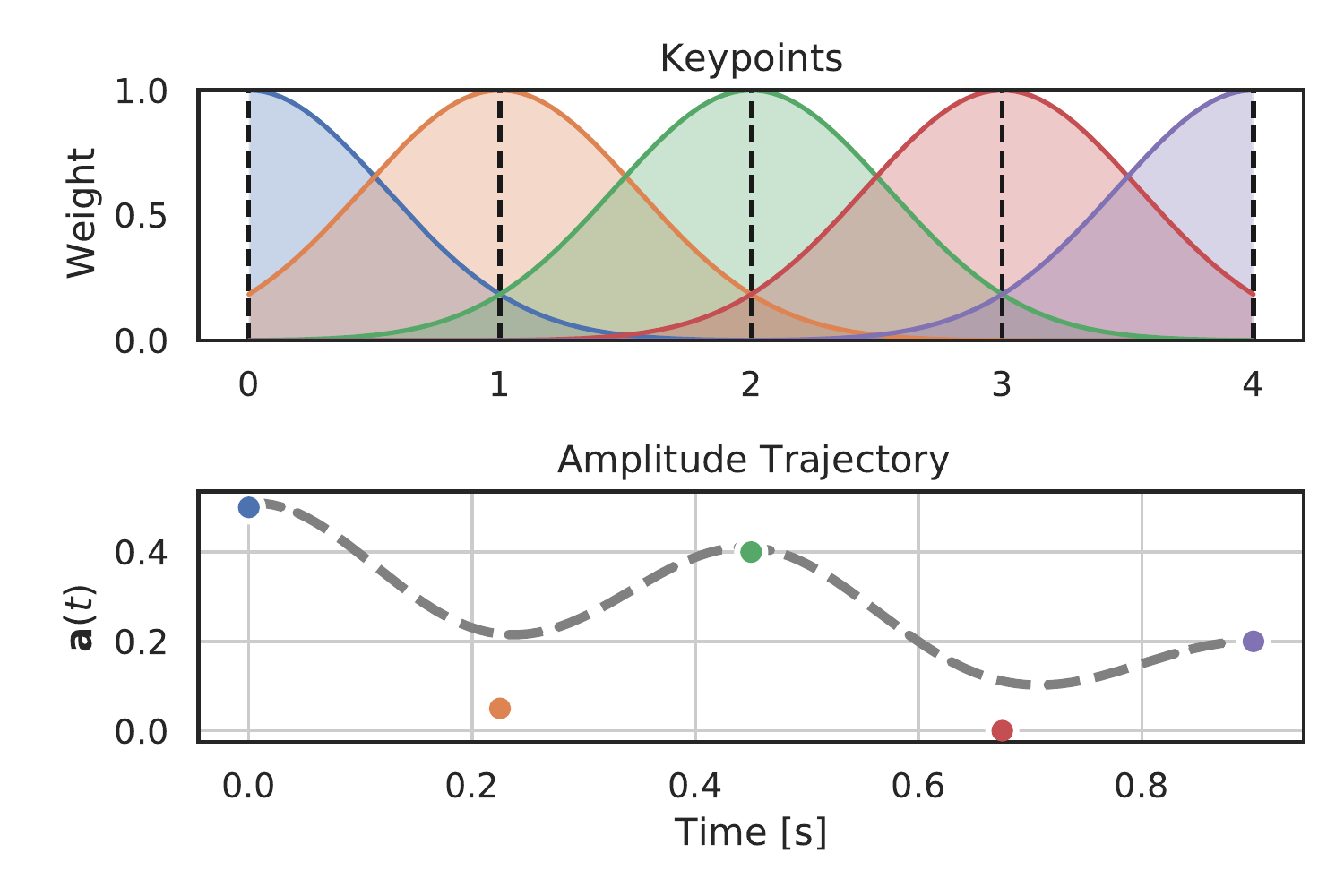}
    \caption{Visualization of five keypoints (top) evenly distributed in time with exemplary amplitude values $\mathbf{a}[i]$ per keyframe $i$ (color-matching dots in lower plot). The resulting continuous trajectory $\mathbf{a}(t)$ resulting from weighting the keyframes via the RBF kernel (\autoref{eq:rbf-weighting}) is shown as the dashed line at the bottom.}
    \label{fig:explain-slicing-trajectory}
\end{figure}

To allow for a smooth interpolation between the keyframes, and propagation of gradients from all trajectory parameters at every time step, we weight the contribution of all keyframe parameters on the entire trajectory via the radial basis function (RBF) kernel. The kernel uses the squared norm of the difference between the current time $t$ difference and the predefined keyframe times to compute the weight contributions $\mathbf{w}\in\mathbb{R}^k$ of the keyframe parameters (see \autoref{fig:explain-slicing-trajectory}):
\begin{align}
\label{eq:rbf-weighting}
    \mathbf{w}(t) = \exp \left( -\frac{\|t - w\|^2}{2\sigma^2} \right)
\end{align}

In effect, a nonzero contribution on the trajectory is maintained from all keyframes at all times, which eases gradient-based optimization.

The kernel width $\sigma$ controls how smoothed out the contributions of the keyframe parameters become. We found $\sigma = \sqrt{0.03}$ to be an appropriate setting (see illustration in \autoref{fig:explain-slicing-trajectory}), given that the duration of the cutting action we optimize for is \SI{0.9}{\second} using $k=5$ keyframes.

To compute the knife's horizontal (sideways) and vertical velocities $\dot{z}_{\text{knife}}(t)$ and $\dot{y}_{\text{knife}}(t)$ at time $t=[0..T]$, the keyframe parameters in vector form $\mathbf{a},\mathbf{b},\mathbf{c}\in\mathbb{R}^k$ are combined with the time-dependent weighting contribution from \autoref{eq:rbf-weighting}:
\begin{align}
\label{eq:cutting-motion}
    \dot{z}_{\text{knife}}(t) &= \mathbf{a}\cdot\mathbf{w}(t) \cos(\mathbf{b}\cdot\mathbf{w}(t) t) \\
    \dot{y}_{\text{knife}}(t) &= \mathbf{c}\cdot\mathbf{w}(t)
\end{align}

We minimize the mean knife force plus the vertical knife velocity integrated over the entire length of the trajectory. Thus, we penalize high actuation effort by the robot controlling the knife while maintaining a fast progression of the cutting process:
\begin{align}
    \label{eq:slicing-objective}
    \mathop{\operatorname{minimize}}_{\mathbf{u} = [\mathbf{a}, \mathbf{b}, \mathbf{c}]} \qquad \mathcal{L} &= \frac{1}{T} \int f(t, \mathbf{a}, \mathbf{b}, \mathbf{c}) + \dot{y}_{\text{knife}}(t) \, dt\\
    \label{eq:slicing-height-constraint}
    \text{s.t.} \qquad  y_{\text{knife}}(T) &= h_\text{end} \\
    \label{eq:slicing-lateral-constraint}
                 \vert z_{\text{knife}}(t) \vert &\leq \frac{1}{2} l_\text{knife},
\end{align}
where $k=5$ is the number of keyframes, $T=\SI{0.9}{\second}$ is the time at the end of the trajectory, $f$ is the simulation step that returns the knife force norm $\|\mathbf{f}_\text{knife}\|$ at time step $t$ given the trajectory parameter vectors $\mathbf{a}, \mathbf{b}, \mathbf{c}$, and
$l_\text{knife}=\SI{15}{\cm}$ is the blade length of the knife. We impose the hard constraint in \autoref{eq:slicing-lateral-constraint} to ensure that the knife does not move too far along $z$ where the blade of the knife ends (which would trivially minimize the knife force, although we assume the knife blade has infinite length in this experiment setup).

Constrained optimization problems are typically solved by converting them to unconstrained optimization problems through the introduction of Lagrange multipliers. However, the critical points to such Lagrangians often tend to be saddle points, which gradient-descent-style algorithms, such as Adam, will not converge to~\citep{platt1988mdmm}. To make the unconstrained objective amenable to gradient descent, following the modified differential method of multipliers (MDMM)~\citep{platt1988mdmm}, we introduce a penalty term for $\mathbf{u} = [\mathbf{a}, \mathbf{b}, \mathbf{c}]$:
$$
E_\text{penalty} = \frac{c}{2}(g(\mathbf{u}))^2.
$$
This term acts as an attractor to the energy function that we are optimizing for (where $c$ acts as a damping factor), where $g(\mathbf{u})$ is an equality constraint.

To include the inequality constraint in \autoref{eq:slicing-lateral-constraint}, we convert it to an equality constraint $g(\mathbf{x})$ by introducing slack variable $\gamma\in\mathbb{R}$ that becomes part of $\mathbf{u}$:
\begin{align}
   g(\mathbf{u}) &= \frac{1}{2} l_\text{knife} - \vert z_{\text{knife}}(t) \vert - \gamma^2.
\end{align}

The update rule for the trajectory parameters $\mathbf{u}$ is then
\begin{align}
   \mathbf{u}^\prime &= \mathbf{u} - \frac{\partial \mathcal{L}}{\partial \mathbf{u}} - \lambda \frac{\partial g}{\partial \mathbf{u}} - c g(\mathbf{u}) \frac{\partial g}{\partial \mathbf{u}} \\
   \lambda^\prime &= \lambda + g(\mathbf{u}).
\end{align}

Using gradient-based trajectory optimization, we observe that such an intuitive cutting strategy emerges. We define the cost function in \autoref{eq:slicing-objective} to penalize the mean knife force and inverse velocity, and we parameterize the knife trajectory via keyframes in time that define the downward velocity and sinusoidal time-varying horizontal velocity (a complete description is given in \autoref{sec:mdmm} of the appendix).

\begin{figure}[t!]
    \centering
    \includegraphics[width=0.48\columnwidth,trim=0.5cm 0 0.25cm 0,clip]{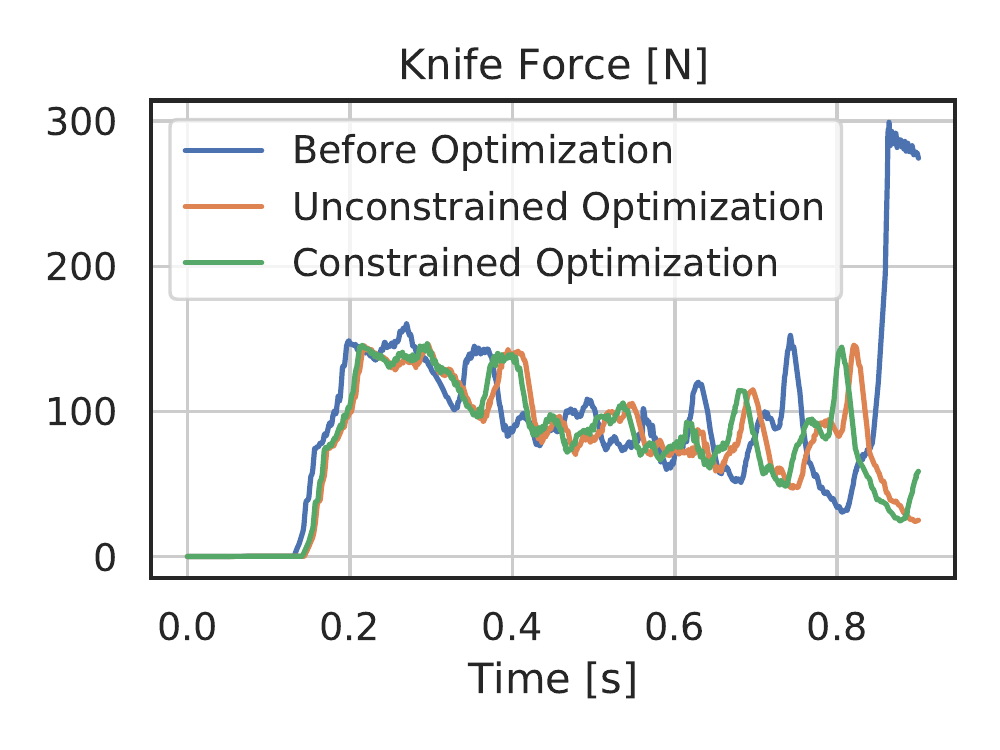}
    \hfill
    \includegraphics[width=0.48\columnwidth,trim=0.5cm 0 0.25cm 0,clip]{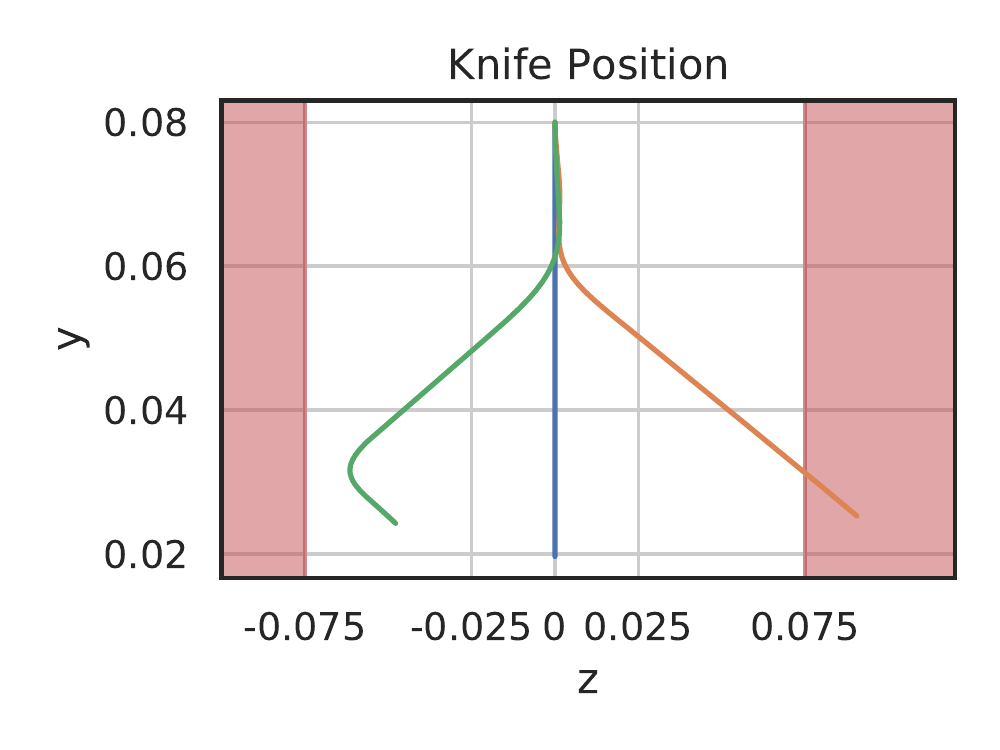}
    \caption{Results from the trajectory optimization experiment (\autoref{sec:exp-control}) on the cylinder mesh with cucumber material properties where the mean knife force is penalized and the overall downward velocity maximized. \textit{Left:} knife force profiles before the trajectory optimization (blue), after unconstrained (orange) and constrained (green) optimization. \textit{Right:} resulting knife motions (starting from $y=\SI{8}{\cm}$ moving downwards), with constraints on the lateral motion shaded in red.}
    \label{fig:exp-control}
\end{figure}

When optimizing \autoref{eq:slicing-objective} through Adam without constraints on the sideways knife position, the resulting motion after 50 iterations (orange line in right subplot of \autoref{fig:exp-control}) consists of an initial pressing phase up to the point of contact with the cucumber, after which the knife continuously moves sideways without sawing. To limit the sideways motion to remain within the bounds of the blade length, we add an inequality constraint in~\autoref{eq:slicing-lateral-constraint}. By performing constrained optimization with the modified differential method of multipliers (MDMM) (see \autoref{sec:mdmm}), we see that the knife moves within the bounds of the \SI{15}{\cm} blade length (green line on the right in \autoref{fig:exp-control}), while incurring only slightly more mean knife force (\SI{76.726}{\newton}) compared to the solution from the earlier unconstrained optimization (\SI{76.498}{\newton}).
For comparison, the mean knife force was \SI{89.698}{\newton} before the trajectory optimization.

\section{Real-robot Optimized Cutting Motion}
\label{sec:robot-opt-cutting}

\begin{figure}
    \centering
    \includegraphics[width=0.9\columnwidth]{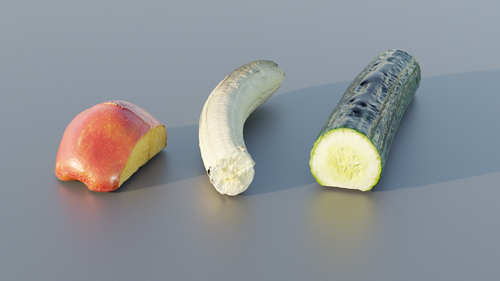}
    \caption{Watertight 3D meshes obtained from 3D scans of a real apple, a peeled banana, and a cucumber, as part of the real-robot experiments from \autoref{sec:robot-opt-cutting}. The textured meshes have been visualized in Blender via the Cycles renderer. For simulation, these meshes are downsampled and tetrahedralized.}
    \label{fig:3d-scans}
\end{figure}

\begin{figure}
    \centering
    \includegraphics[width=\columnwidth]{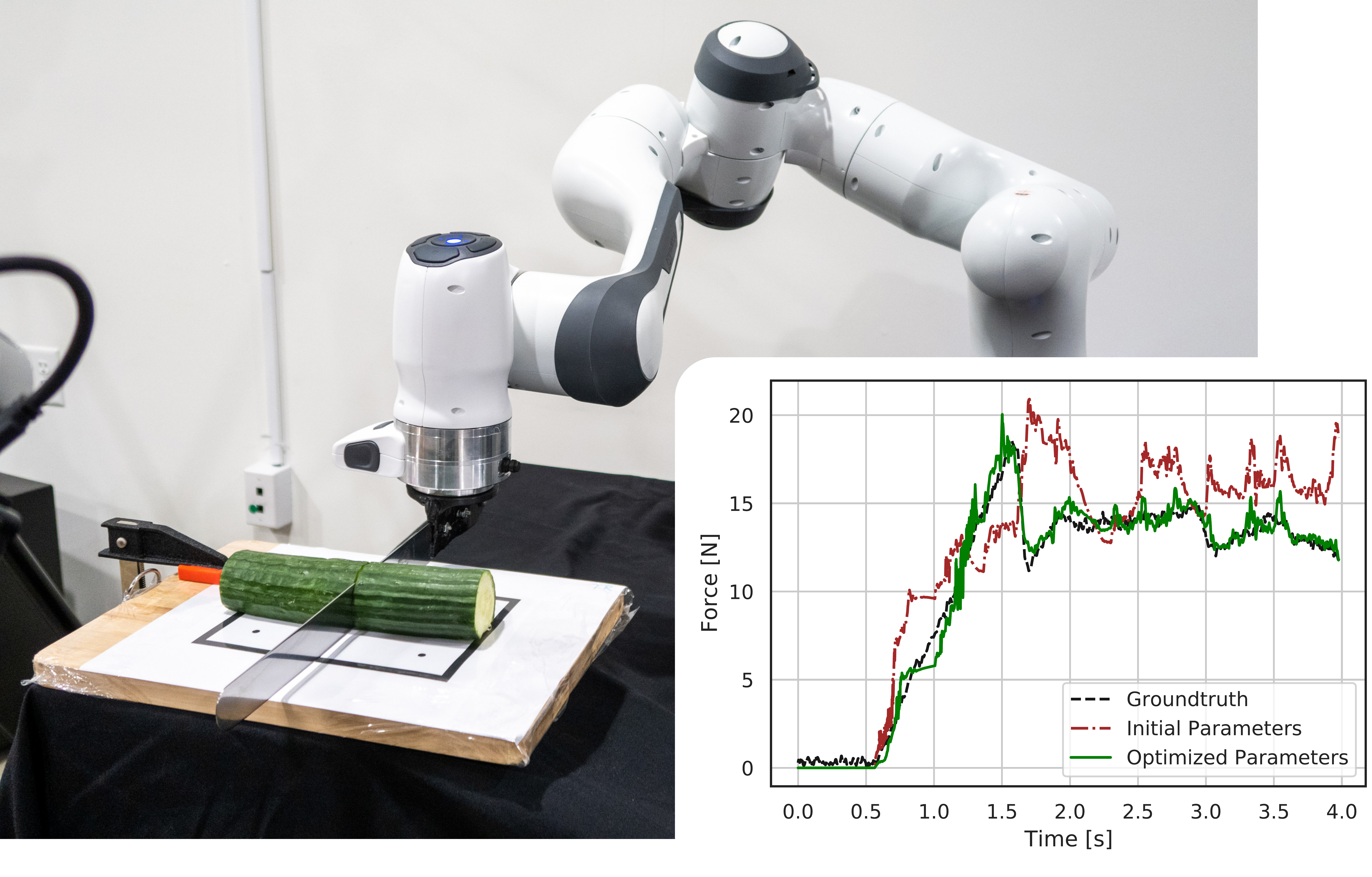}
    \caption{\textit{Left:} Experimental setup of our real-robot experiments where a Panda robot arm equipped with the slicing knife has cut vertically through a piece of cucumber. \textit{Right:} Force profile obtained from cutting this cucumber vertically (black), compared against the simulated force profile from before (red) and after (green) optimization of the simulation parameters.}
    \label{fig:real-robot-cucumber-vertical}
\end{figure}

In our final set of experiments, we validate the capabilities of our simulator to infer the simulation parameters and optimize the knife motion trajectory on a real robot.
We designed and 3D-printed a knife fixture for a thin slicing knife with a blade length of \SI{25}{\cm} that attaches as end-effector to a Franka Emika Panda robot arm (see \autoref{fig:real-robot-cucumber-vertical}).
Before slicing, we obtain highly detailed meshes of all the fruits to be cut via the turn-table-based EinScan SE 3D scanner (see \autoref{fig:3d-scans}). We generate watertight surface meshes and down-sample them before discretization with TetMeshWild~\citep{hu2018tetwild} to obtain adequate tetrahedral meshes used in our simulation. While force-torque sensors attached to the end-effector, such as in the experimental setup from \cite{jamdagni2019robotic}, may provide force measurements of higher accuracy, we found the linear force calculated from the joint forces measured by the torque sensors of the Panda arm very comparable in quality to the previously used dataset of real force profiles in \autoref{sec:experiments}.

\subsection{Parameter Inference}
\label{sec:real-parameter-inference}

Starting with a piece of cucumber, we command the robot arm via Cartesian velocity control to cut in a straight line vertically with a velocity of \SI{4}{\cm\per\second} downwards starting from a height of \SI{11}{\mm} of the blade above the cutting board (see \autoref{fig:real-robot-cucumber-vertical}). We continuously calculate the linear forces acting on the knife from the measurements of the torque sensors of the robot arm given the kinematic and inertial properties of the knife and its fixture computed by the Autodesk Fusion 360 CAD software. As before, we take the norm of the linear knife force at each time step as our ground-truth signal. Given such force profile, we estimate the simulation parameters leading to a significantly reduced reality gap to the simulated knife force (the mean force error at each time step declined from over \SI{14}{\newton} to less than \SI{1}{\newton}).

\begin{figure}
    \centering
    \includegraphics[align=c,width=0.33\columnwidth]{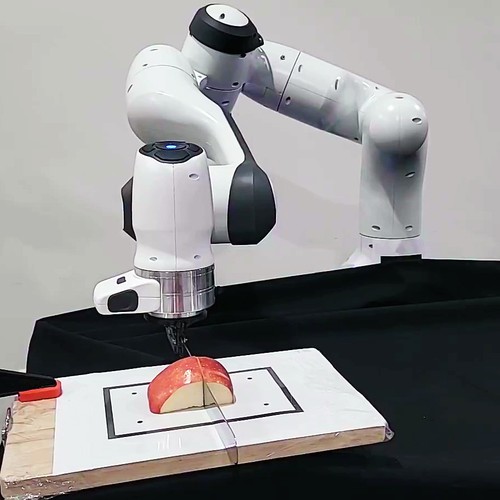}
    \includegraphics[align=c,width=0.65\columnwidth]{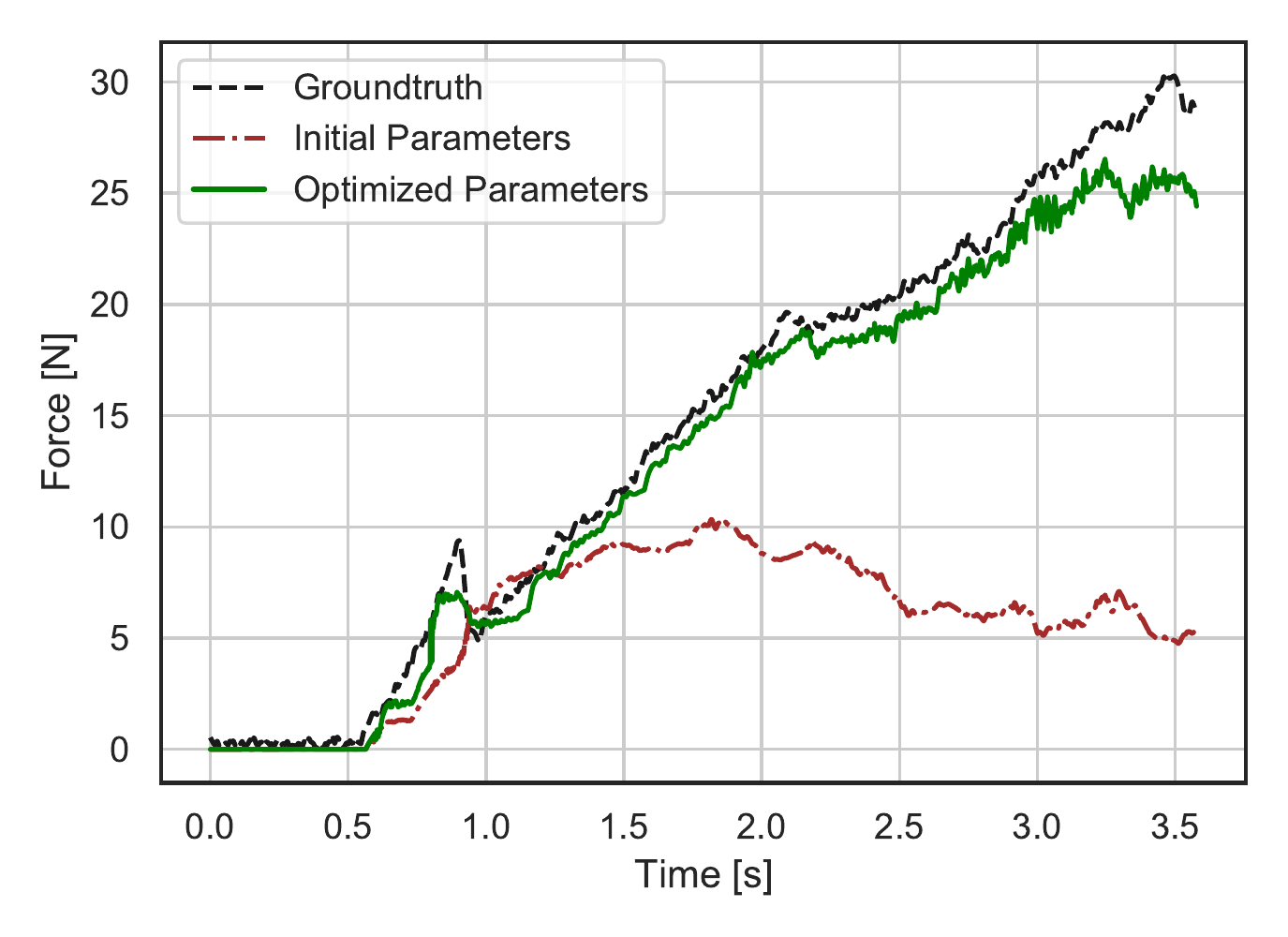}
    \caption{Vertical cutting of an apple.}
    \label{fig:real-apple-cut-vertical}
\end{figure}

\begin{figure}
    \centering
    \includegraphics[align=c,width=0.33\columnwidth]{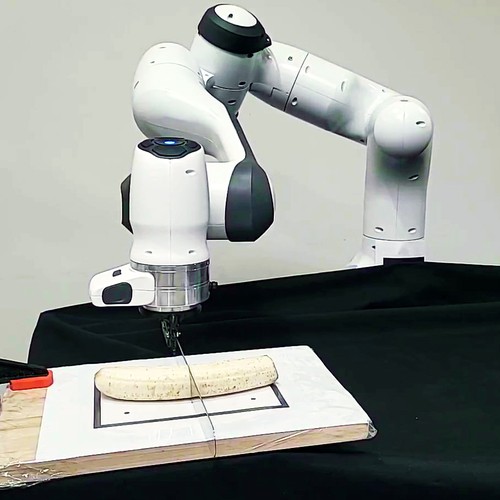}
    \includegraphics[align=c,width=0.65\columnwidth]{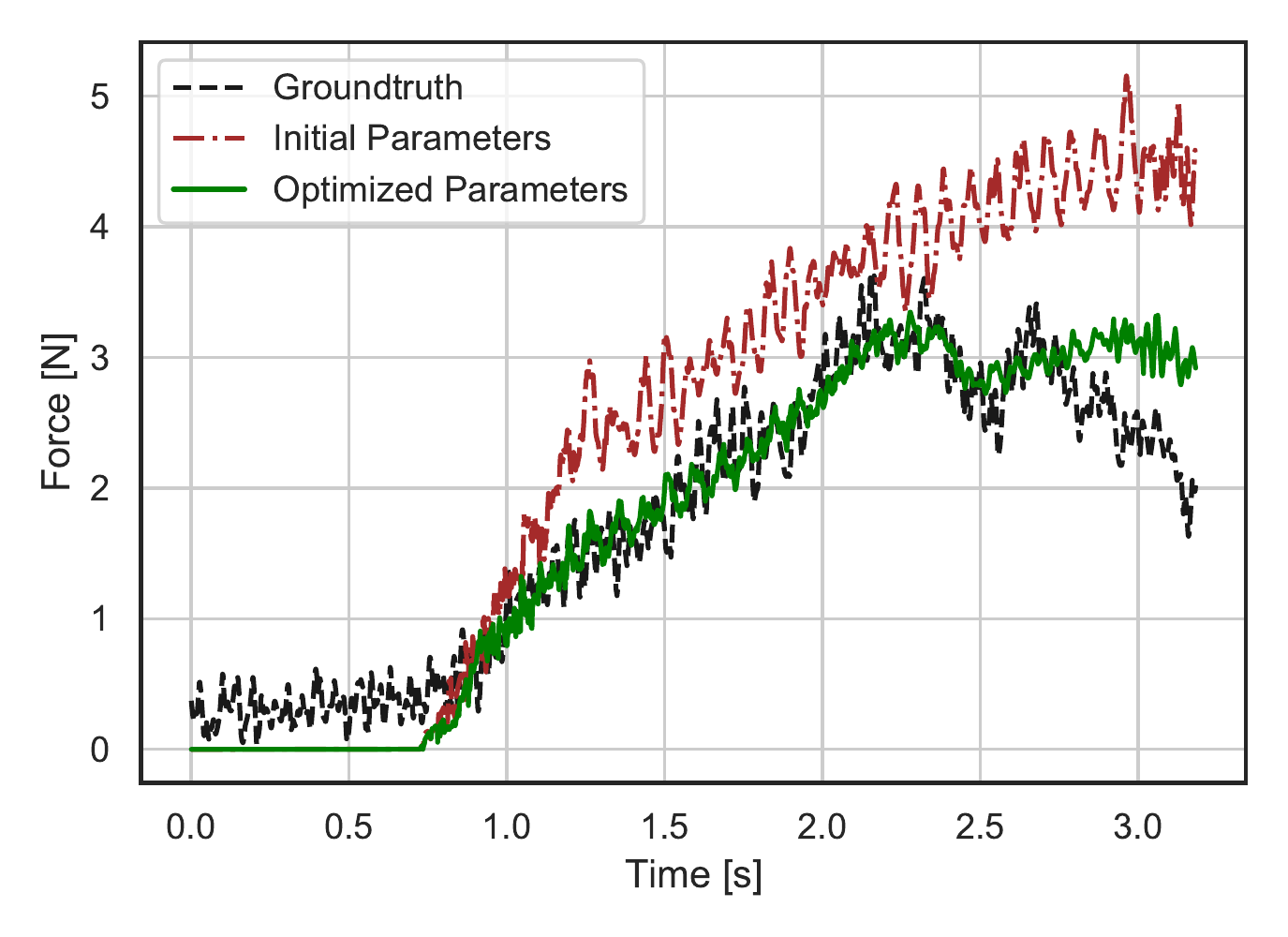}
    \caption{Vertical cutting of a peeled banana.}
    \label{fig:real-banana-cut-vertical}
\end{figure}

Within 200 iterations of the Adam optimizer, we find simulation parameters that yield a simulated force profile which closely matched the real measurements (see the force plot on the right of \autoref{fig:real-robot-cucumber-vertical}). We repeat the same cutting motions on an apple and a peeled banana, which are fruits that are on the extreme ends of stiffness and toughness within the foodstuffs we investigated in this work. As shown in Figures \ref{fig:real-apple-cut-vertical} and \ref{fig:real-banana-cut-vertical} for the apple and the banana, respectively, the optimized simulation parameters achieve a significantly lower reality gap in the generated force profiles than at the beginning of the simulation (the force profile given the initial parameters is plotted in red).

We quantify the accuracy in the simulated force profiles via the normalized mean absolute error metric (NMAE) again, where the MAE between the simulated and real force profile is divided by the mean of the real force profile. As reported in the first two rows of \autoref{tab:real-results}, the NMAE shrunk significantly between the initial setting of the simulation parameters and the final result of the optimization.

\begin{figure*}
    \centering
    \resizebox{\textwidth}{!}{
    \begin{tabular}{ccc}
        \bf Apple & \bf Banana & \bf Cucumber \\
        \includegraphics[width=0.76\columnwidth]{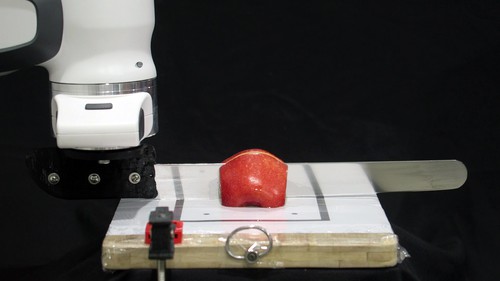} & 
        \includegraphics[width=0.76\columnwidth]{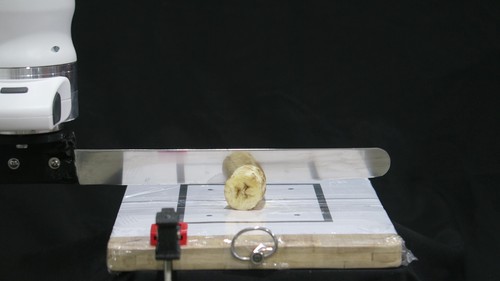} &
        \includegraphics[width=0.76\columnwidth]{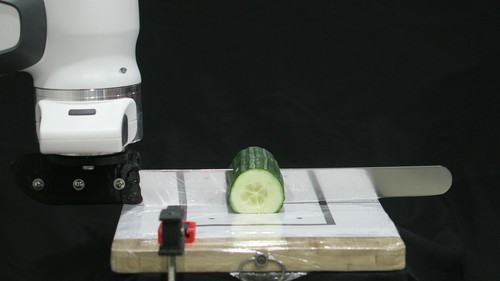} \\
        \includegraphics[width=0.8\columnwidth]{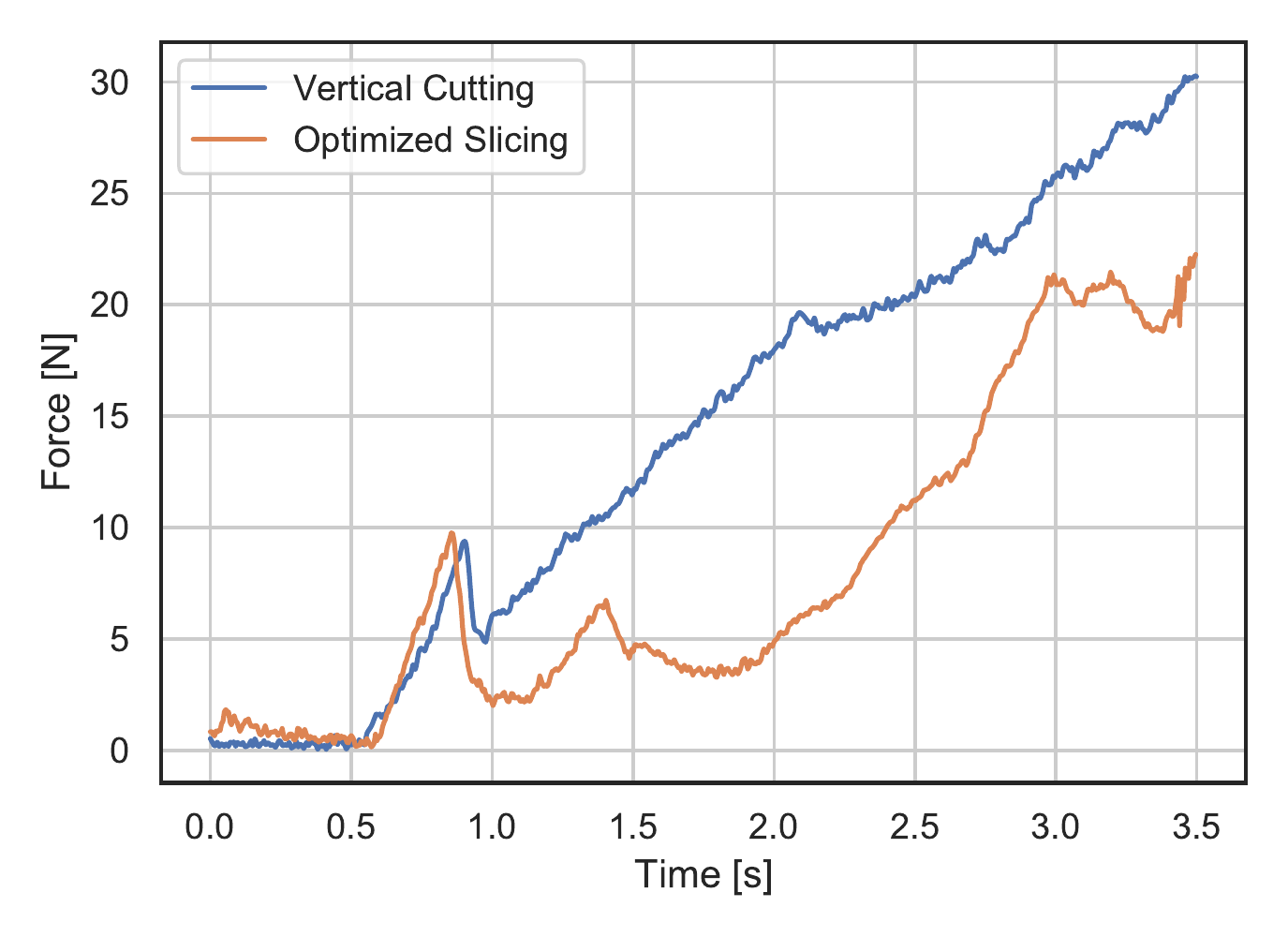} &
        \includegraphics[width=0.8\columnwidth]{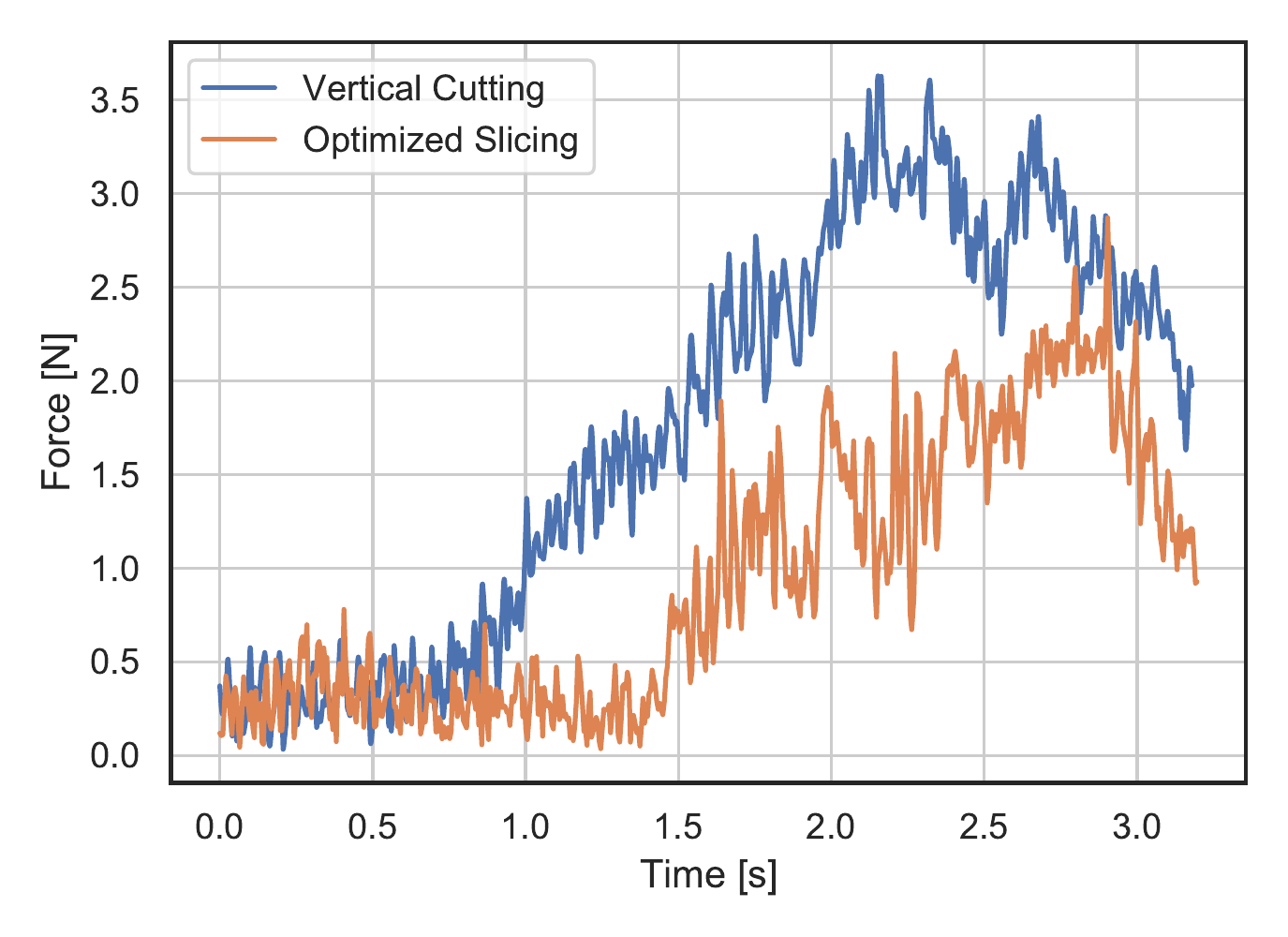} & 
        \includegraphics[width=0.8\columnwidth]{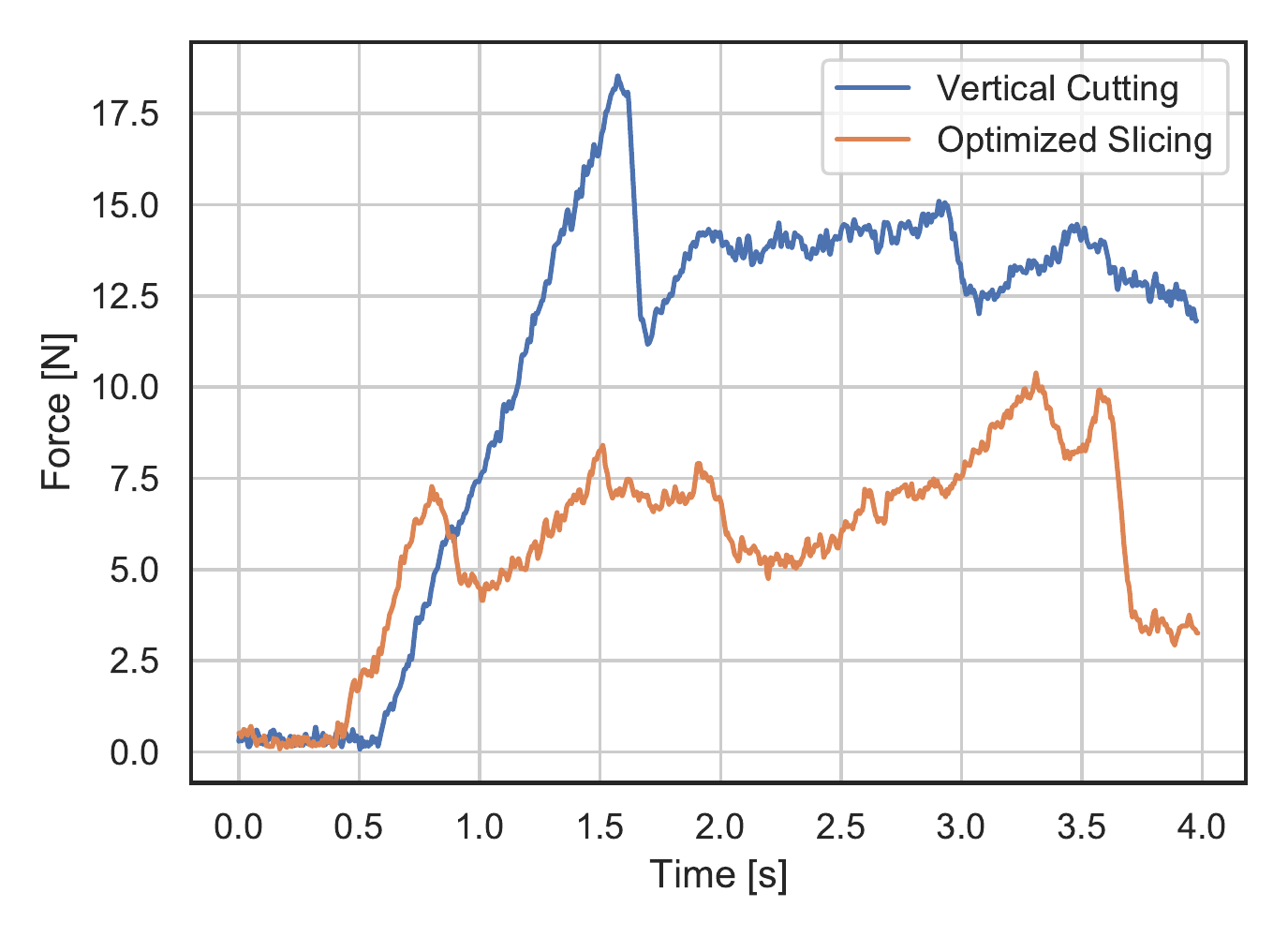} \\
        \includegraphics[width=0.87\columnwidth]{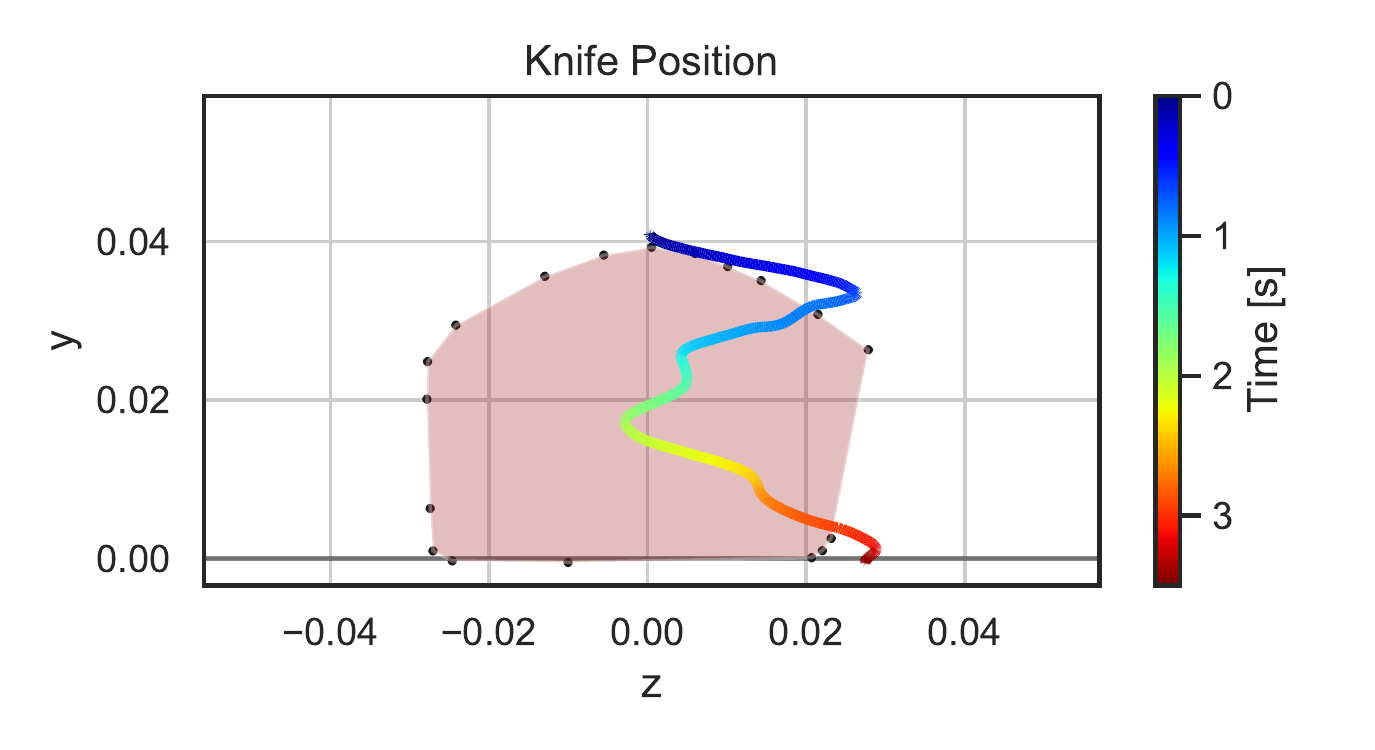} &
        \includegraphics[width=0.87\columnwidth]{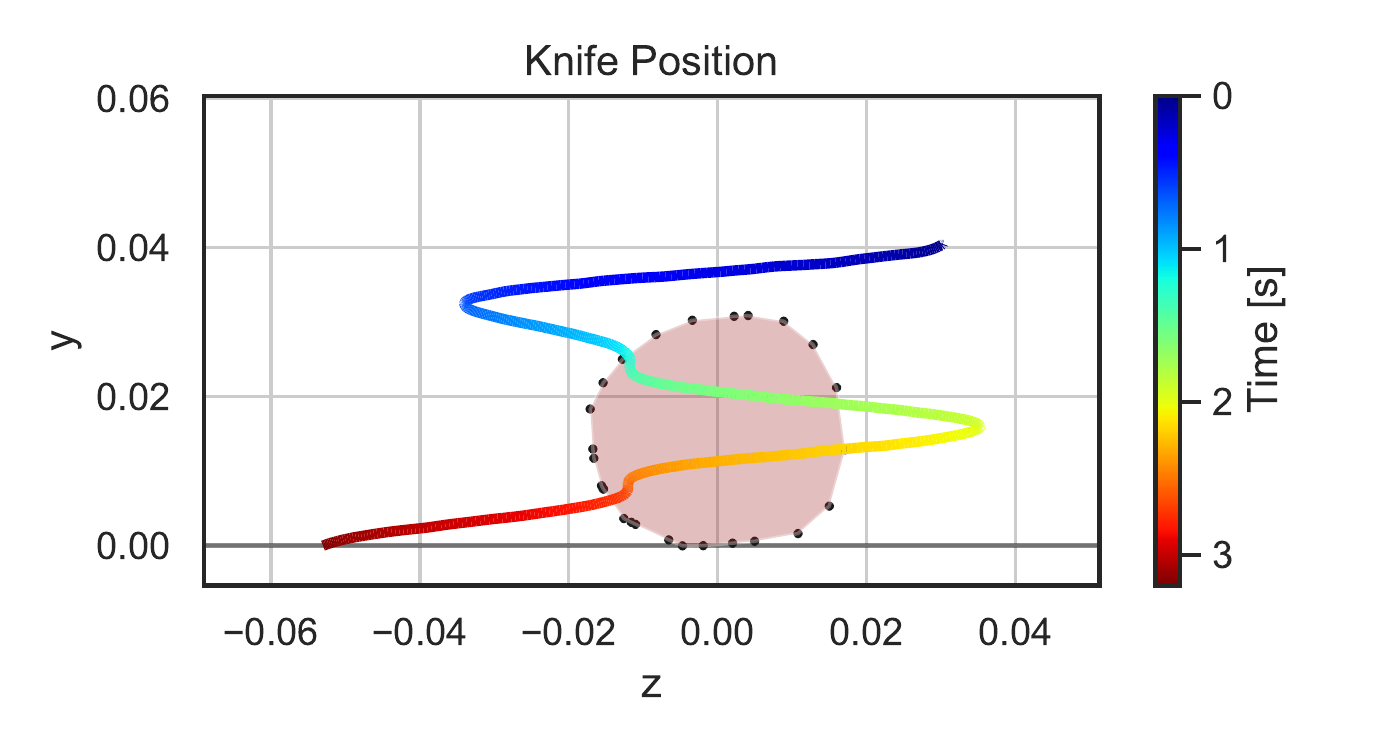} &
        \includegraphics[width=0.87\columnwidth]{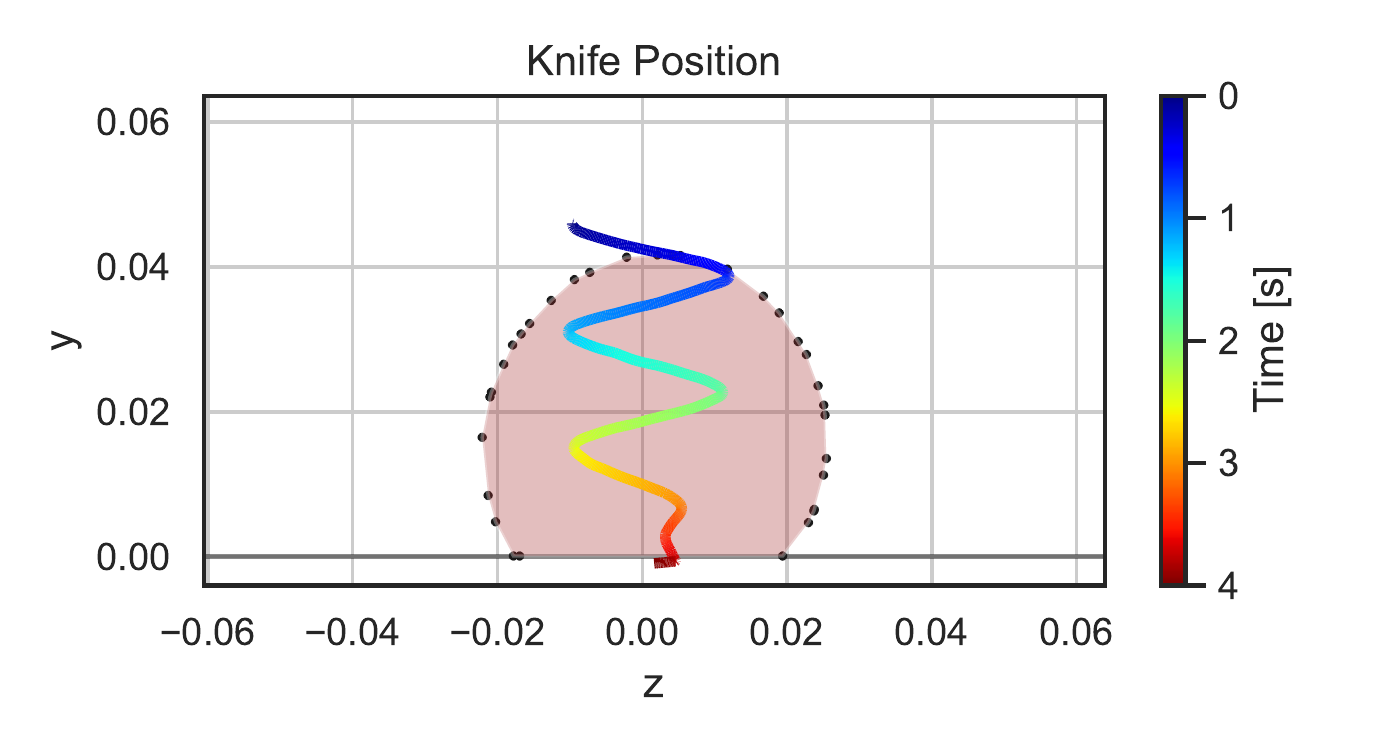}
    \end{tabular}}
    \caption{Real-robot executions of the optimized slicing motions for an apple (left column), a peeled banana (center column), and a cucumber (right column). \textit{First row:} side view of the experimental setup. \textit{Second row:} comparison of measured knife force profiles from a vertical cut (blue) and the optimized slicing motion (orange). \textit{Third row:} trajectory of the knife following the optimized slicing motion. The colorful line indicates the position of the blade center point over time, from the first (dark blue) to the last time step (dark red) of the motion. The shaded polygon behind the trajectory line indicates the silhouette of the mesh being cut (orthographic projection of the convex hull of the fruit's shape).}
    \label{fig:real-slicing}
\end{figure*}

\subsection{Trajectory Optimization}
\label{sec:real-parameter-inference}

Having optimized the simulation parameters from measurements obtained through a vertical cut, we now seek to find cutting motions that require less effort from the robot and thereby exert less force on the material which potentially damages the fruit. To this end, we select a half of the mesh that we cut previously as our new cutting target where we aim to optimize a lateral slicing motion, similar to our setup from \autoref{sec:exp-control}. Since the mesh topology has changed between the previous vertical cut and the new part of the fruit that we want to slice, we transport the previously optimized simulation parameters to the new cutting surface via the optimal transport technique we developed in \autoref{sec:optimal-transport}.

Given the simulation setup and transported cutting spring parameters, we are now ready to optimize the lateral slicing motion of the knife. We follow our formulation from \autoref{eq:slicing-objective} where we parameterize the motion through five way points that encode the slicing amplitude and vertical velocity. At the same time, we again enforce the constraints to (1) ensure the knife touches the cutting board at the end of the cut, and (2) that the lateral motion of the knife does not exceed the blade length. We execute the optimized slicing motions on the real robot, as shown in \autoref{fig:real-slicing}, for the apple, banana, and cucumber. The fruits are fixed to the cutting board via water-resistant super glue to ensure that they remain in place without the need of an external fixture. Analogously, in simulation, we also add boundary conditions to the lowest mesh vertices that touch the cutting board to keep them at a fixed location.

With an initialization of the slicing amplitude to zero, which corresponds to a vertical cut, the optimized motions exhibit a pronounced slicing trajectory (see bottom row in \autoref{fig:real-slicing}). Particularly for the banana (middle column), the knife moves laterally by more than ten centimeters, while the lateral motions on the apple and cucumber remain smaller. In the measured forces, we see a clear benefit of the obtained slicing motions. As shown in the second row of \autoref{fig:real-slicing}, the force profile of the optimized knife trajectory (orange) is considerable lower than its counterpart of the vertical cutting motion (blue). We quantify the reduction in exerted knife force in the bottom two rows in \autoref{tab:real-results}. The mean of the knife force over the entire length of the trajectory is consistently reduced by more than $40\%$, while the peak force (maximum of force profile) has shrunk by $20\%$ (banana) to more than $40\%$ (cucumber). Since the cutting of a peeled banana requires significantly less force than the other fruits, it is expected that the maximal force does not reduce by as much compared to a stiffer material, such as a cucumber or an apple.

\begin{table}[t]
    \centering
    \resizebox{\columnwidth}{!}{\begin{tabular}{lccc}\toprule
         & \bf Apple & \bf Banana & \bf Cucumber \\\midrule
        \multicolumn{4}{l}{Simulation accuracy \textit{before} parameter inference} \\
        \bf Knife force NMAE & 0.606 & 0.487 & 0.231 \\\midrule
        \multicolumn{4}{l}{Simulation accuracy \textit{after} parameter inference} \\
        \bf Knife force NMAE & 0.083 & 0.160 & 0.050 \\\midrule
        \multicolumn{4}{l}{Trajectory optimization performance} \\
        \bf Mean force reduction & 42.7 \% & 46.5 \% & 46.7 \% \\
        \bf Max force reduction & 26.5 \% & 20.9 \% & 43.9 \% \\
        \bottomrule
    \end{tabular}}
    \caption{Summary of results in parameter inference and trajectory optimization from our real-robot experiments in \autoref{sec:robot-opt-cutting}.}
    \label{tab:real-results}
\end{table}

\section{Conclusions}
\label{sec:conclusion}

We presented the first differentiable simulator for the robotic cutting of soft materials. Differentiability was achieved through a continuous contact formulation, the insertion of virtual nodes along a cutting plane, and a continuous damage model based on the progressive weakening of springs. The advantages of differentiability were shown through a systematic comparison of multiple gradient-based and derivative-free methods for optimizing the simulation parameters; leveraging gradients from the simulator enabled highly efficient estimation of posteriors over hundreds of parameters. The calibrated simulator was able to reproduce ground-truth data from a state-of-the-art commercial simulator in a fraction of the time, as well as closely match data from a real-world cutting dataset. Simulator predictions generalized across cutting velocities, object instances, and object geometries. A control experiment was performed in which human pressing-and-slicing behavior emerged from sample-efficient constrained optimization applied to the differentiable simulator, reducing the mean knife force by $15\%$ relative to a vertical cut. Finally, we applied the capabilities of parameter inference, parameter transport, and trajectory optimization to a real-robot cutting scenario where the sim2real gap was closed and the measured knife forces reduced significantly.

In future work, we plan to make multiple extensions. First, we will extend the material model for the soft material to explicitly capture nonlinearity and isotropy, as commonly observed in biomaterials; explicitly specifying such behaviors will facilitate optimization of simulator parameters. In addition, we will extend our modeling approach to accommodate more complex cutting actions, such as carving, in which the knife may arbitrarily rotate and follow more diverse trajectories than sawing motions; this approach will be used in additional control experiments, with the resulting trajectories compared again to human actions. Ultimately, we envision the use of differentiable cutting simulators in applications as challenging as robotic surgery, where tissue parameters can accurately and efficiently be estimated on initial contact with the cutting instrument, and the calibrated simulator can be used for faster-than-real-time roll-outs in a model-predictive framework for online surgical planning.

\section*{Acknowledgments}
We thank Yan-Bin Jia and Prajjwal Jamdagni for providing the dataset of real-world cutting trajectories and meshes that we used in our experiments from \autoref{sec:experiments}.
We are grateful to Krishna Mellachervu for outstanding technical support with commercial solvers.
We thank David Millard for helping to set up the robot infrastructure and guiding us in the fabrication of the knife fixture.
Finally, we thank Mike Skolones for his mentorship. This work was supported by a Google Ph.D. Fellowship.

\bibliography{literature}  %

\begin{thebibliography}{}
\providecommand{\doi}[1]{\url{https://doi.org/#1}}
\bibcommenthead

\bibitem [\protect \citeauthoryear {%
Areias%
\ \BBA {} Rabczuk%
}{%
Areias%
\ \BBA {} Rabczuk%
}{%
{\protect \APACyear {2017}}%
}]{%
areias2017steiner}
\APACinsertmetastar {%
areias2017steiner}%
\begin{APACrefauthors}%
Areias, P.%
\BCBT {}\ \BBA {} Rabczuk, T.%
\end{APACrefauthors}%
\unskip\
\newblock
\APACrefYearMonthDay{2017}{}{}.
\newblock
{\BBOQ}\APACrefatitle {Steiner-point free edge cutting of tetrahedral meshes
  with applications in fracture} {Steiner-point free edge cutting of
  tetrahedral meshes with applications in fracture}.{\BBCQ}
\newblock
\APACjournalVolNumPages{Finite Elements in Analysis and Design}{132}{}{27--41}.
\newblock

\newblock

\PrintBackRefs{\CurrentBib}

\bibitem [\protect \citeauthoryear {%
Atkins%
\ \BBA {} Atkins%
}{%
Atkins%
\ \BBA {} Atkins%
}{%
{\protect \APACyear {2009}}%
}]{%
atkins2009science}
\APACinsertmetastar {%
atkins2009science}%
\begin{APACrefauthors}%
Atkins, A.%
\BCBT {}\ \BBA {} Atkins, T.%
\end{APACrefauthors}%
\unskip\
\newblock
\APACrefYear{2009}.
\newblock
\APACrefbtitle {The Science and Engineering of Cutting: The Mechanics and
  Processes of Separating, Scratching and Puncturing Biomaterials, Metals and
  Non-metals} {The science and engineering of cutting: The mechanics and
  processes of separating, scratching and puncturing biomaterials, metals and
  non-metals}.
\newblock
\APACaddressPublisher{}{Elsevier Science}.
\PrintBackRefs{\CurrentBib}

\bibitem [\protect \citeauthoryear {%
Atkins%
, Xu%
\BCBL {}\ \BBA {} Jeronimidis%
}{%
Atkins%
\ \protect \BOthers {.}}{%
{\protect \APACyear {2004}}%
}]{%
atkins2004cutting}
\APACinsertmetastar {%
atkins2004cutting}%
\begin{APACrefauthors}%
Atkins, A.%
, Xu, X.%
\BCBL {} Jeronimidis, G.%
\end{APACrefauthors}%
\unskip\
\newblock
\APACrefYearMonthDay{2004}{}{}.
\newblock
{\BBOQ}\APACrefatitle {Cutting by `pressing and slicing' of thin floppy slices
  of materials illustrated by experiments on cheddar cheese and salami}
  {Cutting by `pressing and slicing' of thin floppy slices of materials
  illustrated by experiments on cheddar cheese and salami}.{\BBCQ}
\newblock
\APACjournalVolNumPages{Journal of Materials Science}{39}{}{2761-2766}.
\newblock

\newblock

\PrintBackRefs{\CurrentBib}

\bibitem [\protect \citeauthoryear {%
Belytschko%
, Liu%
, Moran%
\BCBL {}\ \BBA {} Elkhodary%
}{%
Belytschko%
\ \protect \BOthers {.}}{%
{\protect \APACyear {2013}}%
}]{%
belytschko2013nonlinear}
\APACinsertmetastar {%
belytschko2013nonlinear}%
\begin{APACrefauthors}%
Belytschko, T.%
, Liu, W.K.%
, Moran, B.%
\BCBL {} Elkhodary, K.%
\end{APACrefauthors}%
\unskip\
\newblock
\APACrefYear{2013}.
\newblock
\APACrefbtitle {Nonlinear finite elements for continua and structures}
  {Nonlinear finite elements for continua and structures}.
\newblock
\APACaddressPublisher{}{John Wiley \& Sons}.
\PrintBackRefs{\CurrentBib}

\bibitem [\protect \citeauthoryear {%
Belytschko%
, Lu%
\BCBL {}\ \BBA {} Gu%
}{%
Belytschko%
\ \protect \BOthers {.}}{%
{\protect \APACyear {1994}}%
}]{%
belytschko1994efg}
\APACinsertmetastar {%
belytschko1994efg}%
\begin{APACrefauthors}%
Belytschko, T.%
, Lu, Y.%
\BCBL {} Gu, L.%
\end{APACrefauthors}%
\unskip\
\newblock
\APACrefYearMonthDay{1994}{}{}.
\newblock
{\BBOQ}\APACrefatitle {Element-free Galerkin methods} {Element-free galerkin
  methods}.{\BBCQ}
\newblock
\APACjournalVolNumPages{International Journal for Numerical Methods in
  Engineering}{37}{}{229--256}.
\newblock

\newblock

\PrintBackRefs{\CurrentBib}

\bibitem [\protect \citeauthoryear {%
{Berndt}%
, {Torchelsen}%
\BCBL {}\ \BBA {} {Maciel}%
}{%
{Berndt}%
\ \protect \BOthers {.}}{%
{\protect \APACyear {2017}}%
}]{%
berndt2017pbd}
\APACinsertmetastar {%
berndt2017pbd}%
\begin{APACrefauthors}%
{Berndt}, I.%
, {Torchelsen}, R.%
\BCBL {} {Maciel}, A.%
\end{APACrefauthors}%
\unskip\
\newblock
\APACrefYearMonthDay{2017}{}{}.
\newblock
{\BBOQ}\APACrefatitle {Efficient Surgical Cutting with Position-Based Dynamics}
  {Efficient surgical cutting with position-based dynamics}.{\BBCQ}
\newblock
\APACjournalVolNumPages{IEEE Computer Graphics and Applications}{37}{3}{24-31}.
\newblock

\newblock

\PrintBackRefs{\CurrentBib}

\bibitem [\protect \citeauthoryear {%
Bielser%
, Glardon%
, Teschner%
\BCBL {}\ \BBA {} Gross%
}{%
Bielser%
\ \protect \BOthers {.}}{%
{\protect \APACyear {2003}}%
}]{%
bielser2003state}
\APACinsertmetastar {%
bielser2003state}%
\begin{APACrefauthors}%
Bielser, D.%
, Glardon, P.%
, Teschner, M.%
\BCBL {} Gross, M.%
\end{APACrefauthors}%
\unskip\
\newblock
\APACrefYearMonthDay{2003}{}{}.
\newblock
{\BBOQ}\APACrefatitle {A state machine for real-time cutting of tetrahedral
  meshes} {A state machine for real-time cutting of tetrahedral meshes}.{\BBCQ}
\newblock
 \APACrefbtitle {11th Pacific Conference on Computer Graphics and Applications}
  {11th pacific conference on computer graphics and applications}\ (\BPGS\
  377--386).
\PrintBackRefs{\CurrentBib}

\bibitem [\protect \citeauthoryear {%
Brown%
}{%
Brown%
}{%
{\protect \APACyear {2017}}%
}]{%
brown2017contact}
\APACinsertmetastar {%
brown2017contact}%
\begin{APACrefauthors}%
Brown, P.%
\end{APACrefauthors}%
\unskip\
\newblock
\APACrefYear{2017}.
\unskip\
\newblock
\APACrefbtitle {Contact modelling for forward dynamics of human motion}
  {Contact modelling for forward dynamics of human motion}\
  \APACtypeAddressSchool {\BUMTh}{}{}.
\unskip\
\newblock
\APACaddressSchool {}{University of Waterloo}.
\PrintBackRefs{\CurrentBib}

\bibitem [\protect \citeauthoryear {%
Burkhart%
, Hamann%
\BCBL {}\ \BBA {} Umlauf%
}{%
Burkhart%
\ \protect \BOthers {.}}{%
{\protect \APACyear {2010}}%
}]{%
burkhart2010adaptive}
\APACinsertmetastar {%
burkhart2010adaptive}%
\begin{APACrefauthors}%
Burkhart, D.%
, Hamann, B.%
\BCBL {} Umlauf, G.%
\end{APACrefauthors}%
\unskip\
\newblock
\APACrefYearMonthDay{2010}{}{}.
\newblock
{\BBOQ}\APACrefatitle {Adaptive and feature-preserving subdivision for
  high-quality tetrahedral meshes} {Adaptive and feature-preserving subdivision
  for high-quality tetrahedral meshes}.{\BBCQ}
\newblock
\APACjournalVolNumPages{Computer Graphics Forum}{29}{1}{117--127}.
\newblock

\newblock

\PrintBackRefs{\CurrentBib}

\bibitem [\protect \citeauthoryear {%
Carpentier%
\ \BBA {} Mansard%
}{%
Carpentier%
\ \BBA {} Mansard%
}{%
{\protect \APACyear {2018}}%
}]{%
carpentier2018analytical}
\APACinsertmetastar {%
carpentier2018analytical}%
\begin{APACrefauthors}%
Carpentier, J.%
\BCBT {}\ \BBA {} Mansard, N.%
\end{APACrefauthors}%
\unskip\
\newblock
\APACrefYearMonthDay{2018}{}{}.
\newblock
{\BBOQ}\APACrefatitle {Analytical Derivatives of Rigid Body Dynamics
  Algorithms} {Analytical derivatives of rigid body dynamics
  algorithms}.{\BBCQ}
\newblock
 \APACrefbtitle {Robotics: Science and Systems.} {Robotics: Science and
  systems.}
\PrintBackRefs{\CurrentBib}

\bibitem [\protect \citeauthoryear {%
Chebotar%
\ \protect \BOthers {.}}{%
Chebotar%
\ \protect \BOthers {.}}{%
{\protect \APACyear {2019}}%
}]{%
chebotar2019closing}
\APACinsertmetastar {%
chebotar2019closing}%
\begin{APACrefauthors}%
Chebotar, Y.%
, Handa, A.%
, Makoviychuk, V.%
, Macklin, M.%
, Issac, J.%
, Ratliff, N.%
\BCBL {} Fox, D.%
\end{APACrefauthors}%
\unskip\
\newblock
\APACrefYearMonthDay{2019}{}{}.
\newblock
{\BBOQ}\APACrefatitle {Closing the sim-to-real loop: Adapting simulation
  randomization with real world experience} {Closing the sim-to-real loop:
  Adapting simulation randomization with real world experience}.{\BBCQ}
\newblock
 \APACrefbtitle {2019 International Conference on Robotics and Automation
  (ICRA)} {2019 international conference on robotics and automation (icra)}\
  (\BPGS\ 8973--8979).
\PrintBackRefs{\CurrentBib}

\bibitem [\protect \citeauthoryear {%
Cranmer%
, Brehmer%
\BCBL {}\ \BBA {} Louppe%
}{%
Cranmer%
\ \protect \BOthers {.}}{%
{\protect \APACyear {2020}}%
}]{%
cranmer2020sbi}
\APACinsertmetastar {%
cranmer2020sbi}%
\begin{APACrefauthors}%
Cranmer, K.%
, Brehmer, J.%
\BCBL {} Louppe, G.%
\end{APACrefauthors}%
\unskip\
\newblock
\APACrefYearMonthDay{2020}{}{}.
\newblock
{\BBOQ}\APACrefatitle {The frontier of simulation-based inference} {The
  frontier of simulation-based inference}.{\BBCQ}
\newblock
\APACjournalVolNumPages{Proceedings of the National Academy of
  Sciences}{117}{48}{30055--30062}.
\newblock

\newblock

\PrintBackRefs{\CurrentBib}

\bibitem [\protect \citeauthoryear {%
Cunningham%
}{%
Cunningham%
}{%
{\protect \APACyear {1976}}%
}]{%
Cunningham1976simplex}
\APACinsertmetastar {%
Cunningham1976simplex}%
\begin{APACrefauthors}%
Cunningham, W.H.%
\end{APACrefauthors}%
\unskip\
\newblock
\APACrefYearMonthDay{1976}{}{}.
\newblock
{\BBOQ}\APACrefatitle {A network simplex method} {A network simplex
  method}.{\BBCQ}
\newblock
\APACjournalVolNumPages{Mathematical Programming}{11}{}{105--116}.
\newblock

\newblock

\PrintBackRefs{\CurrentBib}

\bibitem [\protect \citeauthoryear {%
de Avila Belbute-Peres%
, Smith%
, Allen%
, Tenenbaum%
\BCBL {}\ \BBA {} Kolter%
}{%
de Avila Belbute-Peres%
\ \protect \BOthers {.}}{%
{\protect \APACyear {2018}}%
}]{%
peres2018lcp}
\APACinsertmetastar {%
peres2018lcp}%
\begin{APACrefauthors}%
de Avila Belbute-Peres, F.%
, Smith, K.%
, Allen, K.%
, Tenenbaum, J.%
\BCBL {} Kolter, J.Z.%
\end{APACrefauthors}%
\unskip\
\newblock
\APACrefYearMonthDay{2018}{}{}.
\newblock
{\BBOQ}\APACrefatitle {End-to-End Differentiable Physics for Learning and
  Control} {End-to-end differentiable physics for learning and control}.{\BBCQ}
\newblock
 \APACrefbtitle {Advances in Neural Information Processing Systems 31}
  {Advances in neural information processing systems 31}\ (\BPGS\ 7178--7189).
\PrintBackRefs{\CurrentBib}

\bibitem [\protect \citeauthoryear {%
{Debao Zhou}%
, {Claffee}%
, {Kok-Meng Lee}%
\BCBL {}\ \BBA {} {McMurray}%
}{%
{Debao Zhou}%
\ \protect \BOthers {.}}{%
{\protect \APACyear {2006}}%
{\protect \APACexlab {{\protect \BCnt {1}}}}}]{%
zhou2006cut1}
\APACinsertmetastar {%
zhou2006cut1}%
\begin{APACrefauthors}%
{Debao Zhou}%
, {Claffee}, M.R.%
, {Kok-Meng Lee}%
\BCBL {} {McMurray}, G.V.%
\end{APACrefauthors}%
\unskip\
\newblock
\APACrefYearMonthDay{2006{\protect \BCnt {1}}}{}{}.
\newblock
{\BBOQ}\APACrefatitle {Cutting, "by pressing and slicing", applied to robotic
  cutting bio-materials. I. Modeling of stress distribution} {Cutting, "by
  pressing and slicing", applied to robotic cutting bio-materials. i. modeling
  of stress distribution}.{\BBCQ}
\newblock
 \APACrefbtitle {Proceedings 2006 IEEE International Conference on Robotics and
  Automation} {Proceedings 2006 ieee international conference on robotics and
  automation}\ (\BPG~2896-2901).
\PrintBackRefs{\CurrentBib}

\bibitem [\protect \citeauthoryear {%
{Debao Zhou}%
, {Claffee}%
, {Kok-Meng Lee}%
\BCBL {}\ \BBA {} {McMurray}%
}{%
{Debao Zhou}%
\ \protect \BOthers {.}}{%
{\protect \APACyear {2006}}%
{\protect \APACexlab {{\protect \BCnt {2}}}}}]{%
zhou2006cut2}
\APACinsertmetastar {%
zhou2006cut2}%
\begin{APACrefauthors}%
{Debao Zhou}%
, {Claffee}, M.R.%
, {Kok-Meng Lee}%
\BCBL {} {McMurray}, G.V.%
\end{APACrefauthors}%
\unskip\
\newblock
\APACrefYearMonthDay{2006{\protect \BCnt {2}}}{}{}.
\newblock
{\BBOQ}\APACrefatitle {Cutting, 'by pressing and slicing', applied to the
  robotic cut of bio-materials. II. Force during slicing and pressing cuts}
  {Cutting, 'by pressing and slicing', applied to the robotic cut of
  bio-materials. ii. force during slicing and pressing cuts}.{\BBCQ}
\newblock
 \APACrefbtitle {Proceedings 2006 IEEE International Conference on Robotics and
  Automation, 2006. ICRA 2006.} {Proceedings 2006 ieee international conference
  on robotics and automation, 2006. icra 2006.}\ (\BPG~2256-2261).
\PrintBackRefs{\CurrentBib}

\bibitem [\protect \citeauthoryear {%
El~Said%
, Atallah%
, Khalil%
\BCBL {}\ \BBA {} El-Lithy%
}{%
El~Said%
\ \protect \BOthers {.}}{%
{\protect \APACyear {2011}}%
}]{%
el2011cucumber}
\APACinsertmetastar {%
el2011cucumber}%
\begin{APACrefauthors}%
El~Said, I.%
, Atallah, M.M.%
, Khalil, K.%
\BCBL {} El-Lithy, A.%
\end{APACrefauthors}%
\unskip\
\newblock
\APACrefYearMonthDay{2011}{}{}.
\newblock
{\BBOQ}\APACrefatitle {Physical and mechanical properties of cucumber applied
  to seed extractor} {Physical and mechanical properties of cucumber applied to
  seed extractor}.{\BBCQ}
\newblock
\APACjournalVolNumPages{Journal of Soil Sciences and Agricultural
  Engineering}{2}{8}{871--880}.
\newblock

\newblock

\PrintBackRefs{\CurrentBib}

\bibitem [\protect \citeauthoryear {%
Flamary%
\ \BBA {} Courty%
}{%
Flamary%
\ \BBA {} Courty%
}{%
{\protect \APACyear {2017}}%
}]{%
flamary2017pot}
\APACinsertmetastar {%
flamary2017pot}%
\begin{APACrefauthors}%
Flamary, R.%
\BCBT {}\ \BBA {} Courty, N.%
\end{APACrefauthors}%
\unskip\
\newblock
\APACrefYearMonthDay{2017}{}{}.
\newblock
\APACrefbtitle {POT Python Optimal Transport library.} {Pot python optimal
  transport library.}
\newblock
\begin{APACrefURL} {https://pythonot.github.io/} \end{APACrefURL}
\PrintBackRefs{\CurrentBib}

\bibitem [\protect \citeauthoryear {%
Geilinger%
\ \protect \BOthers {.}}{%
Geilinger%
\ \protect \BOthers {.}}{%
{\protect \APACyear {2020}}%
}]{%
geilinger2020add}
\APACinsertmetastar {%
geilinger2020add}%
\begin{APACrefauthors}%
Geilinger, M.%
, Hahn, D.%
, Zehnder, J.%
, Bächer, M.%
, Thomaszewski, B.%
\BCBL {} Coros, S.%
\end{APACrefauthors}%
\unskip\
\newblock
\APACrefYearMonthDay{2020}{}{}.
\newblock
{\BBOQ}\APACrefatitle {ADD: Analytically Differentiable Dynamics for Multi-Body
  Systems with Frictional Contact} {Add: Analytically differentiable dynamics
  for multi-body systems with frictional contact}.{\BBCQ}
\newblock
 \APACrefbtitle {arXiv.} {arxiv.}
\PrintBackRefs{\CurrentBib}

\bibitem [\protect \citeauthoryear {%
Giftthaler%
\ \protect \BOthers {.}}{%
Giftthaler%
\ \protect \BOthers {.}}{%
{\protect \APACyear {2017}}%
}]{%
giftthaler2017autodiff}
\APACinsertmetastar {%
giftthaler2017autodiff}%
\begin{APACrefauthors}%
Giftthaler, M.%
, Neunert, M.%
, St{\"a}uble, M.%
, Frigerio, M.%
, Semini, C.%
\BCBL {} Buchli, J.%
\end{APACrefauthors}%
\unskip\
\newblock
\APACrefYearMonthDay{2017}{}{}.
\newblock
{\BBOQ}\APACrefatitle {Automatic differentiation of rigid body dynamics for
  optimal control and estimation} {Automatic differentiation of rigid body
  dynamics for optimal control and estimation}.{\BBCQ}
\newblock
\APACjournalVolNumPages{Advanced Robotics}{31}{22}{1225-1237}.
\newblock

\newblock

\PrintBackRefs{\CurrentBib}

\bibitem [\protect \citeauthoryear {%
Gingold%
\ \BBA {} Monaghan%
}{%
Gingold%
\ \BBA {} Monaghan%
}{%
{\protect \APACyear {1977}}%
}]{%
monaghan1977sph}
\APACinsertmetastar {%
monaghan1977sph}%
\begin{APACrefauthors}%
Gingold, R.A.%
\BCBT {}\ \BBA {} Monaghan, J.J.%
\end{APACrefauthors}%
\unskip\
\newblock
\APACrefYearMonthDay{1977}{12}{}.
\newblock
{\BBOQ}\APACrefatitle {{Smoothed particle hydrodynamics: theory and application
  to non-spherical stars}} {{Smoothed particle hydrodynamics: theory and
  application to non-spherical stars}}.{\BBCQ}
\newblock
\APACjournalVolNumPages{Monthly Notices of the Royal Astronomical
  Society}{181}{3}{375-389}.
\newblock

\newblock

\PrintBackRefs{\CurrentBib}

\bibitem [\protect \citeauthoryear {%
Griewank%
\ \BBA {} Walther%
}{%
Griewank%
\ \BBA {} Walther%
}{%
{\protect \APACyear {2003}}%
}]{%
griewank_ad}
\APACinsertmetastar {%
griewank_ad}%
\begin{APACrefauthors}%
Griewank, A.%
\BCBT {}\ \BBA {} Walther, A.%
\end{APACrefauthors}%
\unskip\
\newblock
\APACrefYearMonthDay{2003}{}{}.
\newblock
{\BBOQ}\APACrefatitle {Introduction to Automatic Differentiation} {Introduction
  to automatic differentiation}.{\BBCQ}
\newblock
\APACjournalVolNumPages{PAMM}{2}{1}{45-49}.
\newblock

\newblock

\PrintBackRefs{\CurrentBib}

\bibitem [\protect \citeauthoryear {%
Griffith%
}{%
Griffith%
}{%
{\protect \APACyear {1921}}%
}]{%
griffith1921fracture}
\APACinsertmetastar {%
griffith1921fracture}%
\begin{APACrefauthors}%
Griffith, A.A.%
\end{APACrefauthors}%
\unskip\
\newblock
\APACrefYearMonthDay{1921}{}{}.
\newblock
{\BBOQ}\APACrefatitle {The phenomena of rupture and flow in solids} {The
  phenomena of rupture and flow in solids}.{\BBCQ}
\newblock
\APACjournalVolNumPages{Philosophical transactions of the royal society of
  London. Series A, containing papers of a mathematical or physical
  character}{221}{582-593}{163--198}.
\newblock

\newblock

\PrintBackRefs{\CurrentBib}

\bibitem [\protect \citeauthoryear {%
Hahn%
, Banzet%
, Bern%
\BCBL {}\ \BBA {} Coros%
}{%
Hahn%
\ \protect \BOthers {.}}{%
{\protect \APACyear {2019}}%
}]{%
hahn2019real2sim}
\APACinsertmetastar {%
hahn2019real2sim}%
\begin{APACrefauthors}%
Hahn, D.%
, Banzet, P.%
, Bern, J.M.%
\BCBL {} Coros, S.%
\end{APACrefauthors}%
\unskip\
\newblock
\APACrefYearMonthDay{2019}{}{}.
\newblock
{\BBOQ}\APACrefatitle {Real2sim: Visco-elastic parameter estimation from
  dynamic motion} {Real2sim: Visco-elastic parameter estimation from dynamic
  motion}.{\BBCQ}
\newblock
\APACjournalVolNumPages{ACM Transactions on Graphics (TOG)}{38}{6}{1--13}.
\newblock

\newblock

\PrintBackRefs{\CurrentBib}

\bibitem [\protect \citeauthoryear {%
Heiden%
, Denniston%
, Millard%
, Ramos%
\BCBL {}\ \BBA {} Sukhatme%
}{%
Heiden%
, Denniston%
\BCBL {}\ \protect \BOthers {.}}{%
{\protect \APACyear {2021}}%
}]{%
heiden2022pds}
\APACinsertmetastar {%
heiden2022pds}%
\begin{APACrefauthors}%
Heiden, E.%
, Denniston, C.E.%
, Millard, D.%
, Ramos, F.%
\BCBL {} Sukhatme, G.S.%
\end{APACrefauthors}%
\unskip\
\newblock
\APACrefYearMonthDay{2021}{}{}.
\newblock
{\BBOQ}\APACrefatitle {Probabilistic Inference of Simulation Parameters via
  Parallel Differentiable Simulation} {Probabilistic inference of simulation
  parameters via parallel differentiable simulation}.{\BBCQ}
\newblock
\APACjournalVolNumPages{CoRR}{abs/2109.08815}{}{}.
\newblock
\begin{APACrefURL} {https://arxiv.org/abs/2109.08815} \end{APACrefURL}
\newblock
{\href{https://arxiv.org/abs/2109.08815}{{2109.08815}}}
\newblock

\PrintBackRefs{\CurrentBib}

\bibitem [\protect \citeauthoryear {%
Heiden%
, Liu%
, Ramachandran%
\BCBL {}\ \BBA {} Sukhatme%
}{%
Heiden%
, Liu%
\BCBL {}\ \protect \BOthers {.}}{%
{\protect \APACyear {2020}}%
}]{%
heiden2020lidar}
\APACinsertmetastar {%
heiden2020lidar}%
\begin{APACrefauthors}%
Heiden, E.%
, Liu, Z.%
, Ramachandran, R.K.%
\BCBL {} Sukhatme, G.S.%
\end{APACrefauthors}%
\unskip\
\newblock
\APACrefYearMonthDay{2020}{}{}.
\newblock
{\BBOQ}\APACrefatitle {Physics-based Simulation of Continuous-Wave {LIDAR} for
  Localization, Calibration and Tracking} {Physics-based simulation of
  continuous-wave {LIDAR} for localization, calibration and tracking}.{\BBCQ}
\newblock
 \APACrefbtitle {International Conference on Robotics and Automation (ICRA).}
  {International conference on robotics and automation (icra).}
\PrintBackRefs{\CurrentBib}

\bibitem [\protect \citeauthoryear {%
Heiden%
, Macklin%
\BCBL {}\ \protect \BOthers {.}}{%
Heiden%
, Macklin%
\BCBL {}\ \protect \BOthers {.}}{%
{\protect \APACyear {2021}}%
}]{%
heiden2021disect}
\APACinsertmetastar {%
heiden2021disect}%
\begin{APACrefauthors}%
Heiden, E.%
, Macklin, M.%
, Narang, Y.S.%
, Fox, D.%
, Garg, A.%
\BCBL {} Ramos, F.%
\end{APACrefauthors}%
\unskip\
\newblock
\APACrefYearMonthDay{2021}{July}{}.
\newblock
{\BBOQ}\APACrefatitle {{DiSECt: A Differentiable Simulation Engine for
  Autonomous Robotic Cutting}} {{DiSECt: A Differentiable Simulation Engine for
  Autonomous Robotic Cutting}}.{\BBCQ}
\newblock
 \APACrefbtitle {Proceedings of Robotics: Science and Systems.} {Proceedings of
  robotics: Science and systems.}
\newblock
\APACaddressPublisher{Virtual}{}.
\newblock
\begin{APACrefDOI} \doi{10.15607/RSS.2021.XVII.067} \end{APACrefDOI}
\PrintBackRefs{\CurrentBib}

\bibitem [\protect \citeauthoryear {%
Heiden%
, Millard%
, Coumans%
, Sheng%
\BCBL {}\ \BBA {} Sukhatme%
}{%
Heiden%
, Millard%
\BCBL {}\ \protect \BOthers {.}}{%
{\protect \APACyear {2021}}%
}]{%
heiden2021neuralsim}
\APACinsertmetastar {%
heiden2021neuralsim}%
\begin{APACrefauthors}%
Heiden, E.%
, Millard, D.%
, Coumans, E.%
, Sheng, Y.%
\BCBL {} Sukhatme, G.S.%
\end{APACrefauthors}%
\unskip\
\newblock
\APACrefYearMonthDay{2021}{}{}.
\newblock
{\BBOQ}\APACrefatitle {Neural{S}im: Augmenting Differentiable Simulators with
  Neural Networks} {Neural{S}im: Augmenting differentiable simulators with
  neural networks}.{\BBCQ}
\newblock
 \APACrefbtitle {Proceedings of the IEEE International Conference on Robotics
  and Automation (ICRA).} {Proceedings of the ieee international conference on
  robotics and automation (icra).}
\newblock
\begin{APACrefURL}
  {https://github.com/google-research/tiny-differentiable-simulator}
  \end{APACrefURL}
\PrintBackRefs{\CurrentBib}

\bibitem [\protect \citeauthoryear {%
Heiden%
, Millard%
, Zhang%
\BCBL {}\ \BBA {} Sukhatme%
}{%
Heiden%
, Millard%
\BCBL {}\ \protect \BOthers {.}}{%
{\protect \APACyear {2020}}%
}]{%
heiden2019ids}
\APACinsertmetastar {%
heiden2019ids}%
\begin{APACrefauthors}%
Heiden, E.%
, Millard, D.%
, Zhang, H.%
\BCBL {} Sukhatme, G.S.%
\end{APACrefauthors}%
\unskip\
\newblock
\APACrefYearMonthDay{2020}{}{}.
\newblock
\APACrefbtitle {Interactive Differentiable Simulation.} {Interactive
  differentiable simulation.}
\PrintBackRefs{\CurrentBib}

\bibitem [\protect \citeauthoryear {%
Hsu%
\ \BBA {} Ramos%
}{%
Hsu%
\ \BBA {} Ramos%
}{%
{\protect \APACyear {2019}}%
}]{%
hsu2019likelihoodfree}
\APACinsertmetastar {%
hsu2019likelihoodfree}%
\begin{APACrefauthors}%
Hsu, K.%
\BCBT {}\ \BBA {} Ramos, F.%
\end{APACrefauthors}%
\unskip\
\newblock
\APACrefYearMonthDay{2019}{16--18 Apr}{}.
\newblock
{\BBOQ}\APACrefatitle {Bayesian Learning of Conditional Kernel Mean Embeddings
  for Automatic Likelihood-Free Inference} {Bayesian learning of conditional
  kernel mean embeddings for automatic likelihood-free inference}.{\BBCQ}
\newblock
 K.~Chaudhuri\ \BBA {} M.~Sugiyama\ (\BEDS), \APACrefbtitle {Proceedings of
  Machine Learning Research} {Proceedings of machine learning research}\
  (\BVOL~89, \BPGS\ 2631--2640).
\PrintBackRefs{\CurrentBib}

\bibitem [\protect \citeauthoryear {%
Hu%
\ \protect \BOthers {.}}{%
Hu%
\ \protect \BOthers {.}}{%
{\protect \APACyear {2020}}%
}]{%
hu2020difftaichi}
\APACinsertmetastar {%
hu2020difftaichi}%
\begin{APACrefauthors}%
Hu, Y.%
, Anderson, L.%
, Li, T\BHBI M.%
, Sun, Q.%
, Carr, N.%
, Ragan-Kelley, J.%
\BCBL {} Durand, F.%
\end{APACrefauthors}%
\unskip\
\newblock
\APACrefYearMonthDay{2020}{}{}.
\newblock
{\BBOQ}\APACrefatitle {Diff{T}aichi: Differentiable Programming for Physical
  Simulation} {Diff{T}aichi: Differentiable programming for physical
  simulation}.{\BBCQ}
\newblock
\APACjournalVolNumPages{ICLR}{}{}{}.
\newblock

\newblock

\PrintBackRefs{\CurrentBib}

\bibitem [\protect \citeauthoryear {%
Hu%
, Fang%
\BCBL {}\ \protect \BOthers {.}}{%
Hu%
, Fang%
\BCBL {}\ \protect \BOthers {.}}{%
{\protect \APACyear {2018}}%
}]{%
hu2018mlsmpmcpic}
\APACinsertmetastar {%
hu2018mlsmpmcpic}%
\begin{APACrefauthors}%
Hu, Y.%
, Fang, Y.%
, Ge, Z.%
, Qu, Z.%
, Zhu, Y.%
, Pradhana, A.%
\BCBL {} Jiang, C.%
\end{APACrefauthors}%
\unskip\
\newblock
\APACrefYearMonthDay{2018}{}{}.
\newblock
{\BBOQ}\APACrefatitle {A Moving Least Squares Material Point Method with
  Displacement Discontinuity and Two-Way Rigid Body Coupling} {A moving least
  squares material point method with displacement discontinuity and two-way
  rigid body coupling}.{\BBCQ}
\newblock
\APACjournalVolNumPages{ACM Transactions on Graphics (TOG)}{37}{4}{150}.
\newblock

\newblock

\PrintBackRefs{\CurrentBib}

\bibitem [\protect \citeauthoryear {%
Hu%
\ \protect \BOthers {.}}{%
Hu%
\ \protect \BOthers {.}}{%
{\protect \APACyear {2019}}%
}]{%
hu2019chainqueen}
\APACinsertmetastar {%
hu2019chainqueen}%
\begin{APACrefauthors}%
Hu, Y.%
, Liu, J.%
, Spielberg, A.%
, Tenenbaum, J.B.%
, Freeman, W.T.%
, Wu, J.%
\BDBL {}Matusik, W.%
\end{APACrefauthors}%
\unskip\
\newblock
\APACrefYearMonthDay{2019}{}{}.
\newblock
{\BBOQ}\APACrefatitle {Chain{Q}ueen: A Real-Time Differentiable Physical
  Simulator for Soft Robotics} {Chain{Q}ueen: A real-time differentiable
  physical simulator for soft robotics}.{\BBCQ}
\newblock
\APACjournalVolNumPages{Proceedings of IEEE International Conference on
  Robotics and Automation (ICRA)}{}{}{}.
\newblock

\newblock

\PrintBackRefs{\CurrentBib}

\bibitem [\protect \citeauthoryear {%
Hu%
, Zhou%
\BCBL {}\ \protect \BOthers {.}}{%
Hu%
, Zhou%
\BCBL {}\ \protect \BOthers {.}}{%
{\protect \APACyear {2018}}%
}]{%
hu2018tetwild}
\APACinsertmetastar {%
hu2018tetwild}%
\begin{APACrefauthors}%
Hu, Y.%
, Zhou, Q.%
, Gao, X.%
, Jacobson, A.%
, Zorin, D.%
\BCBL {} Panozzo, D.%
\end{APACrefauthors}%
\unskip\
\newblock
\APACrefYearMonthDay{2018}{{\APACmonth{07}}}{}.
\newblock
{\BBOQ}\APACrefatitle {Tetrahedral Meshing in the Wild} {Tetrahedral meshing in
  the wild}.{\BBCQ}
\newblock
\APACjournalVolNumPages{ACM Trans. Graph.}{37}{4}{60:1--60:14}.
\newblock

\newblock

\PrintBackRefs{\CurrentBib}

\bibitem [\protect \citeauthoryear {%
Hughes%
}{%
Hughes%
}{%
{\protect \APACyear {2012}}%
}]{%
hughes2012fem}
\APACinsertmetastar {%
hughes2012fem}%
\begin{APACrefauthors}%
Hughes, T.J.%
\end{APACrefauthors}%
\unskip\
\newblock
\APACrefYear{2012}.
\newblock
\APACrefbtitle {The finite element method: linear static and dynamic finite
  element analysis} {The finite element method: linear static and dynamic
  finite element analysis}.
\newblock
\APACaddressPublisher{}{Courier Corporation}.
\PrintBackRefs{\CurrentBib}

\bibitem [\protect \citeauthoryear {%
Innes%
}{%
Innes%
}{%
{\protect \APACyear {2019}}%
}]{%
innes2019zygote}
\APACinsertmetastar {%
innes2019zygote}%
\begin{APACrefauthors}%
Innes, M.%
\end{APACrefauthors}%
\unskip\
\newblock
\APACrefYearMonthDay{2019}{}{}.
\newblock
\APACrefbtitle {Don't Unroll Adjoint: Differentiating SSA-Form Programs.}
  {Don't unroll adjoint: Differentiating ssa-form programs.}
\PrintBackRefs{\CurrentBib}

\bibitem [\protect \citeauthoryear {%
Jamdagni%
\ \BBA {} Jia%
}{%
Jamdagni%
\ \BBA {} Jia%
}{%
{\protect \APACyear {2019}}%
}]{%
jamdagni2019robotic}
\APACinsertmetastar {%
jamdagni2019robotic}%
\begin{APACrefauthors}%
Jamdagni, P.%
\BCBT {}\ \BBA {} Jia, Y\BHBI B.%
\end{APACrefauthors}%
\unskip\
\newblock
\APACrefYearMonthDay{2019}{}{}.
\newblock
{\BBOQ}\APACrefatitle {Robotic cutting of solids based on fracture mechanics
  and FEM} {Robotic cutting of solids based on fracture mechanics and
  fem}.{\BBCQ}
\newblock
 \APACrefbtitle {IEEE International Conference on Intelligent Robots and
  Systems (IROS)} {Ieee international conference on intelligent robots and
  systems (iros)}\ (\BPGS\ 8246--8251).
\PrintBackRefs{\CurrentBib}

\bibitem [\protect \citeauthoryear {%
{Jeřábková}%
\ \BBA {} {Kuhlen}%
}{%
{Jeřábková}%
\ \BBA {} {Kuhlen}%
}{%
{\protect \APACyear {2009}}%
}]{%
jerabkova2009xfem}
\APACinsertmetastar {%
jerabkova2009xfem}%
\begin{APACrefauthors}%
{Jeřábková}, L.%
\BCBT {}\ \BBA {} {Kuhlen}, T.%
\end{APACrefauthors}%
\unskip\
\newblock
\APACrefYearMonthDay{2009}{}{}.
\newblock
{\BBOQ}\APACrefatitle {Stable Cutting of Deformable Objects in Virtual
  Environments Using {XFEM}} {Stable cutting of deformable objects in virtual
  environments using {XFEM}}.{\BBCQ}
\newblock
\APACjournalVolNumPages{IEEE Computer Graphics and Applications}{29}{2}{61-71}.
\newblock

\newblock

\PrintBackRefs{\CurrentBib}

\bibitem [\protect \citeauthoryear {%
Khoei%
}{%
Khoei%
}{%
{\protect \APACyear {2014}}%
}]{%
khoei2014extended}
\APACinsertmetastar {%
khoei2014extended}%
\begin{APACrefauthors}%
Khoei, A.R.%
\end{APACrefauthors}%
\unskip\
\newblock
\APACrefYear{2014}.
\newblock
\APACrefbtitle {Extended finite element method: theory and applications}
  {Extended finite element method: theory and applications}.
\newblock
\APACaddressPublisher{}{John Wiley \& Sons}.
\PrintBackRefs{\CurrentBib}

\bibitem [\protect \citeauthoryear {%
Kingma%
\ \BBA {} Ba%
}{%
Kingma%
\ \BBA {} Ba%
}{%
{\protect \APACyear {2015}}%
}]{%
kingma2014adam}
\APACinsertmetastar {%
kingma2014adam}%
\begin{APACrefauthors}%
Kingma, D.P.%
\BCBT {}\ \BBA {} Ba, J.%
\end{APACrefauthors}%
\unskip\
\newblock
\APACrefYearMonthDay{2015}{}{}.
\newblock
{\BBOQ}\APACrefatitle {Adam: A Method for Stochastic Optimization} {Adam: A
  method for stochastic optimization}.{\BBCQ}
\newblock
 \APACrefbtitle {3rd International Conference for Learning Representations
  (ICLR).} {3rd international conference for learning representations (iclr).}
\PrintBackRefs{\CurrentBib}

\bibitem [\protect \citeauthoryear {%
Koolen%
\ \BBA {} Deits%
}{%
Koolen%
\ \BBA {} Deits%
}{%
{\protect \APACyear {2019}}%
}]{%
koolen2019rbd-julia}
\APACinsertmetastar {%
koolen2019rbd-julia}%
\begin{APACrefauthors}%
Koolen, T.%
\BCBT {}\ \BBA {} Deits, R.%
\end{APACrefauthors}%
\unskip\
\newblock
\APACrefYearMonthDay{2019}{05}{}.
\newblock
{\BBOQ}\APACrefatitle {Julia for robotics: simulation and real-time control in
  a high-level programming language} {Julia for robotics: simulation and
  real-time control in a high-level programming language}.{\BBCQ}
\newblock
 \APACrefbtitle {International Conference on Robotics and Automation.}
  {International conference on robotics and automation.}
\PrintBackRefs{\CurrentBib}

\bibitem [\protect \citeauthoryear {%
Koschier%
, Bender%
\BCBL {}\ \BBA {} Thuerey%
}{%
Koschier%
\ \protect \BOthers {.}}{%
{\protect \APACyear {2017}}%
}]{%
koschier2017xfem}
\APACinsertmetastar {%
koschier2017xfem}%
\begin{APACrefauthors}%
Koschier, D.%
, Bender, J.%
\BCBL {} Thuerey, N.%
\end{APACrefauthors}%
\unskip\
\newblock
\APACrefYearMonthDay{2017}{{\APACmonth{07}}}{}.
\newblock
{\BBOQ}\APACrefatitle {Robust EXtended Finite Elements for Complex Cutting of
  Deformables} {Robust extended finite elements for complex cutting of
  deformables}.{\BBCQ}
\newblock
\APACjournalVolNumPages{ACM Trans. Graph.}{36}{4}{}.
\newblock

\newblock

\PrintBackRefs{\CurrentBib}

\bibitem [\protect \citeauthoryear {%
Koschier%
, Lipponer%
\BCBL {}\ \BBA {} Bender%
}{%
Koschier%
\ \protect \BOthers {.}}{%
{\protect \APACyear {2014}}%
}]{%
koschier2014adaptive}
\APACinsertmetastar {%
koschier2014adaptive}%
\begin{APACrefauthors}%
Koschier, D.%
, Lipponer, S.%
\BCBL {} Bender, J.%
\end{APACrefauthors}%
\unskip\
\newblock
\APACrefYearMonthDay{2014}{}{}.
\newblock
{\BBOQ}\APACrefatitle {Adaptive Tetrahedral Meshes for Brittle Fracture
  Simulation} {Adaptive tetrahedral meshes for brittle fracture
  simulation}.{\BBCQ}
\newblock
\APACjournalVolNumPages{Symposium on Computer Animation}{}{}{}.
\newblock

\newblock

\PrintBackRefs{\CurrentBib}

\bibitem [\protect \citeauthoryear {%
Krishnan%
\ \protect \BOthers {.}}{%
Krishnan%
\ \protect \BOthers {.}}{%
{\protect \APACyear {2018}}%
}]{%
swirl-ijrr18}
\APACinsertmetastar {%
swirl-ijrr18}%
\begin{APACrefauthors}%
Krishnan, S.%
, Garg, A.%
, Liaw, R.%
, Thananjeyan, B.%
, Miller, L.%
, Pokorny, F.T.%
\BCBL {} Goldberg, K.%
\end{APACrefauthors}%
\unskip\
\newblock
\APACrefYearMonthDay{2018}{jul}{}.
\newblock
{\BBOQ}\APACrefatitle {{SWIRL: A Sequential Windowed Inverse Reinforcement
  Learning Algorithm for Robot Tasks With Delayed Rewards}} {{SWIRL: A
  Sequential Windowed Inverse Reinforcement Learning Algorithm for Robot Tasks
  With Delayed Rewards}}.{\BBCQ}
\newblock
\APACjournalVolNumPages{International Journal of Robotics Research
  (IJRR)}{}{}{}.
\newblock

\newblock

\PrintBackRefs{\CurrentBib}

\bibitem [\protect \citeauthoryear {%
Kulkarni%
, Satyanarayana%
, Rohatgi%
\BCBL {}\ \BBA {} Vijayan%
}{%
Kulkarni%
\ \protect \BOthers {.}}{%
{\protect \APACyear {1983}}%
}]{%
kulkarni1983banana}
\APACinsertmetastar {%
kulkarni1983banana}%
\begin{APACrefauthors}%
Kulkarni, A.%
, Satyanarayana, K.%
, Rohatgi, P.%
\BCBL {} Vijayan, K.%
\end{APACrefauthors}%
\unskip\
\newblock
\APACrefYearMonthDay{1983}{}{}.
\newblock
{\BBOQ}\APACrefatitle {Mechanical properties of banana fibres (Musa sepientum)}
  {Mechanical properties of banana fibres (musa sepientum)}.{\BBCQ}
\newblock
\APACjournalVolNumPages{Journal of materials science}{18}{8}{2290--2296}.
\newblock

\newblock

\PrintBackRefs{\CurrentBib}

\bibitem [\protect \citeauthoryear {%
Lenz%
, Knepper%
\BCBL {}\ \BBA {} Saxena%
}{%
Lenz%
\ \protect \BOthers {.}}{%
{\protect \APACyear {2015}}%
}]{%
lenz2015deepmpc}
\APACinsertmetastar {%
lenz2015deepmpc}%
\begin{APACrefauthors}%
Lenz, I.%
, Knepper, R.A.%
\BCBL {} Saxena, A.%
\end{APACrefauthors}%
\unskip\
\newblock
\APACrefYearMonthDay{2015}{}{}.
\newblock
{\BBOQ}\APACrefatitle {{DeepMPC}: Learning deep latent features for model
  predictive control} {{DeepMPC}: Learning deep latent features for model
  predictive control}.{\BBCQ}
\newblock
\APACjournalVolNumPages{Robotics: Science and Systems}{}{}{}.
\newblock

\newblock

\PrintBackRefs{\CurrentBib}

\bibitem [\protect \citeauthoryear {%
C.~Li%
, Chen%
, Carlson%
\BCBL {}\ \BBA {} Carin%
}{%
C.~Li%
\ \protect \BOthers {.}}{%
{\protect \APACyear {2016}}%
}]{%
Li2016pSGLD}
\APACinsertmetastar {%
Li2016pSGLD}%
\begin{APACrefauthors}%
Li, C.%
, Chen, C.%
, Carlson, D.%
\BCBL {} Carin, L.%
\end{APACrefauthors}%
\unskip\
\newblock
\APACrefYearMonthDay{2016}{}{}.
\newblock
{\BBOQ}\APACrefatitle {Preconditioned stochastic gradient Langevin dynamics for
  deep neural networks} {Preconditioned stochastic gradient langevin dynamics
  for deep neural networks}.{\BBCQ}
\newblock
 \APACrefbtitle {AAAI.} {Aaai.}
\PrintBackRefs{\CurrentBib}

\bibitem [\protect \citeauthoryear {%
Y.~Li%
, Du%
, Wang%
\BCBL {}\ \BBA {} Lei%
}{%
Y.~Li%
\ \protect \BOthers {.}}{%
{\protect \APACyear {2018/02}}%
}]{%
li2018apple}
\APACinsertmetastar {%
li2018apple}%
\begin{APACrefauthors}%
Li, Y.%
, Du, X.%
, Wang, J.%
\BCBL {} Lei, C.%
\end{APACrefauthors}%
\unskip\
\newblock
\APACrefYearMonthDay{2018/02}{}{}.
\newblock
{\BBOQ}\APACrefatitle {Study on mechanical properties of apple picking damage}
  {Study on mechanical properties of apple picking damage}.{\BBCQ}
\newblock
 \APACrefbtitle {Proceedings of the 2017 3rd International Forum on Energy,
  Environment Science and Materials (IFEESM 2017)} {Proceedings of the 2017 3rd
  international forum on energy, environment science and materials (ifeesm
  2017)}\ (\BPG~1666-1670).
\newblock
\APACaddressPublisher{}{Atlantis Press}.
\PrintBackRefs{\CurrentBib}

\bibitem [\protect \citeauthoryear {%
Liu%
\ \BBA {} Liu%
}{%
Liu%
\ \BBA {} Liu%
}{%
{\protect \APACyear {2010}}%
}]{%
liu2010sph}
\APACinsertmetastar {%
liu2010sph}%
\begin{APACrefauthors}%
Liu, M.%
\BCBT {}\ \BBA {} Liu, G.%
\end{APACrefauthors}%
\unskip\
\newblock
\APACrefYearMonthDay{2010}{}{}.
\newblock
{\BBOQ}\APACrefatitle {Smoothed particle hydrodynamics (SPH): An overview and
  recent developments} {Smoothed particle hydrodynamics (sph): An overview and
  recent developments}.{\BBCQ}
\newblock
\APACjournalVolNumPages{Archives of computational methods in
  engineering}{17}{}{25--76}.
\newblock

\newblock

\PrintBackRefs{\CurrentBib}

\bibitem [\protect \citeauthoryear {%
Ljung%
}{%
Ljung%
}{%
{\protect \APACyear {1999}}%
}]{%
ljung1999sysid}
\APACinsertmetastar {%
ljung1999sysid}%
\begin{APACrefauthors}%
Ljung, L.%
\end{APACrefauthors}%
\unskip\
\newblock
\APACrefYear{1999}.
\newblock
\APACrefbtitle {System Identification (2nd Ed.): Theory for the User} {System
  identification (2nd ed.): Theory for the user}.
\newblock
\APACaddressPublisher{USA}{Prentice Hall PTR}.
\PrintBackRefs{\CurrentBib}

\bibitem [\protect \citeauthoryear {%
Long%
, Moughlbay%
, Khalil%
\BCBL {}\ \BBA {} Martinet%
}{%
Long%
\ \protect \BOthers {.}}{%
{\protect \APACyear {2013}}%
}]{%
long2013robotic}
\APACinsertmetastar {%
long2013robotic}%
\begin{APACrefauthors}%
Long, P.%
, Moughlbay, A.%
, Khalil, W.%
\BCBL {} Martinet, P.%
\end{APACrefauthors}%
\unskip\
\newblock
\APACrefYearMonthDay{2013}{}{}.
\newblock
{\BBOQ}\APACrefatitle {Robotic meat cutting} {Robotic meat cutting}.{\BBCQ}
\newblock
 \APACrefbtitle {ICT-PAMM Workshop.} {Ict-pamm workshop.}
\PrintBackRefs{\CurrentBib}

\bibitem [\protect \citeauthoryear {%
Macklin%
\ \protect \BOthers {.}}{%
Macklin%
\ \protect \BOthers {.}}{%
{\protect \APACyear {2020}}%
}]{%
macklin2020sdf}
\APACinsertmetastar {%
macklin2020sdf}%
\begin{APACrefauthors}%
Macklin, M.%
, Erleben, K.%
, M\"{u}ller, M.%
, Chentanez, N.%
, Jeschke, S.%
\BCBL {} Corse, Z.%
\end{APACrefauthors}%
\unskip\
\newblock
\APACrefYearMonthDay{2020}{{\APACmonth{04}}}{}.
\newblock
{\BBOQ}\APACrefatitle {Local Optimization for Robust Signed Distance Field
  Collision} {Local optimization for robust signed distance field
  collision}.{\BBCQ}
\newblock
\APACjournalVolNumPages{Proc. ACM Comput. Graph. Interact. Tech.}{3}{1}{}.
\newblock

\newblock

\PrintBackRefs{\CurrentBib}

\bibitem [\protect \citeauthoryear {%
Mahnken%
}{%
Mahnken%
}{%
{\protect \APACyear {2017}}%
}]{%
mahnken2017identification}
\APACinsertmetastar {%
mahnken2017identification}%
\begin{APACrefauthors}%
Mahnken, R.%
\end{APACrefauthors}%
\unskip\
\newblock
\APACrefYearMonthDay{2017}{}{}.
\newblock
{\BBOQ}\APACrefatitle {Identification of material parameters for constitutive
  equations} {Identification of material parameters for constitutive
  equations}.{\BBCQ}
\newblock
\APACjournalVolNumPages{Encyclopedia of Computational Mechanics Second
  Edition}{}{}{1--21}.
\newblock

\newblock

\PrintBackRefs{\CurrentBib}

\bibitem [\protect \citeauthoryear {%
Margossian%
}{%
Margossian%
}{%
{\protect \APACyear {2019}}%
}]{%
Margossian_2019}
\APACinsertmetastar {%
Margossian_2019}%
\begin{APACrefauthors}%
Margossian, C.C.%
\end{APACrefauthors}%
\unskip\
\newblock
\APACrefYearMonthDay{2019}{Mar}{}.
\newblock
{\BBOQ}\APACrefatitle {A review of automatic differentiation and its efficient
  implementation} {A review of automatic differentiation and its efficient
  implementation}.{\BBCQ}
\newblock
\APACjournalVolNumPages{WIREs Data Mining and Knowledge Discovery}{9}{4}{}.
\newblock

\newblock

\PrintBackRefs{\CurrentBib}

\bibitem [\protect \citeauthoryear {%
{Matl}%
, {Narang}%
, {Bajcsy}%
, {Ramos}%
\BCBL {}\ \BBA {} {Fox}%
}{%
{Matl}%
\ \protect \BOthers {.}}{%
{\protect \APACyear {2020}}%
}]{%
matl2020granular}
\APACinsertmetastar {%
matl2020granular}%
\begin{APACrefauthors}%
{Matl}, C.%
, {Narang}, Y.%
, {Bajcsy}, R.%
, {Ramos}, F.%
\BCBL {} {Fox}, D.%
\end{APACrefauthors}%
\unskip\
\newblock
\APACrefYearMonthDay{2020}{}{}.
\newblock
{\BBOQ}\APACrefatitle {Inferring the Material Properties of Granular Media for
  Robotic Tasks} {Inferring the material properties of granular media for
  robotic tasks}.{\BBCQ}
\newblock
 \APACrefbtitle {2020 IEEE International Conference on Robotics and Automation
  (ICRA)} {2020 ieee international conference on robotics and automation
  (icra)}\ (\BPG~2770-2777).
\PrintBackRefs{\CurrentBib}

\bibitem [\protect \citeauthoryear {%
Matl%
, Narang%
, Fox%
, Bajcsy%
\BCBL {}\ \BBA {} Ramos%
}{%
Matl%
\ \protect \BOthers {.}}{%
{\protect \APACyear {2020}}%
}]{%
matl2020stressd}
\APACinsertmetastar {%
matl2020stressd}%
\begin{APACrefauthors}%
Matl, C.%
, Narang, Y.S.%
, Fox, D.%
, Bajcsy, R.%
\BCBL {} Ramos, F.%
\end{APACrefauthors}%
\unskip\
\newblock
\APACrefYearMonthDay{2020}{}{}.
\newblock
{\BBOQ}\APACrefatitle {{STReSSD}: Sim-To-Real from Sound for Stochastic
  Dynamics} {{STReSSD}: Sim-to-real from sound for stochastic dynamics}.{\BBCQ}
\newblock
\APACjournalVolNumPages{Conference on Robot Learning}{}{}{}.
\newblock

\newblock

\PrintBackRefs{\CurrentBib}

\bibitem [\protect \citeauthoryear {%
Mehta%
, Diaz%
, Golemo%
, Pal%
\BCBL {}\ \BBA {} Paull%
}{%
Mehta%
\ \protect \BOthers {.}}{%
{\protect \APACyear {2019}}%
}]{%
mehta2019adr}
\APACinsertmetastar {%
mehta2019adr}%
\begin{APACrefauthors}%
Mehta, B.%
, Diaz, M.%
, Golemo, F.%
, Pal, C.%
\BCBL {} Paull, L.%
\end{APACrefauthors}%
\unskip\
\newblock
\APACrefYearMonthDay{2019}{}{}.
\newblock
{\BBOQ}\APACrefatitle {Active Domain Randomization} {Active domain
  randomization}.{\BBCQ}
\newblock
\APACjournalVolNumPages{Conference on Robot Learning}{}{}{}.
\newblock

\newblock

\PrintBackRefs{\CurrentBib}

\bibitem [\protect \citeauthoryear {%
Mehta%
, Handa%
, Fox%
\BCBL {}\ \BBA {} Ramos%
}{%
Mehta%
\ \protect \BOthers {.}}{%
{\protect \APACyear {2020}}%
}]{%
mehta2020user}
\APACinsertmetastar {%
mehta2020user}%
\begin{APACrefauthors}%
Mehta, B.%
, Handa, A.%
, Fox, D.%
\BCBL {} Ramos, F.%
\end{APACrefauthors}%
\unskip\
\newblock
\APACrefYearMonthDay{2020}{}{}.
\newblock
{\BBOQ}\APACrefatitle {A User's Guide to Calibrating Robotics Simulators} {A
  user's guide to calibrating robotics simulators}.{\BBCQ}
\newblock
\APACjournalVolNumPages{Conference on Robot Learning}{}{}{}.
\newblock

\newblock

\PrintBackRefs{\CurrentBib}

\bibitem [\protect \citeauthoryear {%
Merchant%
}{%
Merchant%
}{%
{\protect \APACyear {1945}}%
}]{%
merchant1945mechanics}
\APACinsertmetastar {%
merchant1945mechanics}%
\begin{APACrefauthors}%
Merchant, M.E.%
\end{APACrefauthors}%
\unskip\
\newblock
\APACrefYearMonthDay{1945}{}{}.
\newblock
{\BBOQ}\APACrefatitle {Mechanics of the metal cutting process. I. Orthogonal
  cutting and a type 2 chip} {Mechanics of the metal cutting process. i.
  orthogonal cutting and a type 2 chip}.{\BBCQ}
\newblock
\APACjournalVolNumPages{Journal of applied physics}{16}{5}{267--275}.
\newblock

\newblock

\PrintBackRefs{\CurrentBib}

\bibitem [\protect \citeauthoryear {%
{Mitsioni}%
, {Karayiannidis}%
, {Stork}%
\BCBL {}\ \BBA {} {Kragic}%
}{%
{Mitsioni}%
\ \protect \BOthers {.}}{%
{\protect \APACyear {2019}}%
}]{%
mitsioni2019mpc}
\APACinsertmetastar {%
mitsioni2019mpc}%
\begin{APACrefauthors}%
{Mitsioni}, I.%
, {Karayiannidis}, Y.%
, {Stork}, J.A.%
\BCBL {} {Kragic}, D.%
\end{APACrefauthors}%
\unskip\
\newblock
\APACrefYearMonthDay{2019}{}{}.
\newblock
{\BBOQ}\APACrefatitle {Data-Driven Model Predictive Control for the
  Contact-Rich Task of Food Cutting} {Data-driven model predictive control for
  the contact-rich task of food cutting}.{\BBCQ}
\newblock
 \APACrefbtitle {IEEE-RAS International Conference on Humanoid Robots
  (Humanoids)} {Ieee-ras international conference on humanoid robots
  (humanoids)}\ (\BPG~244-250).
\PrintBackRefs{\CurrentBib}

\bibitem [\protect \citeauthoryear {%
Mo{\"e}s%
, Dolbow%
\BCBL {}\ \BBA {} Belytschko%
}{%
Mo{\"e}s%
\ \protect \BOthers {.}}{%
{\protect \APACyear {1999}}%
}]{%
moes1999xfem}
\APACinsertmetastar {%
moes1999xfem}%
\begin{APACrefauthors}%
Mo{\"e}s, N.%
, Dolbow, J.%
\BCBL {} Belytschko, T.%
\end{APACrefauthors}%
\unskip\
\newblock
\APACrefYearMonthDay{1999}{}{}.
\newblock
{\BBOQ}\APACrefatitle {A finite element method for crack growth without
  remeshing} {A finite element method for crack growth without
  remeshing}.{\BBCQ}
\newblock
\APACjournalVolNumPages{International journal for numerical methods in
  engineering}{46}{1}{131--150}.
\newblock

\newblock

\PrintBackRefs{\CurrentBib}

\bibitem [\protect \citeauthoryear {%
Molino%
, Bao%
\BCBL {}\ \BBA {} Fedkiw%
}{%
Molino%
\ \protect \BOthers {.}}{%
{\protect \APACyear {2004}}%
}]{%
molino2004vna}
\APACinsertmetastar {%
molino2004vna}%
\begin{APACrefauthors}%
Molino, N.%
, Bao, Z.%
\BCBL {} Fedkiw, R.%
\end{APACrefauthors}%
\unskip\
\newblock
\APACrefYearMonthDay{2004}{{\APACmonth{08}}}{}.
\newblock
{\BBOQ}\APACrefatitle {A Virtual Node Algorithm for Changing Mesh Topology
  during Simulation} {A virtual node algorithm for changing mesh topology
  during simulation}.{\BBCQ}
\newblock
\APACjournalVolNumPages{ACM Trans. Graph.}{23}{3}{385–392}.
\newblock

\newblock

\PrintBackRefs{\CurrentBib}

\bibitem [\protect \citeauthoryear {%
Monaghan%
}{%
Monaghan%
}{%
{\protect \APACyear {1992}}%
}]{%
monaghan1992sph}
\APACinsertmetastar {%
monaghan1992sph}%
\begin{APACrefauthors}%
Monaghan, J.%
\end{APACrefauthors}%
\unskip\
\newblock
\APACrefYearMonthDay{1992}{}{}.
\newblock
{\BBOQ}\APACrefatitle {Smoothed particle hydrodynamics} {Smoothed particle
  hydrodynamics}.{\BBCQ}
\newblock
\APACjournalVolNumPages{Annual Review of Astronomy and
  Astrophysics}{30}{}{543--574}.
\newblock

\newblock

\PrintBackRefs{\CurrentBib}

\bibitem [\protect \citeauthoryear {%
Mousavizadeh%
, Mashayekhi%
, Daraei~Garmakhany%
, Ehtesham~Nia%
\BCBL {}\ \BBA {} M%
}{%
Mousavizadeh%
\ \protect \BOthers {.}}{%
{\protect \APACyear {2010}}%
}]{%
mousavizadeh2010cucumber}
\APACinsertmetastar {%
mousavizadeh2010cucumber}%
\begin{APACrefauthors}%
Mousavizadeh, S.J.%
, Mashayekhi, A.N.%
, Daraei~Garmakhany, A.%
, Ehtesham~Nia, A.%
\BCBL {} M, J.%
\end{APACrefauthors}%
\unskip\
\newblock
\APACrefYearMonthDay{2010}{01}{}.
\newblock
{\BBOQ}\APACrefatitle {Evaluation of Some Physical Properties of Cucumber
  (Cucumis sativus L.).} {Evaluation of some physical properties of cucumber
  (cucumis sativus l.).}{\BBCQ}
\newblock
\APACjournalVolNumPages{Journal of Agricultural Science and
  Technology}{4}{}{107-115}.
\newblock

\newblock

\PrintBackRefs{\CurrentBib}

\bibitem [\protect \citeauthoryear {%
Mozifian%
, Higuera%
, Meger%
\BCBL {}\ \BBA {} Dudek%
}{%
Mozifian%
\ \protect \BOthers {.}}{%
{\protect \APACyear {2019}}%
}]{%
mozifian2019learning}
\APACinsertmetastar {%
mozifian2019learning}%
\begin{APACrefauthors}%
Mozifian, M.%
, Higuera, J.C.G.%
, Meger, D.%
\BCBL {} Dudek, G.%
\end{APACrefauthors}%
\unskip\
\newblock
\APACrefYearMonthDay{2019}{}{}.
\newblock
{\BBOQ}\APACrefatitle {Learning Domain Randomization Distributions for Training
  Robust Locomotion Policies} {Learning domain randomization distributions for
  training robust locomotion policies}.{\BBCQ}
\newblock
\APACjournalVolNumPages{arXiv preprint arXiv:1906.00410}{}{}{}.
\newblock

\newblock

\PrintBackRefs{\CurrentBib}

\bibitem [\protect \citeauthoryear {%
M.~Sánchez-Banderas%
\ \BBA {} A.~Otaduy%
}{%
M.~Sánchez-Banderas%
\ \BBA {} A.~Otaduy%
}{%
{\protect \APACyear {2018}}%
}]{%
banderas2018}
\APACinsertmetastar {%
banderas2018}%
\begin{APACrefauthors}%
M.~Sánchez-Banderas, R.%
\BCBT {}\ \BBA {} A.~Otaduy, M.%
\end{APACrefauthors}%
\unskip\
\newblock
\APACrefYearMonthDay{2018}{}{}.
\newblock
{\BBOQ}\APACrefatitle {Strain Rate Dissipation for Elastic Deformations}
  {Strain rate dissipation for elastic deformations}.{\BBCQ}
\newblock
\APACjournalVolNumPages{Computer Graphics Forum}{37}{8}{161-170}.
\newblock

\newblock

\PrintBackRefs{\CurrentBib}

\bibitem [\protect \citeauthoryear {%
Mu%
, Xue%
\BCBL {}\ \BBA {} Jia%
}{%
Mu%
\ \protect \BOthers {.}}{%
{\protect \APACyear {2019}}%
}]{%
mu2019robotic}
\APACinsertmetastar {%
mu2019robotic}%
\begin{APACrefauthors}%
Mu, X.%
, Xue, Y.%
\BCBL {} Jia, Y\BHBI B.%
\end{APACrefauthors}%
\unskip\
\newblock
\APACrefYearMonthDay{2019}{}{}.
\newblock
{\BBOQ}\APACrefatitle {Robotic cutting: Mechanics and control of knife motion}
  {Robotic cutting: Mechanics and control of knife motion}.{\BBCQ}
\newblock
 \APACrefbtitle {IEEE International Conference on Robotics and Automation
  (ICRA)} {Ieee international conference on robotics and automation (icra)}\
  (\BPGS\ 3066--3072).
\PrintBackRefs{\CurrentBib}

\bibitem [\protect \citeauthoryear {%
M{\"u}ller%
, Heidelberger%
, Hennix%
\BCBL {}\ \BBA {} Ratcliff%
}{%
M{\"u}ller%
\ \protect \BOthers {.}}{%
{\protect \APACyear {2007}}%
}]{%
muller2007position}
\APACinsertmetastar {%
muller2007position}%
\begin{APACrefauthors}%
M{\"u}ller, M.%
, Heidelberger, B.%
, Hennix, M.%
\BCBL {} Ratcliff, J.%
\end{APACrefauthors}%
\unskip\
\newblock
\APACrefYearMonthDay{2007}{}{}.
\newblock
{\BBOQ}\APACrefatitle {Position based dynamics} {Position based
  dynamics}.{\BBCQ}
\newblock
\APACjournalVolNumPages{Journal of Visual Communication and Image
  Representation}{18}{2}{109--118}.
\newblock

\newblock

\PrintBackRefs{\CurrentBib}

\bibitem [\protect \citeauthoryear {%
Murthy%
\ \protect \BOthers {.}}{%
Murthy%
\ \protect \BOthers {.}}{%
{\protect \APACyear {2021}}%
}]{%
murthy2021gradsim}
\APACinsertmetastar {%
murthy2021gradsim}%
\begin{APACrefauthors}%
Murthy, J.K.%
, Macklin, M.%
, Golemo, F.%
, Voleti, V.%
, Petrini, L.%
, Weiss, M.%
\BDBL {}Fidler, S.%
\end{APACrefauthors}%
\unskip\
\newblock
\APACrefYearMonthDay{2021}{}{}.
\newblock
{\BBOQ}\APACrefatitle {grad{S}im: Differentiable simulation for system
  identification and visuomotor control} {grad{S}im: Differentiable simulation
  for system identification and visuomotor control}.{\BBCQ}
\newblock
 \APACrefbtitle {International Conference on Learning Representations.}
  {International conference on learning representations.}
\PrintBackRefs{\CurrentBib}

\bibitem [\protect \citeauthoryear {%
Narang%
, Sundaralingam%
, Macklin%
, Mousavian%
\BCBL {}\ \BBA {} Fox%
}{%
Narang%
\ \protect \BOthers {.}}{%
{\protect \APACyear {2021}}%
}]{%
narang2021tactile}
\APACinsertmetastar {%
narang2021tactile}%
\begin{APACrefauthors}%
Narang, Y.%
, Sundaralingam, B.%
, Macklin, M.%
, Mousavian, A.%
\BCBL {} Fox, D.%
\end{APACrefauthors}%
\unskip\
\newblock
\APACrefYearMonthDay{2021}{}{}.
\newblock
{\BBOQ}\APACrefatitle {Sim-to-Real for Robotic Tactile Sensing via
  Physics-Based Simulation and Learned Latent Projections} {Sim-to-real for
  robotic tactile sensing via physics-based simulation and learned latent
  projections}.{\BBCQ}
\newblock
\APACjournalVolNumPages{International Conference on Robotics and
  Automation}{}{}{}.
\newblock

\newblock

\PrintBackRefs{\CurrentBib}

\bibitem [\protect \citeauthoryear {%
Pan%
, Bai%
, Zhao%
, Hao%
\BCBL {}\ \BBA {} Qin%
}{%
Pan%
\ \protect \BOthers {.}}{%
{\protect \APACyear {2015}}%
}]{%
pan2015real}
\APACinsertmetastar {%
pan2015real}%
\begin{APACrefauthors}%
Pan, J.%
, Bai, J.%
, Zhao, X.%
, Hao, A.%
\BCBL {} Qin, H.%
\end{APACrefauthors}%
\unskip\
\newblock
\APACrefYearMonthDay{2015}{}{}.
\newblock
{\BBOQ}\APACrefatitle {Real-time haptic manipulation and cutting of hybrid soft
  tissue models by extended position-based dynamics} {Real-time haptic
  manipulation and cutting of hybrid soft tissue models by extended
  position-based dynamics}.{\BBCQ}
\newblock
\APACjournalVolNumPages{Computer Animation and Virtual
  Worlds}{26}{3-4}{321--335}.
\newblock

\newblock

\PrintBackRefs{\CurrentBib}

\bibitem [\protect \citeauthoryear {%
Papamakarios%
\ \BBA {} Murray%
}{%
Papamakarios%
\ \BBA {} Murray%
}{%
{\protect \APACyear {2016}}%
}]{%
papamakarios2016epsilon}
\APACinsertmetastar {%
papamakarios2016epsilon}%
\begin{APACrefauthors}%
Papamakarios, G.%
\BCBT {}\ \BBA {} Murray, I.%
\end{APACrefauthors}%
\unskip\
\newblock
\APACrefYearMonthDay{2016}{}{}.
\newblock
{\BBOQ}\APACrefatitle {Fast $\epsilon$-free Inference of Simulation Models with
  {B}ayesian Conditional Density Estimation} {Fast $\epsilon$-free inference of
  simulation models with {B}ayesian conditional density estimation}.{\BBCQ}
\newblock
 D.~Lee, M.~Sugiyama, U.~Luxburg, I.~Guyon\BCBL {}\ \BBA {} R.~Garnett\
  (\BEDS), \APACrefbtitle {Advances in Neural Information Processing Systems}
  {Advances in neural information processing systems}\ (\BVOL~29).
\newblock
\APACaddressPublisher{}{Curran Associates, Inc.}
\PrintBackRefs{\CurrentBib}

\bibitem [\protect \citeauthoryear {%
Paszke%
\ \protect \BOthers {.}}{%
Paszke%
\ \protect \BOthers {.}}{%
{\protect \APACyear {2019}}%
}]{%
paszke2019pytorch}
\APACinsertmetastar {%
paszke2019pytorch}%
\begin{APACrefauthors}%
Paszke, A.%
, Gross, S.%
, Massa, F.%
, Lerer, A.%
, Bradbury, J.%
, Chanan, G.%
\BDBL {}Chintala, S.%
\end{APACrefauthors}%
\unskip\
\newblock
\APACrefYearMonthDay{2019}{}{}.
\newblock
{\BBOQ}\APACrefatitle {PyTorch: An Imperative Style, High-Performance Deep
  Learning Library} {Pytorch: An imperative style, high-performance deep
  learning library}.{\BBCQ}
\newblock
 H.~Wallach, H.~Larochelle, A.~Beygelzimer, F.~d\textquotesingle Alch\'{e}-Buc,
  E.~Fox\BCBL {}\ \BBA {} R.~Garnett\ (\BEDS), \APACrefbtitle {Advances in
  Neural Information Processing Systems 32} {Advances in neural information
  processing systems 32}\ (\BPGS\ 8024--8035).
\newblock
\APACaddressPublisher{}{Curran Associates, Inc.}
\newblock
\begin{APACrefURL}
  {http://papers.neurips.cc/paper/9015-pytorch-an-imperative-style-high-performance-deep-learning-library.pdf}
  \end{APACrefURL}
\PrintBackRefs{\CurrentBib}

\bibitem [\protect \citeauthoryear {%
Paulus%
, Untereiner%
, Courtecuisse%
, Cotin%
\BCBL {}\ \BBA {} Cazier%
}{%
Paulus%
\ \protect \BOthers {.}}{%
{\protect \APACyear {2015}}%
}]{%
paulus2015virtual}
\APACinsertmetastar {%
paulus2015virtual}%
\begin{APACrefauthors}%
Paulus, C.J.%
, Untereiner, L.%
, Courtecuisse, H.%
, Cotin, S.%
\BCBL {} Cazier, D.%
\end{APACrefauthors}%
\unskip\
\newblock
\APACrefYearMonthDay{2015}{}{}.
\newblock
{\BBOQ}\APACrefatitle {Virtual cutting of deformable objects based on efficient
  topological operations} {Virtual cutting of deformable objects based on
  efficient topological operations}.{\BBCQ}
\newblock
\APACjournalVolNumPages{The Visual Computer}{31}{6}{831--841}.
\newblock

\newblock

\PrintBackRefs{\CurrentBib}

\bibitem [\protect \citeauthoryear {%
Peyr{\'e}%
\ \BBA {} Cuturi%
}{%
Peyr{\'e}%
\ \BBA {} Cuturi%
}{%
{\protect \APACyear {2019}}%
}]{%
peyre2019ot}
\APACinsertmetastar {%
peyre2019ot}%
\begin{APACrefauthors}%
Peyr{\'e}, G.%
\BCBT {}\ \BBA {} Cuturi, M.%
\end{APACrefauthors}%
\unskip\
\newblock
\APACrefYear{2019}.
\newblock
\APACrefbtitle {Computational Optimal Transport: With Applications to Data
  Science} {Computational optimal transport: With applications to data
  science}.
\newblock
\APACaddressPublisher{}{Now Publishers}.
\PrintBackRefs{\CurrentBib}

\bibitem [\protect \citeauthoryear {%
Platt%
\ \BBA {} Barr%
}{%
Platt%
\ \BBA {} Barr%
}{%
{\protect \APACyear {1988}}%
}]{%
platt1988mdmm}
\APACinsertmetastar {%
platt1988mdmm}%
\begin{APACrefauthors}%
Platt, J.C.%
\BCBT {}\ \BBA {} Barr, A.H.%
\end{APACrefauthors}%
\unskip\
\newblock
\APACrefYearMonthDay{1988}{}{}.
\newblock
{\BBOQ}\APACrefatitle {Constrained differential optimization for neural
  networks} {Constrained differential optimization for neural networks}.{\BBCQ}
\newblock
\APACjournalVolNumPages{Advances in Neural Information Processing
  Systems}{}{}{}.
\newblock

\newblock

\PrintBackRefs{\CurrentBib}

\bibitem [\protect \citeauthoryear {%
Qiao%
, Liang%
, Koltun%
\BCBL {}\ \BBA {} Lin%
}{%
Qiao%
\ \protect \BOthers {.}}{%
{\protect \APACyear {2020}}%
}]{%
qiao2020scalable}
\APACinsertmetastar {%
qiao2020scalable}%
\begin{APACrefauthors}%
Qiao, Y\BHBI L.%
, Liang, J.%
, Koltun, V.%
\BCBL {} Lin, M.C.%
\end{APACrefauthors}%
\unskip\
\newblock
\APACrefYearMonthDay{2020}{}{}.
\newblock
{\BBOQ}\APACrefatitle {Scalable Differentiable Physics for Learning and
  Control} {Scalable differentiable physics for learning and control}.{\BBCQ}
\newblock
 \APACrefbtitle {ICML.} {Icml.}
\PrintBackRefs{\CurrentBib}

\bibitem [\protect \citeauthoryear {%
Ramos%
, Possas%
\BCBL {}\ \BBA {} Fox%
}{%
Ramos%
\ \protect \BOthers {.}}{%
{\protect \APACyear {2019}}%
}]{%
ramos2019bayessim}
\APACinsertmetastar {%
ramos2019bayessim}%
\begin{APACrefauthors}%
Ramos, F.%
, Possas, R.%
\BCBL {} Fox, D.%
\end{APACrefauthors}%
\unskip\
\newblock
\APACrefYearMonthDay{2019}{}{}.
\newblock
{\BBOQ}\APACrefatitle {Bayes{S}im: adaptive domain randomization via
  probabilistic inference for robotics simulators} {Bayes{S}im: adaptive domain
  randomization via probabilistic inference for robotics simulators}.{\BBCQ}
\newblock
 \APACrefbtitle {Robotics: Science and Systems (RSS).} {Robotics: Science and
  systems (rss).}
\PrintBackRefs{\CurrentBib}

\bibitem [\protect \citeauthoryear {%
Rubner%
, Tomasi%
\BCBL {}\ \BBA {} Guibas%
}{%
Rubner%
\ \protect \BOthers {.}}{%
{\protect \APACyear {2000}}%
}]{%
Rubner2000EMD}
\APACinsertmetastar {%
Rubner2000EMD}%
\begin{APACrefauthors}%
Rubner, Y.%
, Tomasi, C.%
\BCBL {} Guibas, L.J.%
\end{APACrefauthors}%
\unskip\
\newblock
\APACrefYearMonthDay{2000}{}{}.
\newblock
{\BBOQ}\APACrefatitle {The Earth Mover's Distance as a Metric for Image
  Retrieval.} {The earth mover's distance as a metric for image
  retrieval.}{\BBCQ}
\newblock
\APACjournalVolNumPages{International Journal of Computer
  Vision}{40}{2}{99-121}.
\newblock

\newblock

\PrintBackRefs{\CurrentBib}

\bibitem [\protect \citeauthoryear {%
Sifakis%
, Der%
\BCBL {}\ \BBA {} Fedkiw%
}{%
Sifakis%
, Der%
\BCBL {}\ \BBA {} Fedkiw%
}{%
{\protect \APACyear {2007}}%
}]{%
sifakis2007arbitrary}
\APACinsertmetastar {%
sifakis2007arbitrary}%
\begin{APACrefauthors}%
Sifakis, E.%
, Der, K.G.%
\BCBL {} Fedkiw, R.%
\end{APACrefauthors}%
\unskip\
\newblock
\APACrefYearMonthDay{2007}{}{}.
\newblock
{\BBOQ}\APACrefatitle {Arbitrary cutting of deformable tetrahedralized objects}
  {Arbitrary cutting of deformable tetrahedralized objects}.{\BBCQ}
\newblock
 \APACrefbtitle {Proceedings of the 2007 ACM SIGGRAPH/Eurographics symposium on
  Computer animation} {Proceedings of the 2007 acm siggraph/eurographics
  symposium on computer animation}\ (\BPGS\ 73--80).
\PrintBackRefs{\CurrentBib}

\bibitem [\protect \citeauthoryear {%
Sifakis%
, Shinar%
, Irving%
\BCBL {}\ \BBA {} Fedkiw%
}{%
Sifakis%
, Shinar%
\BCBL {}\ \protect \BOthers {.}}{%
{\protect \APACyear {2007}}%
}]{%
sifakis2007hybrid}
\APACinsertmetastar {%
sifakis2007hybrid}%
\begin{APACrefauthors}%
Sifakis, E.%
, Shinar, T.%
, Irving, G.%
\BCBL {} Fedkiw, R.%
\end{APACrefauthors}%
\unskip\
\newblock
\APACrefYearMonthDay{2007}{}{}.
\newblock
{\BBOQ}\APACrefatitle {Hybrid simulation of deformable solids} {Hybrid
  simulation of deformable solids}.{\BBCQ}
\newblock
 \APACrefbtitle {Proceedings of the 2007 ACM SIGGRAPH/Eurographics symposium on
  Computer animation} {Proceedings of the 2007 acm siggraph/eurographics
  symposium on computer animation}\ (\BPGS\ 81--90).
\PrintBackRefs{\CurrentBib}

\bibitem [\protect \citeauthoryear {%
Smith%
, Goes%
\BCBL {}\ \BBA {} Kim%
}{%
Smith%
\ \protect \BOthers {.}}{%
{\protect \APACyear {2018}}%
}]{%
smith2018neohookean}
\APACinsertmetastar {%
smith2018neohookean}%
\begin{APACrefauthors}%
Smith, B.%
, Goes, F.D.%
\BCBL {} Kim, T.%
\end{APACrefauthors}%
\unskip\
\newblock
\APACrefYearMonthDay{2018}{{\APACmonth{03}}}{}.
\newblock
{\BBOQ}\APACrefatitle {Stable Neo-Hookean Flesh Simulation} {Stable neo-hookean
  flesh simulation}.{\BBCQ}
\newblock
\APACjournalVolNumPages{ACM Trans. Graph.}{37}{2}{}.
\newblock

\newblock

\PrintBackRefs{\CurrentBib}

\bibitem [\protect \citeauthoryear {%
Thananjeyan%
\ \protect \BOthers {.}}{%
Thananjeyan%
\ \protect \BOthers {.}}{%
{\protect \APACyear {2017}}%
}]{%
thananjeyan2017multilateral}
\APACinsertmetastar {%
thananjeyan2017multilateral}%
\begin{APACrefauthors}%
Thananjeyan, B.%
, Garg, A.%
, Krishnan, S.%
, Chen, C.%
, Miller, L.%
\BCBL {} Goldberg, K.%
\end{APACrefauthors}%
\unskip\
\newblock
\APACrefYearMonthDay{2017}{jun}{}.
\newblock
{\BBOQ}\APACrefatitle {Multilateral Surgical Pattern Cutting in 2D Orthotropic
  Gauze with Deep Reinforcement Learning Policies for Tensioning} {Multilateral
  surgical pattern cutting in 2d orthotropic gauze with deep reinforcement
  learning policies for tensioning}.{\BBCQ}
\newblock
 \APACrefbtitle {IEEE International Conference on Robotics and Automation
  (ICRA).} {Ieee international conference on robotics and automation (icra).}
\PrintBackRefs{\CurrentBib}

\bibitem [\protect \citeauthoryear {%
Tieleman%
\ \BBA {} Hinton%
}{%
Tieleman%
\ \BBA {} Hinton%
}{%
{\protect \APACyear {2012}}%
}]{%
Tieleman2012rmsprop}
\APACinsertmetastar {%
Tieleman2012rmsprop}%
\begin{APACrefauthors}%
Tieleman, T.%
\BCBT {}\ \BBA {} Hinton, G.%
\end{APACrefauthors}%
\unskip\
\newblock
\APACrefYearMonthDay{2012}{}{}.
\newblock
\APACrefbtitle {Coursera: Neural netwroks for machine learning (lecture 6.5 -
  RMSProprop).} {Coursera: Neural netwroks for machine learning (lecture 6.5 -
  rmsproprop).}
\PrintBackRefs{\CurrentBib}

\bibitem [\protect \citeauthoryear {%
Toni%
, Welch%
, Strelkowa%
, Ipsen%
\BCBL {}\ \BBA {} Stumpf%
}{%
Toni%
\ \protect \BOthers {.}}{%
{\protect \APACyear {2008}}%
}]{%
toni2008abc}
\APACinsertmetastar {%
toni2008abc}%
\begin{APACrefauthors}%
Toni, T.%
, Welch, D.%
, Strelkowa, N.%
, Ipsen, A.%
\BCBL {} Stumpf, M.P.%
\end{APACrefauthors}%
\unskip\
\newblock
\APACrefYearMonthDay{2008}{Jul}{}.
\newblock
{\BBOQ}\APACrefatitle {Approximate {B}ayesian computation scheme for parameter
  inference and model selection in dynamical systems} {Approximate {B}ayesian
  computation scheme for parameter inference and model selection in dynamical
  systems}.{\BBCQ}
\newblock
\APACjournalVolNumPages{Journal of The Royal Society
  Interface}{6}{31}{187–202}.
\newblock

\newblock

\PrintBackRefs{\CurrentBib}

\bibitem [\protect \citeauthoryear {%
U.S. Department~of Agriculture%
}{%
U.S. Department~of Agriculture%
}{%
{\protect \APACyear {2019}}%
}]{%
usda2019potato}
\APACinsertmetastar {%
usda2019potato}%
\begin{APACrefauthors}%
U.S. Department~of Agriculture, A.R.S.%
\end{APACrefauthors}%
\unskip\
\newblock
\APACrefYearMonthDay{2019}{}{}.
\newblock
\APACrefbtitle {FoodData Central.} {Fooddata central.}
\newblock
\begin{APACrefURL} {fdc.nal.usda.gov} \end{APACrefURL}
\PrintBackRefs{\CurrentBib}

\bibitem [\protect \citeauthoryear {%
S.~Wang%
\ \protect \BOthers {.}}{%
S.~Wang%
\ \protect \BOthers {.}}{%
{\protect \APACyear {2019}}%
}]{%
wang2019mpm}
\APACinsertmetastar {%
wang2019mpm}%
\begin{APACrefauthors}%
Wang, S.%
, Ding, M.%
, Gast, T.F.%
, Zhu, L.%
, Gagniere, S.%
, Jiang, C.%
\BCBL {} Teran, J.M.%
\end{APACrefauthors}%
\unskip\
\newblock
\APACrefYearMonthDay{2019}{{\APACmonth{07}}}{}.
\newblock
{\BBOQ}\APACrefatitle {Simulation and Visualization of Ductile Fracture with
  the Material Point Method} {Simulation and visualization of ductile fracture
  with the material point method}.{\BBCQ}
\newblock
\APACjournalVolNumPages{Proc. ACM Comput. Graph. Interact. Tech.}{2}{2}{}.
\newblock

\newblock

\PrintBackRefs{\CurrentBib}

\bibitem [\protect \citeauthoryear {%
Y.~Wang%
}{%
Y.~Wang%
}{%
{\protect \APACyear {2014}}%
}]{%
wang2014vna}
\APACinsertmetastar {%
wang2014vna}%
\begin{APACrefauthors}%
Wang, Y.%
\end{APACrefauthors}%
\unskip\
\newblock
\APACrefYear{2014}.
\unskip\
\newblock
\APACrefbtitle {Virtual Node Algorithms for Simulating and Cutting Deformable
  Solids} {Virtual node algorithms for simulating and cutting deformable
  solids}\ \APACtypeAddressSchool {\BUPhD}{}{}.
\unskip\
\newblock
\APACaddressSchool {}{UCLA}.
\PrintBackRefs{\CurrentBib}

\bibitem [\protect \citeauthoryear {%
Welling%
\ \BBA {} Teh%
}{%
Welling%
\ \BBA {} Teh%
}{%
{\protect \APACyear {2011}}%
}]{%
welling2011sgld}
\APACinsertmetastar {%
welling2011sgld}%
\begin{APACrefauthors}%
Welling, M.%
\BCBT {}\ \BBA {} Teh, Y.W.%
\end{APACrefauthors}%
\unskip\
\newblock
\APACrefYearMonthDay{2011}{}{}.
\newblock
{\BBOQ}\APACrefatitle {Bayesian Learning via Stochastic Gradient Langevin
  Dynamics.} {Bayesian learning via stochastic gradient langevin
  dynamics.}{\BBCQ}
\newblock
 \APACrefbtitle {International Conference on Machine Learning (ICML)}
  {International conference on machine learning (icml)}\ (\BPG~681-688).
\PrintBackRefs{\CurrentBib}

\bibitem [\protect \citeauthoryear {%
{Wijayarathne}%
, {Sima}%
, {Zhou}%
, {Zhao}%
\BCBL {}\ \BBA {} {Hammond}%
}{%
{Wijayarathne}%
\ \protect \BOthers {.}}{%
{\protect \APACyear {2020}}%
}]{%
wijayarathne2020cut}
\APACinsertmetastar {%
wijayarathne2020cut}%
\begin{APACrefauthors}%
{Wijayarathne}, L.%
, {Sima}, Q.%
, {Zhou}, Z.%
, {Zhao}, Y.%
\BCBL {} {Hammond}, F.L.%
\end{APACrefauthors}%
\unskip\
\newblock
\APACrefYearMonthDay{2020}{}{}.
\newblock
{\BBOQ}\APACrefatitle {Simultaneous Trajectory Optimization and Force Control
  with Soft Contact Mechanics} {Simultaneous trajectory optimization and force
  control with soft contact mechanics}.{\BBCQ}
\newblock
 \APACrefbtitle {IEEE/RSJ International Conference on Intelligent Robots and
  Systems (IROS)} {Ieee/rsj international conference on intelligent robots and
  systems (iros)}\ (\BPG~3164-3171).
\PrintBackRefs{\CurrentBib}

\bibitem [\protect \citeauthoryear {%
Wolper%
\ \protect \BOthers {.}}{%
Wolper%
\ \protect \BOthers {.}}{%
{\protect \APACyear {2020}}%
}]{%
wolper2020anisompm}
\APACinsertmetastar {%
wolper2020anisompm}%
\begin{APACrefauthors}%
Wolper, J.%
, Chen, Y.%
, Li, M.%
, Fang, Y.%
, Qu, Z.%
, Lu, J.%
\BDBL {}Jiang, C.%
\end{APACrefauthors}%
\unskip\
\newblock
\APACrefYearMonthDay{2020}{}{}.
\newblock
{\BBOQ}\APACrefatitle {AnisoMPM: Animating Anisotropic Damage Mechanics}
  {Anisompm: Animating anisotropic damage mechanics}.{\BBCQ}
\newblock
\APACjournalVolNumPages{ACM Trans. Graph.}{39}{4}{}.
\newblock

\newblock

\PrintBackRefs{\CurrentBib}

\bibitem [\protect \citeauthoryear {%
Wolper%
\ \protect \BOthers {.}}{%
Wolper%
\ \protect \BOthers {.}}{%
{\protect \APACyear {2019}}%
}]{%
wolper2019cdmpm}
\APACinsertmetastar {%
wolper2019cdmpm}%
\begin{APACrefauthors}%
Wolper, J.%
, Fang, Y.%
, Li, M.%
, Lu, J.%
, Gao, M.%
\BCBL {} Jiang, C.%
\end{APACrefauthors}%
\unskip\
\newblock
\APACrefYearMonthDay{2019}{}{}.
\newblock
{\BBOQ}\APACrefatitle {CD-MPM: Continuum Damage Material Point Methods for
  Dynamic Fracture Animation} {Cd-mpm: Continuum damage material point methods
  for dynamic fracture animation}.{\BBCQ}
\newblock
\APACjournalVolNumPages{ACM Trans. Graph.}{38}{4}{}.
\newblock

\newblock

\PrintBackRefs{\CurrentBib}

\bibitem [\protect \citeauthoryear {%
C.~Wu%
, Guo%
\BCBL {}\ \BBA {} Hu%
}{%
C.~Wu%
\ \protect \BOthers {.}}{%
{\protect \APACyear {2014}}%
}]{%
wu2014spg}
\APACinsertmetastar {%
wu2014spg}%
\begin{APACrefauthors}%
Wu, C.%
, Guo, Y.%
\BCBL {} Hu, W.%
\end{APACrefauthors}%
\unskip\
\newblock
\APACrefYearMonthDay{2014}{}{}.
\newblock
{\BBOQ}\APACrefatitle {An introduction to the LS-DYNA smoothed particle
  Galerkin method for severe deformation and failure analyses in solids} {An
  introduction to the ls-dyna smoothed particle galerkin method for severe
  deformation and failure analyses in solids}.{\BBCQ}
\newblock
 \APACrefbtitle {International LS-DYNA Users Conference} {International ls-dyna
  users conference}\ (\BPGS\ 1--20).
\PrintBackRefs{\CurrentBib}

\bibitem [\protect \citeauthoryear {%
C.~Wu%
, Wu%
, Crawford%
\BCBL {}\ \BBA {} Magallanes%
}{%
C.~Wu%
\ \protect \BOthers {.}}{%
{\protect \APACyear {2017}}%
}]{%
wU2017spg}
\APACinsertmetastar {%
wU2017spg}%
\begin{APACrefauthors}%
Wu, C.%
, Wu, Y.%
, Crawford, J.E.%
\BCBL {} Magallanes, J.M.%
\end{APACrefauthors}%
\unskip\
\newblock
\APACrefYearMonthDay{2017}{}{}.
\newblock
{\BBOQ}\APACrefatitle {Three-dimensional concrete impact and penetration
  simulations using the smoothed particle {G}alerkin method} {Three-dimensional
  concrete impact and penetration simulations using the smoothed particle
  {G}alerkin method}.{\BBCQ}
\newblock
\APACjournalVolNumPages{International Journal of Impact
  Engineering}{106}{}{1-17}.
\newblock

\newblock

\PrintBackRefs{\CurrentBib}

\bibitem [\protect \citeauthoryear {%
J.~Wu%
, Westermann%
\BCBL {}\ \BBA {} Dick%
}{%
J.~Wu%
\ \protect \BOthers {.}}{%
{\protect \APACyear {2015}}%
}]{%
wu2015survey}
\APACinsertmetastar {%
wu2015survey}%
\begin{APACrefauthors}%
Wu, J.%
, Westermann, R.%
\BCBL {} Dick, C.%
\end{APACrefauthors}%
\unskip\
\newblock
\APACrefYearMonthDay{2015}{}{}.
\newblock
{\BBOQ}\APACrefatitle {A survey of physically based simulation of cuts in
  deformable bodies} {A survey of physically based simulation of cuts in
  deformable bodies}.{\BBCQ}
\newblock
\APACjournalVolNumPages{Computer Graphics Forum}{34}{6}{161--187}.
\newblock

\newblock

\PrintBackRefs{\CurrentBib}

\end{thebibliography}

\clearpage

\begin{appendices}

\newcommand{\hbAppendixPrefix}{A}
\renewcommand{\thefigure}{\hbAppendixPrefix\arabic{figure}}
\setcounter{figure}{0}
\renewcommand{\thetable}{\hbAppendixPrefix\arabic{table}} 
\setcounter{table}{0}
\renewcommand{\theequation}{\hbAppendixPrefix\arabic{equation}} 
\setcounter{equation}{0}

\section{Additional Algorithmic Details}
\begin{algorithm}[b!]
\caption{Frank-Wolfe method}
\label{alg:frank-wolfe}
\begin{algorithmic}[1]
\State\textbf{Result:} Barycentric coordinate $u$ of point on edge $(p_1,p_2)$ with lowest signed distance w.r.t. SDF $d: \mathbb{R}^3\to\mathbb{R}$
 \State $u \gets \nicefrac{1}{2}$
 \For{$i = 0 \dots \operatorname{max\_iterations}$}
  \State $\displaystyle \delta \gets \frac{\partial}{\partial u} d \left((1-u) p_1 + u p_2\right)$
  \If{$\delta < 0$}
   \State $s \gets 1$
  \Else
   \State $s \gets 0$
  \EndIf
  \State $\displaystyle \gamma \gets \frac{2}{2 + i} $
  \State $u \gets u + \gamma(s - u)$
 \EndFor
\end{algorithmic}
\end{algorithm}

\subsection{Adam Optimizer}
\label{sec:adam}
At its core, Adam updates the parameters $\theta$ interactively as $\theta_i \gets \theta_{i-1} - \alpha \cdot \hat{m}_i / (\sqrt{\hat{v}_i} + \epsilon)$, where $\alpha$ is the learning rate, $\hat{m}_i$ and $\hat{v}_i$ represent the first and second order decaying averages (or momentum) after correcting for biases, and $\epsilon$ is a small value to prevent numerical issues.
The full algorithm is shown in Algorithm~\ref{alg:adam}. In Algorithm~\ref{alg:sgld}, we describe Stochastic Gradient Langevin Dynamics (SGLD) with the Adam update rule. A detailed description of SGLD is given in \autoref{sec:sgld}.

\begin{algorithm}[b!]
\caption{Stochastic gradient descent with Adam}
\label{alg:adam}
\begin{algorithmic}[1]
 \State \textbf{Given:} learning rate $\alpha$, initial parameters $\theta_0$, likelihood function $l(\cdot)$
 \State $m_0 \gets 0$
 \State $v_0 \gets 0$
 \For{$i = 1 \dots \operatorname{max\_iterations}$}
  \State $g_i \gets \nabla_\theta l(\theta_{i-1})$
  \State $m_i \gets \beta_1\cdot m_{i-1} + (1-\beta_1)\cdot g_i$
  \State $v_i \gets \beta_2\cdot v_{i-1} + (1-\beta_2)\cdot g_i^2$
  \State $\hat{m}_i \gets m_i / (1-\beta_1^i)$
  \State $\hat{v}_i \gets v_i / (1-\beta_2^i)$
  \State $\theta_i \gets \theta_{i-1} - \alpha \cdot \hat{m}_i / (\sqrt{\hat{v}_i} + \epsilon)$
 \EndFor
 \State \Return $\theta_i$
\end{algorithmic}
\end{algorithm}

\begin{algorithm}[t!]
 \caption{SGLD with Adam update rule}
 \label{alg:sgld}
\begin{algorithmic}[1]
\State \textbf{Given:} learning rate $\alpha$, initial parameters $\theta_0$, likelihood function $l(\cdot)$
 \State $m_0 \gets 0$
 \State $v_0 \gets 0$
 \For{$i = 1 \dots \operatorname{max\_iterations}$}
  \State $g_i \gets \nabla_\theta U(\theta_{i-1})$
  \State $m_i \gets \beta_1\cdot m_{i-1} + (1-\beta_1)\cdot g_i$
  \State $v_i \gets \beta_2\cdot v_{i-1} + (1-\beta_2)\cdot g_i^2$
  \State $\hat{m}_i \gets m_i / (1-\beta_1^i)$
  \State $\hat{v}_i \gets v_i / (1-\beta_2^i)$
  \State $A_i \gets \operatorname{diag}\left(\mathbf{1} \oslash (\sqrt{\hat{v}_i} + \epsilon) \right)$
  \State $\eta_i \sim \mathcal{N}(0, \alpha A_i)$
  \State $\theta_i \gets \theta_{i-1} - \alpha \cdot \hat{m}_i  \cdot A_i + \eta_i$
 \EndFor
 \State \textbf{return} $\theta_i$
\end{algorithmic}
\end{algorithm}

\subsection{Loss Functions}
\label{sec:loss_functions}
We investigated various loss functions for evaluating the closeness between knife force profiles effectively. Given the two-dimensional parameter inference experiment from \autoref{sec:exp-infer-self}, we test the following cost functions which we minimize via the Adam optimizer:
\begin{itemize}
    \item $L1$ loss
    \item $L2$ loss
    \item Inverse Cosine Similarity
    \item LogSumExp
\end{itemize}

\begin{figure}
    \centering
    \small
    \textbf{L1 Loss} $\left\vert \mathbf{\phi}^s - \mathbf{\phi}^r \right\vert$\\
    \adjincludegraphics[width=\columnwidth]{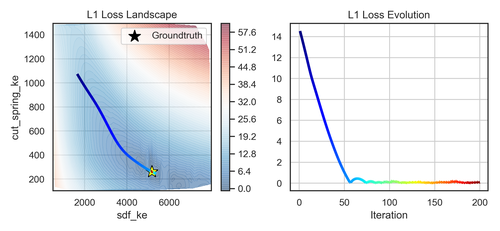}\\
    \textbf{L2 Loss} $\left\| \mathbf{\phi}^s - \mathbf{\phi}^r \right\|^2$\\
    \adjincludegraphics[width=\columnwidth]{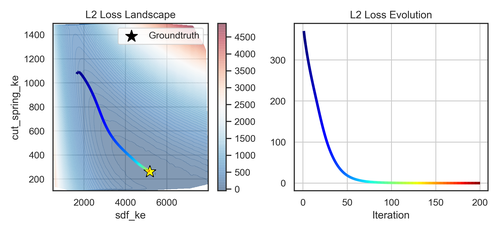}\\
    \textbf{Inverse Cosine Similarity} $1-\dfrac{\mathbf{\phi}^s \cdot \mathbf{\phi}^r}{\max(\Vert \mathbf{\phi}^s \Vert _2 \cdot \Vert \mathbf{\phi}^r \Vert _2, \epsilon)}$\\
    \adjincludegraphics[width=\columnwidth]{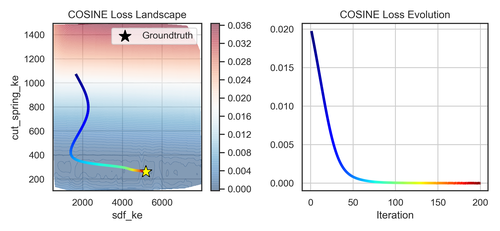}\\
    \textbf{LogSumExp} $\log\left[\sum_t \exp\left(\mathbf{\phi}^s[t] - \mathbf{\phi}^r[t]\right)\right]$\\
    \adjincludegraphics[width=\columnwidth]{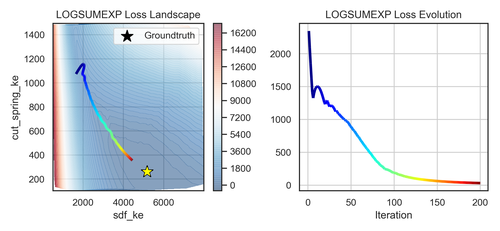}\\
    \caption{Evaluation of various loss functions on the \texttt{actual\_2d} scenario from \autoref{sec:exp-infer-self}. For 500 randomly sampled parameter vectors, the trajectories are simulated (shown as dots) and the loss is computed between the simulated trajectory $\mathbf{\phi}^s$ and the ground-truth knife force profile $\mathbf{\phi}^r$. The interpolated loss landscape is shown one the left, and the training loss evolution using the Adam optimizer is shown on the right. Overall, the $L1$ loss performs favorably compared to the other cost functions and is therefore our choice to evaluate the closeness between knife force trajectories across this work.}
    \label{fig:loss-fn}
\end{figure}

We visualize the trace of the optimized parameters in \autoref{fig:loss-fn}. The $L1$ loss function compares favorably to the other formulations as it allows for a relatively fast convergence to the true parameters, while also yielding a ``bouncing'' behavior when the iterates are close to the (local) optimum. Since we use the $L1$ loss function in the likelihood term for our probabilistic parameter inference experiments~\ref{sec:exp-infer-self}, the likelihood distribution corresponds to a Laplace distribution which is known to have sharp peaks and heavy tales.

\subsection{BayesSim Implementation}
\label{sec:bayessim-extra}
BayesSim approximates the posterior over the simulation parameters as
\begin{align*}
    p(\theta\mid\mathbf{\phi})\approx p(\theta)/\tilde{p}(\theta) \; q(\theta\mid\mathbf{\phi}),
\end{align*}
where $p(\theta)$ is the prior distribution
and $\tilde{p}(\theta)$ is a proposal prior used to 
sample the simulator and collect $N$ samples, $ \{\theta_i, \mathbf{\phi}_i^s\}_{i=1}^N $, to learn $ q(\theta\mid\mathbf{\phi}) $. 
When a real trajectory $ \mathbf{\phi}^r $ is observed, BayesSim computes
$ p(\theta\mid\mathbf{\phi}=\mathbf{\phi}^r) $ that represents the posterior over
simulation parameters given the real data. 
As inferring simulator parameters given trajectories is a type of inverse problem, it can admit a multitude of solutions.

For the BayesSim baseline, we train a mixture density network (MDN) representing $q(\theta\mid\mathbf{\phi})$ with 10 components on a dataset consisting of 500 knife force trajectories that have been generated in our simulator by uniformly sampling the simulation parameters within their respective bounds. Depending on the experiment, we limit ourselves to only a subset of all the available parameters, due to the exponential increase in sample complexity.
As summary statistics input to BayesSim, we downsample the \SI{0.9}{\second} knife force profiles consisting of 90,000 time steps by a factor of 1000 to 90-dimensional summary statistics using polyphase filtering.
While training the MDN, we project the true parameter values to the unit interval using their possible ranges (see~\autoref{tab:parameters}), as we found the MDN to be sensitive to the scale of the parameters, and to perform more accurately with homogeneous value ranges.

\subsection{Optimal Transport of Cutting Spring Parameters}
\label{sec:optimal-transport}

The Earth Mover's Distance can be interpreted as the minimal cost associated with transforming a constant volume pile of dirt into another, where the cost is defined as the amount of dirt moved multiplied by the distance travelled. Formally, given two sets of vertices $P$ and $Q$, 
associated sets of vertex weights $\mathbf{w}_P$ and $\mathbf{w}_Q$,
and a cost matrix given by the Euclidean distance   
between two points $D=\vert d_{i,j} \vert = \|p_i - q_j\|^2$, optimal transport finds the solution of the optimization problem
\begin{align}
    \label{eq:emd}
    \min_F \frac{1}{Z}\sum_{i=1}^m\sum_{j=1}^n f_{i,j} \, d_{i,j},
\end{align}
where $F=f_{i,j}$ is the movement or flow between $p_i$ and $q_j$ which we try to minimize, and $Z=\sum_{i=1}^m \sum_{j=1}^n f_{i,j}$ is a normalization constant. 
In our formulation, we assume uniform weights for both sets of vertices. 

This problem can be efficiently solved using the network simplex algorithm~\cite{Cunningham1976simplex} and has typical complexity $O(n^3)$, but sparsity can be exploited to reduce this cost.

\begin{figure}
    \centering
    \renewcommand{\figheight}[0]{5cm}
    \includegraphics[width=\columnwidth]{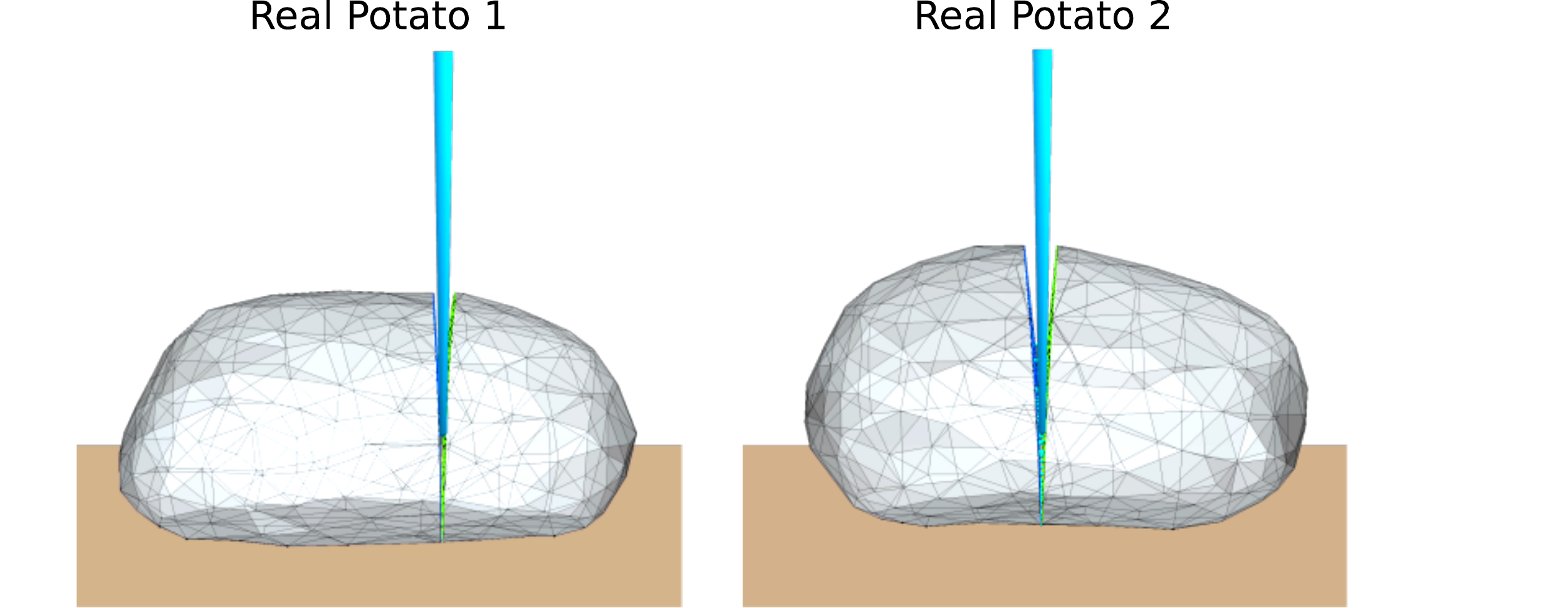}\\
    \includegraphics[width=0.9\linewidth]{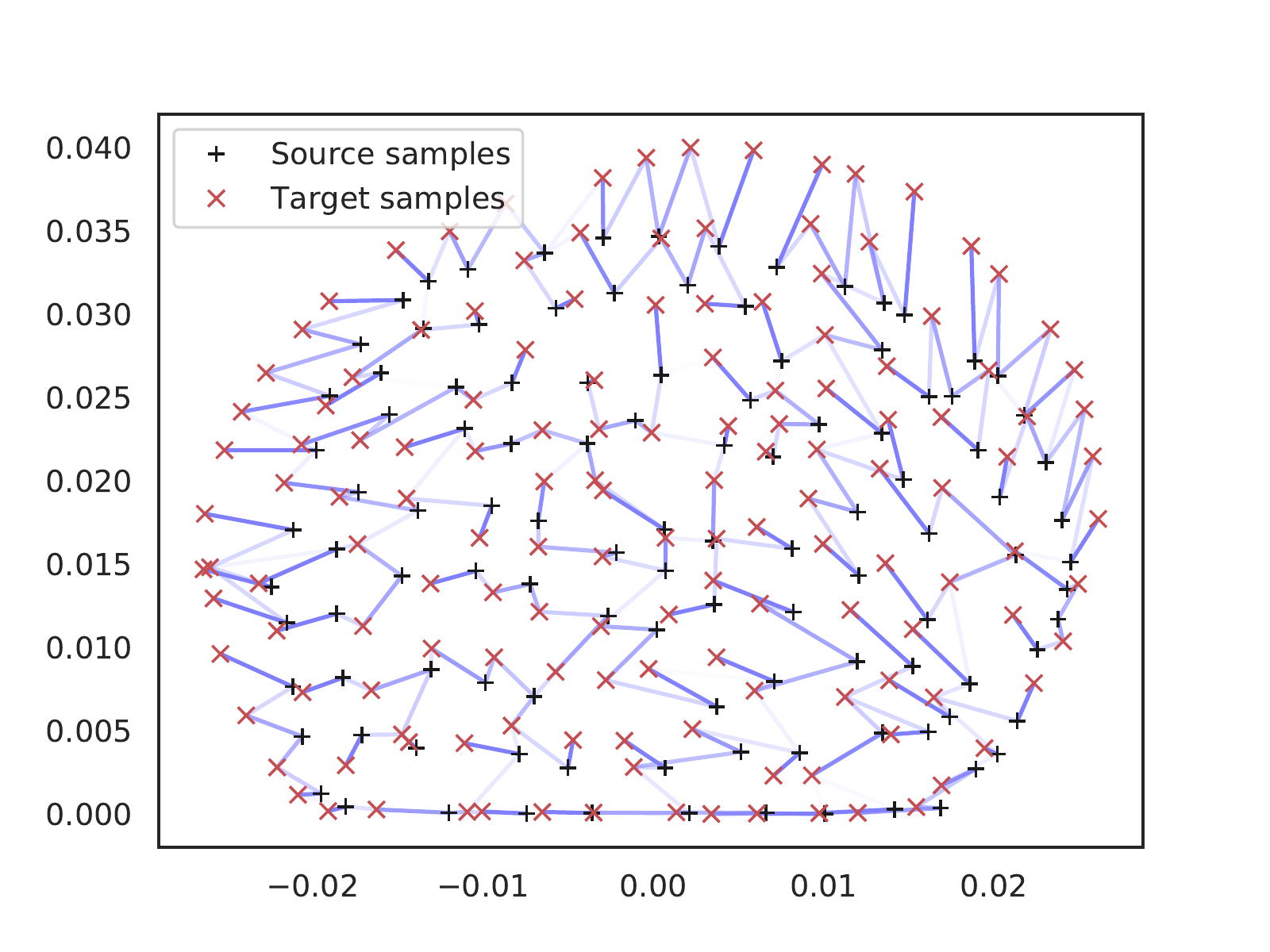}\\
    \includegraphics[width=0.9\linewidth]{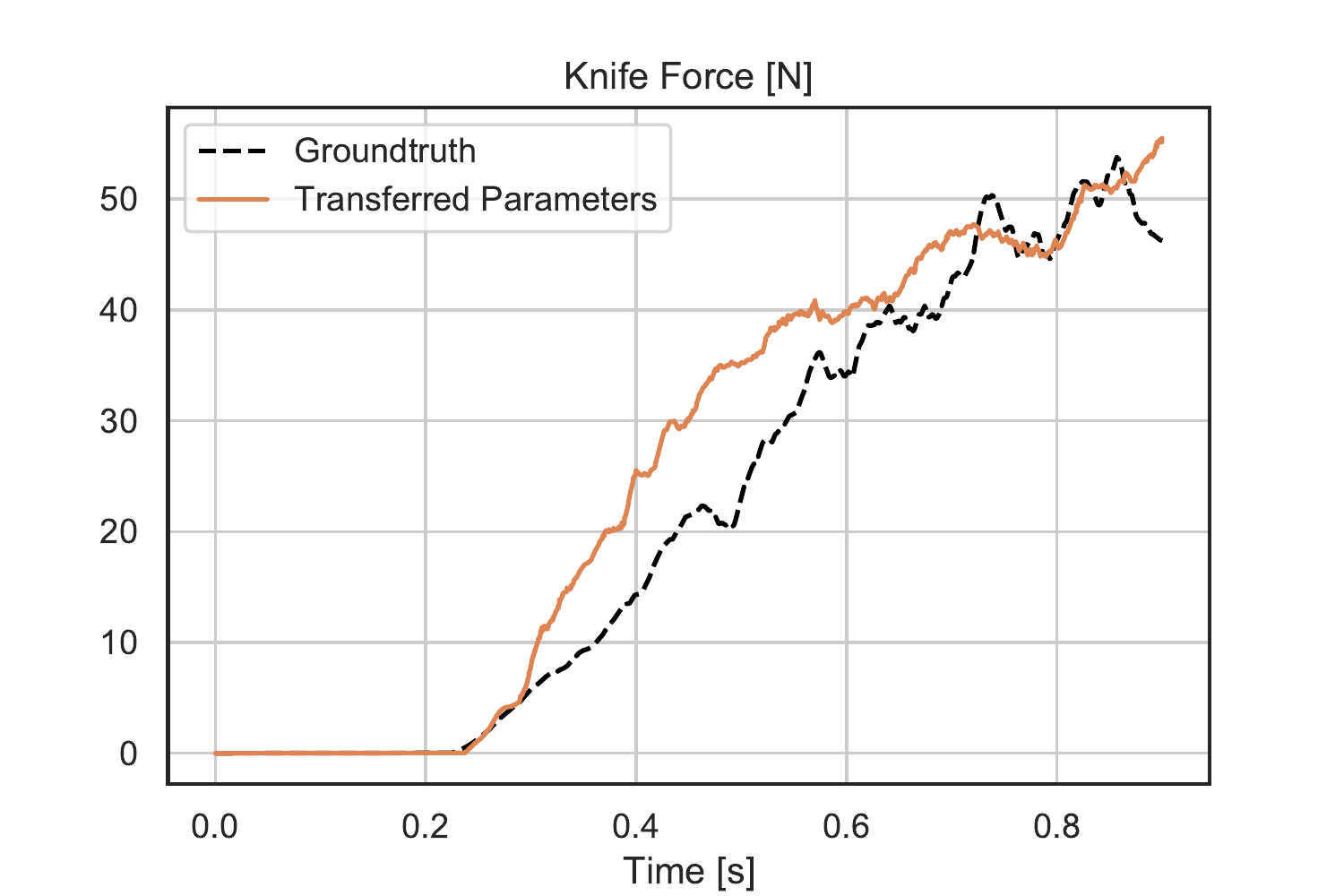}
    \caption{Optimal transport of simulation parameters from one mesh of a real potato (top left) to another real potato mesh (top right). The correspondences (center) have been found via optimal transport using the Earth Mover's Distance (\autoref{eq:emd}) between cutting springs in two potato meshes. The weighted mapping between the 2D positions of the springs at the cutting interface from the source domain (\texttt{ybj\_potato1}) to the target domain (\texttt{ybj\_potato2}) is used to transfer the cutting spring parameters between the two topologically different meshes, resulting in a close fit to the ground-truth trajectory (bottom).}
    \label{fig:ybj-potato-optimal-transport}
\end{figure}

\section{Details of Experimental Setup}

\subsection{Simulation Parameters}
\methodname introduces additional degrees of freedom via the insertion of virtual nodes and cutting springs that connect them (see \autoref{sec:preprocess}). In \autoref{tab:parameters}, we list each available parameter with its description and default value. Parameters that are allowed to be individually tuned for each spring can be set in two modes:
\begin{itemize}
    \item Shared parameterization: the parameter is a single scalar that gets replicated across all cutting springs.
    \item Individual parameterization: the parameter is a vector where each entry can be tuned separately for each spring.
\end{itemize}

To enforce hard parameter limits in our simulator, throughout our experiments, we impose bounds on the estimated simulation parameters by projecting them through the sigmoid function. Thus, given an unconstrained real number $x$ to be optimized, the resulting projected parameter value is $\operatorname{sigmoid}(x) \cdot (p_{ub}-p_{lb}) + p_{lb}$, where $p_{lb}$ and $p_{ub}$ are the upper and lower bounds of the parameter, and $\operatorname{sigmoid}(x) = 1/(1+\exp(-x))$.

\begin{table}[]
    \centering
    \begin{tabularx}{\columnwidth}{Xrcc}
    \toprule
        \bf Material & \bf $E$ (\si{\N\per\m\tothe{2}}) & $\nu$ & $\rho$ (\si{\kg\per\m\tothe{3}}) \\ \midrule
        Apple\footnote{\cite{jamdagni2019robotic}} & $3.0\times10^6$ & 0.17 & 787 \\
        Potato\footnotemark[4] & $2.0\times10^6$ & 0.45 & 630 \\
        Cucumber\footnotemark[4] & $2.5\times10^6$ & 0.37 & 950 \\
        Banana\footnote{\cite{kulkarni1983banana}} & $0.003048\times10^6$ & 0.28 & 1350 \\
    \bottomrule\\
    \end{tabularx}
    \caption{Properties of common biomaterials: Young's modulus $E$, Poisson's ratio $\nu$, density $\rho$.}
    \label{tab:materials}
\end{table}

\begin{table*}[]
    \centering
    \begin{adjustbox}{width=\textwidth}
    \begin{tabular}{llll}
        \toprule
        \bf Name & \bf Description & \multicolumn{2}{c}{\bf Default value} \\ \midrule
        \multicolumn{4}{l}{\bf Knife geometry parameters (fixed)} \\
        \verb|edge_dim| & Lower diameter of knife (see \autoref{fig:knife-params} right) & 0.08 & \si{\mm} \\
        \verb|spine_dim| & Upper diameter of knife (spine) & 2 & \si{\mm} \\
        \verb|spine_height| & Height of knife spine & 40 & \si{\mm} \\
        \verb|tip_height| & Height of knife tip & 0.04 & \si{\mm} \\
        \verb|depth| & Length of knife blade (along $z$ axis) & 150 & \si{\mm} \\
        \midrule%
        \multicolumn{4}{l}{\bf Knife motion} \\ 
        \verb|velocity_y| & Vertical knife velocity & -0.05 & \si{\m\per\s} \\
        \verb|initial_y| & Initial vertical knife position (height) & 80 & \si{\mm} \\
        \midrule%
        \multicolumn{4}{l}{\bf Spring-damper parameters (individual for each spring)} \\
        \verb|cut_spring_ke| & Spring stiffness coefficient at the start of the simulation (initial stiffness of the cutting spring) & 500 &  \\
        \verb|cut_spring_kd| & Spring damping coefficient & 0.1 &  \\
        \verb|cut_spring_softness| & Softness coefficient $\gamma$ used in the linear spring loosening (\autoref{eq:loosen-spring}) & 500 &  \\
        \midrule%
        \multicolumn{4}{l}{\bf Knife contact dynamics parameters (individual for each spring)} \\
        \verb|sdf_radius| & Radius around SDF to consider for contact dynamics & 0.5 & \si{\mm} \\
        \verb|sdf_ke| & Positional penalty coefficient (contact normal stiffness) & 1000 &  \\
        \verb|sdf_kd| & Damping coefficient & 1 &  \\
        \verb|sdf_kf| & Contact friction stiffness (tangential stiffness used in Coulomb friction model) & 0.01 &  \\
        \verb|sdf_mu| & Friction coefficient ($\mu$) & 0.5 &  \\
        \midrule%
        \multicolumn{4}{l}{\bf Ground contact dynamics parameters (fixed)} \\
        \verb|ground_ke| & Positional penalty coefficient (contact normal stiffness) & 100 &  \\
        \verb|ground_kd| & Damping coefficient & 0.1 &  \\
        \verb|ground_kf| & Contact friction stiffness (Coulomb friction model) & 0.2 &  \\
        \verb|ground_mu| & Friction coefficient ($\mu$) & 0.6 &  \\
        \verb|ground_radius| & Radius around mesh vertices & 1 & \si{\mm} \\
        \midrule%
        \multicolumn{4}{l}{\bf Material properties} \\
        \verb|young| & Young's modulus $E$ & \multicolumn{2}{c}{see \autoref{tab:materials}} \\
        \verb|poisson| & Poisson's ratio $\nu$ & \multicolumn{2}{c}{see \autoref{tab:materials}} \\
        \verb|density| & Density $\rho$ & \multicolumn{2}{c}{see \autoref{tab:materials}} \\
        \bottomrule\\
    \end{tabular}
    \end{adjustbox}
    \caption{Overview of the model parameters used by our cutting simulator.}
    \label{tab:parameters}
\end{table*}

\begin{figure}
    \centering
    \includegraphics[width=0.3\textwidth]{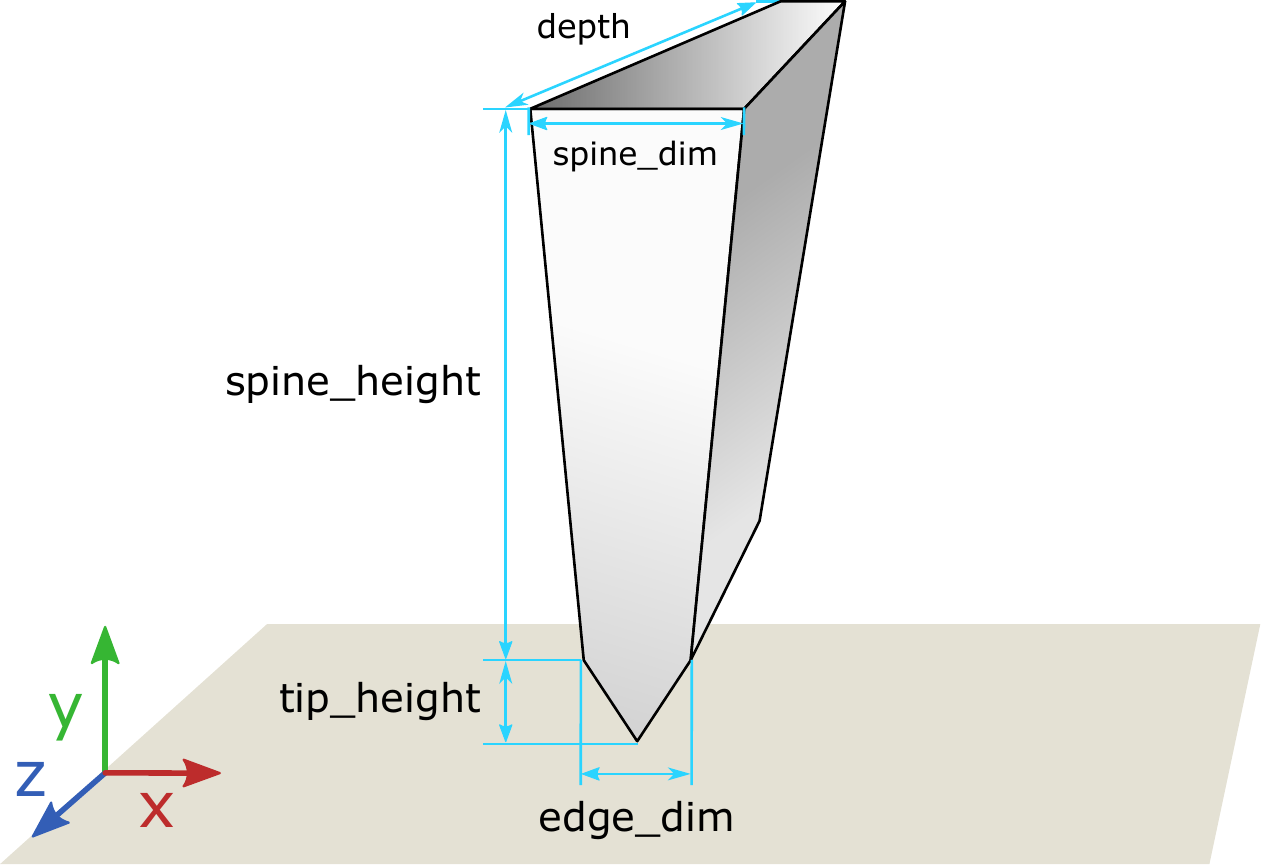}
    \caption{Parameters that describe the knife geometry.}
    \label{fig:knife-params}
\end{figure}

\subsection{Commercial Simulation Setup}
\label{sec:ansys-setup}
In the commercial solver, each simulation consisted of a rigid knife, 1 of 3 deformable fruits/vegetables (apple, cucumber, or potato), and a rigid table. The apple, cucumber, and potato geometries were represented by sphere, cylinder, and rectangular shape primitives, respectively. Each primitive had a \SI{10}{\mm}-thick slice, centered along the long axis, that defined the volumetric region that could be cut. All regions were assigned an isotropic elastic failure material model, with density, Lam\'e parameters (see \autoref{tab:materials}), yield stress, and failure strain obtained from the agricultural mechanics literature~\citep{li2018apple,mousavizadeh2010cucumber,el2011cucumber} and the FoodData Central database from the U.S. Department of Agriculture~\citep{usda2019potato}. The non-slice regions were simulated using FEM with a tetrahedral mesh-based discretization. The slice region was simulated using the smoothed particle Galerkin (SPG) method with a particle-based discretization. SPG is a numerical method related to smoothed particle hydrodynamics (SPH)~\citep{liu2010sph, monaghan1992sph} and the element-free Galerkin method (EFG)~\citep{belytschko1994efg}, and has been validated for simulating large deformation and failure of elastoplastic solids~\citep{wu2014spg}. Continuity conditions were imposed between the non-slice mesh and the slice particles. Coulomb frictional contact was defined between the knife and the deformable object, as well as between the object and the table, with a coefficient of friction of 0.6. A constant downward velocity was applied to the knife until contact with the surface of the table, and gravity was applied to the deformable object. 

\section{Additional Experiments}

\subsection{Generalization Results}
\label{sec:more-generalization}
\begin{figure}[t!]
    \centering
    \includegraphics[width=0.9\linewidth]{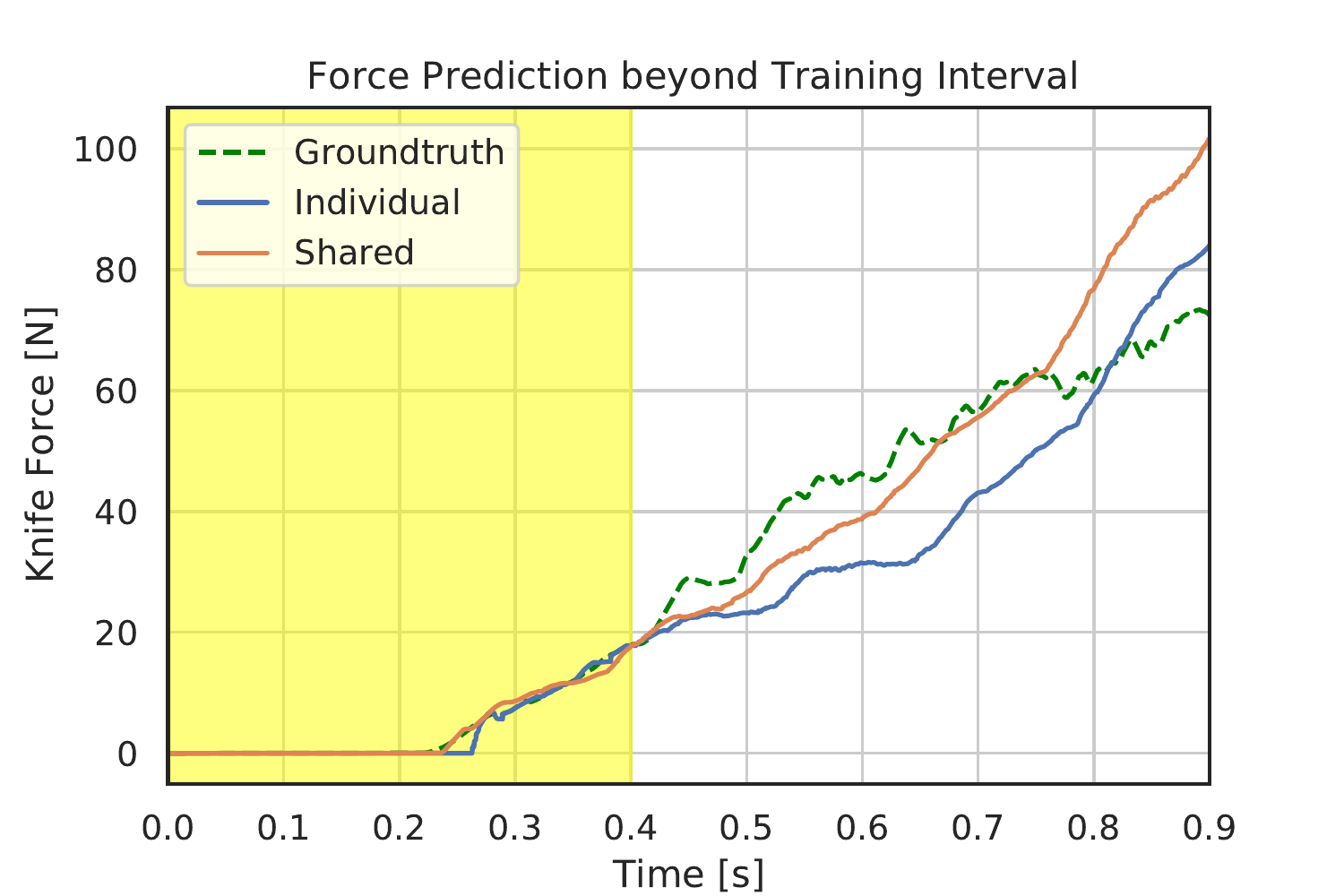}
    \caption{Generalization performance on the apple cutting trajectory by simulating the parameters which have been optimized from a \SI{0.4}{\second} ground-truth trajectory (highlighted in yellow) from a commercial simulator. At test time, the force profile is simulated over a duration of \SI{0.9}{\second} (see \autoref{sec:exp-ansys-duration}). The blue line shows the estimated trajectory when all parameters were optimized individually for each cutting spring. Shown in orange is the resulting trajectory from optimizing parameters shared across all cutting springs.}
    \label{fig:exp-ansys-duration}
\end{figure}

\autoref{fig:exp-ansys-duration} shows the trajectories for the generalization experiment (\autoref{sec:exp-ansys-duration}) where the parameters have been optimized for a shorter simulation time window of \SI{0.4}{\second}, and tested against the ground-truth trajectory (green, dashed line) over a duration of \SI{0.9}{\second}.

\begin{figure}[t!]
    \centering
    \includegraphics[width=\columnwidth]{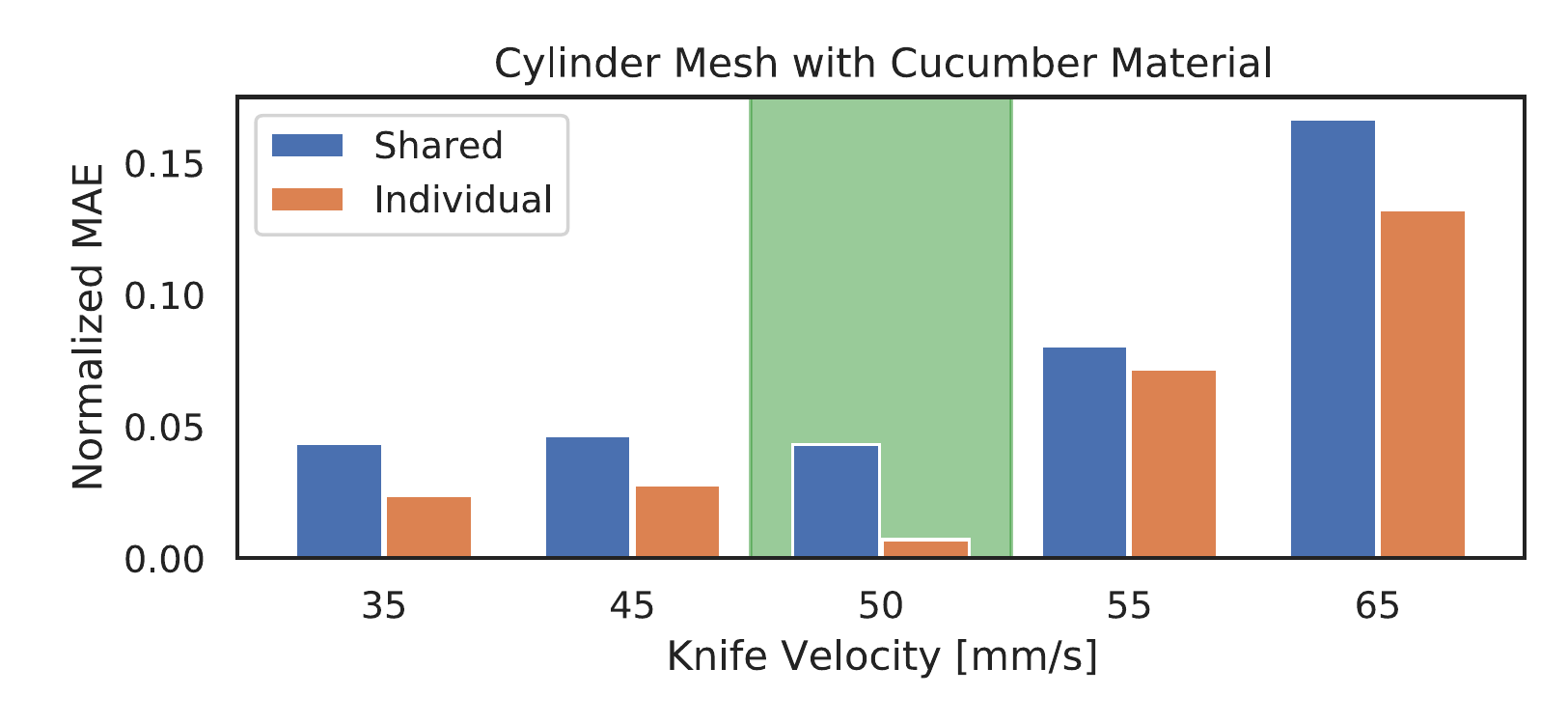}
    \caption{Velocity generalization results for a cylinder mesh with cucumber material properties given ground-truth simulations with different vertical knife velocities from a commercial solver. Two versions of \methodname were calibrated: by sharing the parameters across all cutting springs (blue) and by optimizing each value individually (orange), given a ground-truth trajectory with the knife sliding down at \SI{50}{\mm\per\second} speed (highlighted in green). The normalized mean absolute error (MAE) is evaluated against the ground-truth by rolling out the estimated parameters for the given knife velocity.}
    \label{fig:exp-ansys-velocity-cucumber}
\end{figure}

Analogous to the apple cutting results with ground-truth from the commercial solver in \autoref{fig:exp-ansys-velocity-apple}, the bar plot in \autoref{fig:exp-ansys-velocity-cucumber} visualizes the normalized mean absolute error (NMAE) for different knife velocities at test time. The simulation parameters have been optimized for a vertical downward velocity of \SI{50}{\mm} (green shade in the background), given a knife force profile of cutting a cylindrical mesh with cucumber material properties from the commercial simulator. At test time, we use the same optimized simulation parameters to evaluate the accuracy of the force profile simulation against the commercial simulator on different downward velocities. As in the experiments of cutting an apple (\autoref{fig:exp-ansys-velocity-apple}), the results in \autoref{fig:exp-ansys-velocity-cucumber} show that the individual cutting spring parameterization outperforms the shared parameterization across all tested velocities. While the performance improvement over the shared parameterization becomes less significant, it can be concluded that the individual parameterization generalizes better to novel test velocities.

\subsection{Posterior Over Simulation Parameters}
\label{sec:ansys-apple-shared}
\begin{figure}[H]
    \centering
    \includegraphics[width=0.9\textwidth]{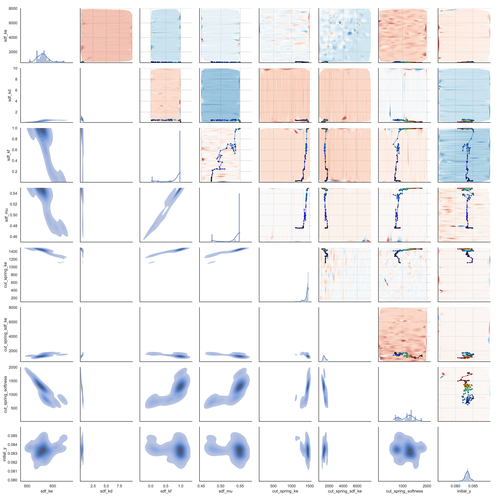}
    \caption{Posterior obtained by SGLD in our differentiable simulator after 300 trajectory roll-outs. Marginals are shown on the diagonal. The sampled chain is visualized for pairs of parameter dimensions in the upper triangle, along with an approximate rendering of the loss surface as heatmap in the background. The heatmaps in the lower triangle visualize the kernel densities for all pair-wise combinations of the parameters.}
    \label{fig:ansys-apple-sgld-posterior}
\end{figure}

\begin{figure}[H]
    \centering
    \includegraphics[width=0.9\textwidth]{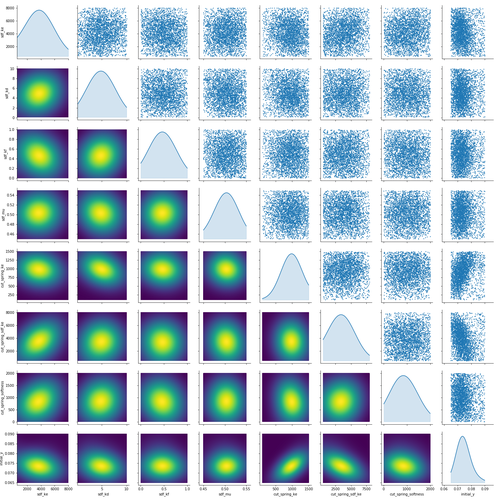}
    \caption{Posterior obtained by BayesSim after 100 iterative updates of the training dataset with 20 new trajectories per iteration. Marginals are shown on the diagonal. Parameter samples are represented by the dots in the scatter plots from the upper triangle. The heatmaps in the lower triangle visualize the kernel densities for all pair-wise combinations of the parameters.}
    \label{fig:ansys-apple-bayessim-posterior}
\end{figure}

The marginal plots in \autoref{fig:ansys-apple-sgld-posterior} and \autoref{fig:ansys-apple-bayessim-posterior} show the posteriors from iterative BayesSim and SGLD, respectively. These densities over a subset of the simulation parameters (shared across the cutting springs) have been inferred from knife force profiles of cutting a sphere mesh using apple material properties. The ground-truth, from which the posteriors are inferred, stems from a high-fidelity commercial simulator (see description in \autoref{sec:exp-ansys}).

\section{Code Snippets}
\label{sec:code-snippets}

We provide abridged Python code samples that demonstrate the usage of our simulator, whose source code we released previously\footnote{\url{https://github.com/NVlabs/DiSECt}}. Our simulator provides gradients of all its parameters w.r.t. the calculated force profile (and any other simulation outputs). We embed the differentiable dynamics function as a PyTorch~\citep{paszke2019pytorch} module, such that DiSECt can be leveraged by the existing deep learning ecosystem, including gradient-based optimizers and tensor operations that manipulate the inputs and outputs of the simulator.

\newpage
\onecolumn

\newpage

\newpage

\subsection{Parameter Inference}
The following code snippet demonstrates how the differentiability of DiSECt can be leveraged to infer the simulation parameters from real-world force measurements. We instantiate our simulator to cut an apple in the same way it was cut in the real world, by applying a prescribed knife motion that follows the recorded end-effector trajectory. Note that the simulation settings which are loaded from a JSON file include the file name of the tetrahedral mesh of the fruit we cut, as well as settings pertaining to the cutting springs and contact model, the material properties, boundary conditions, and integration procedure, among others (see \autoref{tab:parameters}). Given the recorded force profile, we select the Adam optimizer to tune the simulation parameters in a way that the L2 error between the simulated and real force profile is minimized.

\begin{lstlisting}[language=Python, caption=Code example where the simulation parameters are inferred from real force measurements through gradient-based optimization.]
from cutting import CuttingSim, load_settings, PredefinedCartesianMotion
# load the simulation settings for a real apple
settings = load_settings("experiments/config/real_apple.json")
# create DiSECt instance with GPU support (CUDA) and enabled gradient computation
sim = CuttingSim(settings, dataset="real", adapter="cuda", requires_grad=True)
# load data (force, position measurements) that we recorded from the real robot
with open("assets/real/apple/vertical_cutting_apple.pkl", "rb") as f:
    real_data = pickle.load(f)
real_force = np.linalg.norm(real_data["force"][:, :3], axis=1)  # force profile
real_dt = real_data["time"][1] - real_data["time"][0]  # time step of the recorded measurements
ee_pos = real_data["position"][:, :3]  # trajectory of end-effector 3D positions
ee_vel = real_data["velocity"][:, :3]  # trajectory of end-effector 3D velocities
times = real_data["time"]  # time stamps of the recorded real measurements
# create a prescribed knife motion from the recorded end-effector trajectory
sim.motion = PredefinedCartesianMotion(q=ee_pos, qd=ee_vel, time=times)
# select reference force profile, optionally interpolate in case the recording time step does 
# not match the simulation time step
sim.load_groundtruth(real_force, groundtruth_dt=real_dt)
sim.cut()  # preprocessing step that cuts the mesh and inserts cutting springs
opt_params = sim.init_parameters()  # retrieve tensors of simulation parameters to be optimized
opt = torch.optim.Adam(opt_params, lr=learning_rate)
# optimization loop in which the Adam optimizer tunes the simulation parameters to minimize the L2 error
# between the real and simulated force profile
for iteration in range(num_iters):
    hist_knife_force = sim.simulate()  # simulate a force profile
    # compute L2 loss of force profile against ground-truth
    loss = torch.square(hist_knife_force - sim.groundtruth_torch[:len(hist_knife_force)]).mean()
    # log some information of the inference process (loss, plot of force profile and knife trajectory)
    logger.add_scalar("loss", loss.item(), iteration)
    fig = sim.plot_simulation_results()
    logger.add_figure("simulation", fig, iteration)
    opt.zero_grad()
    loss.backward()
    # update the simulation parameters
    opt.step()
\end{lstlisting}

\newpage

\subsection{Slicing Motion Optimization}
In the following code snippet, we implement our trajectory optimization loop from \autoref{sec:exp-control} where a lateral slicing motion is tuned to minimize the force that acts on the knife. Following \autoref{eq:slicing-objective}, we add hard constraints on the motion of the knife to ensure that the blade touches the cutting board at the end of the cut (i.e., the object is fully cut), and that the lateral motion does not exceed the length of the blade.

\begin{lstlisting}[language=Python, caption=Code example where the knife's slicing motion is optimized to reduce the knife force while adhering to constraints.]
import mdmm  # package that implements the Modified Differential Method of Multipliers (MDMM)
from cutting import CuttingSim, load_settings, SlicingMotion
# load the simulation settings for the right half of a real apple we cut before
settings = load_settings("experiments/config/real_apple_right.json")
# create DiSECt instance with GPU support (CUDA) and enabled gradient computation
sim = CuttingSim(settings, dataset="real", adapter="cuda", requires_grad=True)
slicing_waypoints = 5  # number of waypoints in the trajectory
slicing_kernel_width = 0.4  # widths of the kernels of the trajectory waypoints
# the following tensors parameterize the knife motion, and require gradients to be optimizable
slicing_amplitudes = torch.tensor(np.ones(slicing_waypoints) * 0.001, requires_grad=True)
pressing_velocities = torch.tensor(np.ones(slicing_waypoints) * velocity_y, requires_grad=True)
slicing_frequency = torch.tensor(5., requires_grad=True)
# set up slicing motion of the knife with the parameters that we will optimize
sim.motion = SlicingMotion(
    initial_pos=torch.tensor([0.0, sim.settings.initial_y, 0.], device="cuda"),
    slicing_frequency=slicing_frequency, slicing_amplitudes=slicing_amplitudes,
    pressing_velocities=pressing_velocities, slicing_kernel_width=slicing_kernel_width,
    slicing_times=np.linspace(0.0, settings.sim_duration, slicing_waypoints))
sim.cut()  # preprocessing step that cuts the mesh and inserts cutting springs
# add constraint to have the blade touch the ground at the end of the cut
knife_height_constraint = mdmm.EqConstraint(
    lambda: sim.hist_knife_pos[-1, 1] + sim.blade_bottom_center_point_offset[1],
    0.0, scale=height_constraint_scale, damping=height_constraint_damping)
# determine blade length and maximal lateral extents of the mesh
blade_length = sim.knife.depth
mesh_min_z = np.array(sim.builder.particle_q)[:, 2].min()
mesh_max_z = np.array(sim.builder.particle_q)[:, 2].max()
blade_min = mesh_max_z - blade_length / 2
blade_max = mesh_min_z + blade_length / 2
# add constraint to prevent the knife from moving laterally further than its blade length
blade_length_constraint = mdmm.EqConstraint(
    lambda: (sim.hist_knife_pos[:, 2].clamp(blade_min, blade_max) - sim.hist_knife_pos[:, 2]).sum(),
    0.0, scale=blade_constraint_scale, damping=blade_constraint_damping)
# set up MDMM optimization problem with the constraints; use Adamax as optimizer
constraints = [knife_height_constraint, blade_length_constraint]
mdmm_module = mdmm.MDMM(constraints)
parameters = [slicing_frequency, slicing_amplitudes, pressing_velocities]
opt = mdmm_module.make_optimizer(parameters, optimizer=torch.optim.Adamax)
# optimize knife motion parameters to minimize mean knife force while adhering to the constraints
for iteration in range(num_iters):
    # objective is to minimize mean knife force
    hist_knife_force = sim.simulate()
    mean_knife_force = torch.mean(hist_knife_force)
    mdmm_return = mdmm_module(mean_knife_force)
    # log the current average knife force, add plots of the simulation and the trajectory parameters
    logger.add_scalar("mean_knife_force", mean_knife_force.item(), iteration)
    fig = sim.plot_simulation_results()
    logger.add_figure("simulation", fig, iteration)
    fig = sim.motion.plot(sim_duration=settings.sim_duration)
    logger.add_figure("motion", fig, iteration)
    opt.zero_grad()
    # calculate gradient of the MDMM Lagrangian
    mdmm_return.value.backward()
    opt.step()  # update motion parameters
\end{lstlisting}

\end{appendices}

\end{document}